\documentclass{theme/uniprthesis}

\usepackage[english,italian]{babel}		
\usepackage{csquotes}
\usepackage[acronym]{glossaries}
\usepackage{todonotes}			
\usepackage[style=apa]{biblatex}
\usepackage{bm,bbm,amsfonts,amssymb}
\usepackage{xspace}
\usepackage{eucal} 
\usepackage{xurl}
\usepackage[T1]{fontenc}
\usepackage[inline]{enumitem}
\usepackage{pdflscape}

\usepackage{booktabs}
\usepackage{longtable}
\usepackage[pass]{geometry}
\usepackage{hyperref}
\usepackage{varioref}
\usepackage{cleveref}
\usepackage{notation}

\makeglossaries
\newacronym{pbge}{BGE}{Bidimensional Gripper Environment}
\newacronym{ale}{ALE}{Arcade Learning Environment}

\newacronym{rl}{RL}{Reinforcement Learning}
\newacronym{drl}{DRL}{Deep Reinforcement Learning}
\newacronym{rlhf}{RLHF}{Reinforcement Learning from Human Feedback}

\newacronym{ai}{AI}{Artificial Intelligence}
\newacronym{ml}{ML}{Machine Learning}
\newacronym{dl}{DL}{Deep Learning}
\newacronym{sl}{SL}{Supervised Learning}
\newacronym{marl}{MARL}{Multi-Agent Reinforcement Learning}

\newacronym{aan}{AAN}{Artificial Neural Network}
\newacronym{cnn}{CNN}{Convolutional Neural Network}
\newacronym{dnn}{DNN}{Deep Neural Network}

\newacronym{dqn}{DQN}{Deep Q-Network}
\newacronym{ppo}{PPO}{Proximal Policy Optimisation}
\newacronym{qrdqn}{QRDQN}{Quantile Regression DQN}
\newacronym{trpo}{TRPO}{Trust Region Policy Optimisation}
\newacronym{a2c}{A2C}{Advantage Actor-Critic}
\newacronym{a3c}{A3C}{Asynchronous Advantage Actor-Critic}
\newacronym{sarsa}{SARSA}{State–action–reward–state–action}
\newacronym{reinforce}{REINFORCE}{REward Increment = Nonnegative Factor × Offset Reinforcement × Characteristic Eligibility}
\newacronym{DQfD}{DQfD}{Deep Q-learning from Demonstrations}

\newacronym{bc}{BC}{Behavioural Cloning}

\newacronym{td}{TD}{Temporal Difference}
\newacronym{mc}{MC}{Monte Carlo}
\newacronym{mcts}{MCTS}{Monte Carlo Tree Search}
\newacronym{dp}{DP}{Dynamic Programming}
\newacronym{mpc}{MPC}{Model Predictive Control}

\newacronym{mdp}{MDP}{Markov Decision Process}
\newacronym{pomdp}{POMDP}{Partially Observable Markov Decision Process}
\newacronym{gae}{GAE}{Generalised Advantage Estimation}
\newacronym{gpi}{GPI}{Generalised Policy Iteration}

\newacronym{sgd}{SGD}{Stochastic Gradient Descent}
\newacronym{fps}{FPS}{Frames Per Second}

\newacronym{cdt}{CDT}{Centre for Doctoral Research}

\title{Deep Reinforcement Learning per la Manipolazione con Fisica di Oggetti in Ambienti Ingombri Bidimensionali}
\author{Luca Renna}
\advisor{Prof. Francesco Morandin}
\college{Dipartimento di Scienze Matematiche, Fisiche e Informatiche}
\degree{Corso di Laurea Triennale in Informatica}
\degreeyears{2022--2023}

\newcommand{\subTitle}{Deep Reinforcement Learning for 2D Physics-Based Object Manipulation in Clutter} 

\newcommand{\coadvisor}{Prof. Matteo Leonetti} 

\IfFileExists{\jobname.run.xml} 
{
}
{
}

\theoremstyle{definition}
\newtheorem{definition}{Definition}[section] 
\theoremstyle{plain}
\newtheorem{theorem}{Theorem}[section]

\begin{document}

\maketitle


\frontmatter
\pagestyle{plain}
\pagenumbering{roman}

\selectlanguage{italian}
\section*{Sommario}
Il \acrfull{drl} è un ambito di ricerca in rapida evoluzione con radici nella ricerca operativa e nella psicologia comportamentale, e con potenziali applicazioni che si estendono in vari domini, compreso quello della robotica. Questa tesi delinea le basi per il \acrfull{rl} moderno, partendo dall'impostazione matematica data dai Markov decision processes, la Markov property, goals e rewards, interazioni agente-ambiente e policies. Spieghiamo i principali tipi di algoritmi comunemente utilizzati nell'\acrshort{rl}, inclusi i metodi value-based, policy gradient, e actor-critic, con particolare attenzione agli algoritmi \acrshort{dqn}, \acrshort{a2c} e \acrshort{ppo}.
Successivamente, forniamo una breve rassegna della letteratura corrente su alcuni dei framework di uso comune per lo sviluppo di algoritmi e ambienti di \acrshort{rl}. In seguito, presentiamo il \acrfull{pbge}, un simulatore virtuale basato sul motore fisico Pymunk, che abbiamo sviluppato per la analisi della manipolazione bidimensionale di oggetti con vista dall'alto. La sezione della metodologia spiega come la nostra interazione agente-ambiente si possa inquadrare come un Markov decision process, in modo tale da poter applicare i nostri algoritmi di \acrshort{rl}. Dopodichè elenchiamo varie strategie di formulazione dei goal, incluse il reward shaping e il curriculum learning. Impieghiamo anche numerosi passaggi di pre-processing delle osservazioni per ridurre il carico computazionale necessario.
Nella fase sperimentale, affrontiamo una serie di scenari di difficoltà crescente. Iniziamo con uno scenario statico semplice, per poi aumentare gradualmente il livello di stocasticità. Ogni volta che gli agenti mostrano difficoltà nell'apprendimento, controbilanciamo aumentando il grado di reward shaping e curriculum learning. Questi esperimenti mostrano le forti limitazioni e le insidie degli algoritmi model-free in ambienti dinamici.
In conclusione, presentiamo un riepilogo dei risultati accompagnati da alcune osservazioni. Quindi delineiamo alcuni possibili sviluppi futuri per migliorare sia la nostra metodologia sia per un possibile ampliamento a sistemi in ambiente reale.

\newpage
\selectlanguage{english}
\section*{Abstract}
\acrfull{drl} is a quickly evolving research field rooted in operations research and behavioural psychology, with potential applications extending across various domains, including robotics. This thesis delineates the background of modern \acrfull{rl}, starting with the framework constituted by the Markov decision processes, Markov properties, goals and rewards, agent-environment interactions, and policies. We explain the main types of algorithms commonly used in \acrshort{rl}, including value-based, policy gradient, and actor-critic methods, with a special emphasis on \acrshort{dqn}, \acrshort{a2c} and \acrshort{ppo}.
We then give a short literature review on some widely adopted frameworks for implementing \acrshort{rl} algorithms and environments. Subsequently, we present \acrfull{pbge}, a virtual simulator based on the Pymunk physics engine we developed to analyse top-down bidimensional object manipulation. The methodology section frames our agent-environment interaction as a Markov decision process, such that we can apply our \acrshort{rl} algorithms. We list various goal formulation strategies, including reward shaping and curriculum learning. We also employ different steps of observation preprocessing to reduce the computational workload required.
In the experimental phase, we run through a series of scenarios of increasing difficulty. We start with a simple static scenario and then gradually increase the amount of stochasticity. Whenever the agents show difficulty in learning, we counteract by increasing the degree of reward shaping and curriculum learning. These experiments demonstrate the substantial limitations and pitfalls of model-free algorithms under changing dynamics.
In conclusion, we present a summary of our findings and remarks. We then outline potential future work to improve our methodology and possibly expand to real-world systems.
\tableofcontents
\chapter*{Acknowledgements}

I wish to extend my heartfelt gratitude to my supervisor, Professor Matteo Leonetti, for his guidance and support throughout this project. The opportunity to work at the King's College London Centre for Doctoral Research (CDT) has been invaluable, and his passion, expertise, and constructive feedback have inspired me to continue my studies in this field. Additionally, I am grateful to Professor Francesco Morandin, my home supervisor at the University of Parma, for his invaluable insight and contribution.
My appreciation extends to all the members of the CDT. Their assistance and camaraderie made my journey enjoyable and enriching. In particular, I would like to thank Dr. Gabriele La Malfa, whose words of encouragement and insightful discussions were a constant source of motivation, and Dr. Nathan Schneider Gavenski for his invaluable advice regarding my experiments and methodology.
I express my gratitude to the Computer Science and Erasmus Department staff at the University of Parma, especially to Professor Roberto Bagnara and Dr. Alessandro Bernazzoli for making the exchange a reality.
On a personal note, I am eternally grateful to my friends and family. A special thank you to my parents, Claudia and Antonio, whose unwavering encouragement and support have been my pillars of strength.
Lastly, I wish to acknowledge the immense contribution of the developers and contributors behind the open-source tools and libraries utilized in my thesis. Without their dedication and expertise, this project would not have come to fruition.
Please note that this research benefits from the High-Performance Computing facility of the University of Parma, Italy.

\mainmatter
\pagenumbering{arabic}

%
\pagestyle{fancy}
\chapter{Introduction}\label{chapter:introduction}

\gls{rl} is an approach deeply inspired by how humans acquire knowledge. We engage with our environment, assess the outcomes of our actions, and learn from the feedback we receive. This learning method allows us to adapt our behaviours to attain specific objectives, whether playing a musical instrument, learning to drive, playing a game like chess, or even acquiring a new language.

In Computer Science, \gls{rl} emerges as a computational paradigm emphasising learning through interaction. Its primary goal is to maximise the expected return of a numerical reward signal, positioning itself as a contrast to the traditional \gls{sl} methods. Unlike \gls{sl}, which necessitates labelled datasets comprising observations and corresponding ground truths, \gls{rl} does not rely on such explicit supervision. This distinction is pivotal, as constructing large labelled datasets can be labour-intensive, inefficient, and challenging, especially when it is difficult to discern a definitive ground truth.

\gls{rl} methods dispense with the need for ground truths, empowering models to learn through self-play, trial-and-error, and direct interaction with their environment. Data is procured autonomously, with outcomes, such as success or failure, attributed to this data post-collection.

\gls{dl} offers a solution to navigate the unpredictability of the real world. By training a highly parametrised model, like a deep neural network, \gls{drl} can efficiently handle vast state spaces where traditional \gls{rl} methods fall short. A typical configuration includes a convolutional layer followed by a specialised, fully connected layer, where the entire architecture is trained end-to-end. In this setup, the convolutional layer can be perceived as a "visual cortex" that autonomously sifts through observations to identify pertinent features. These features are subsequently evaluated by the fully connected layer, functioning akin to a "motor cortex" to determine the most appropriate action for the agent.

While \gls{rl} lays the mathematical and algorithmic groundwork, deep models supplement it with powerful representations. Together, they enable scaling to real-world systems.

\section{Brief History of Reinforcement Learning}\label{sec:brief_history}

\begin{figure}[h]
    \centering
    \begin{minipage}[t]{.49\textwidth}
        \centering
        \includegraphics[height=.8\linewidth]{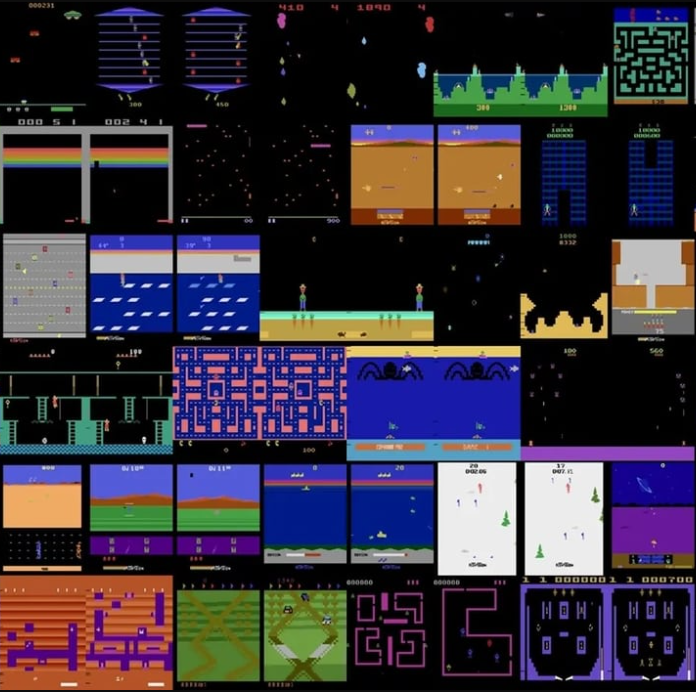}
        \captionof{figure}[Frames of different Atari games]{Frames of different Atari games. Source: \textcite{deepmind_agent57_2020}.}
        \label{fig:atari}
    \end{minipage}%
    \hfill
    \begin{minipage}[t]{.49\textwidth}
        \centering
        \includegraphics[height=.8\linewidth]{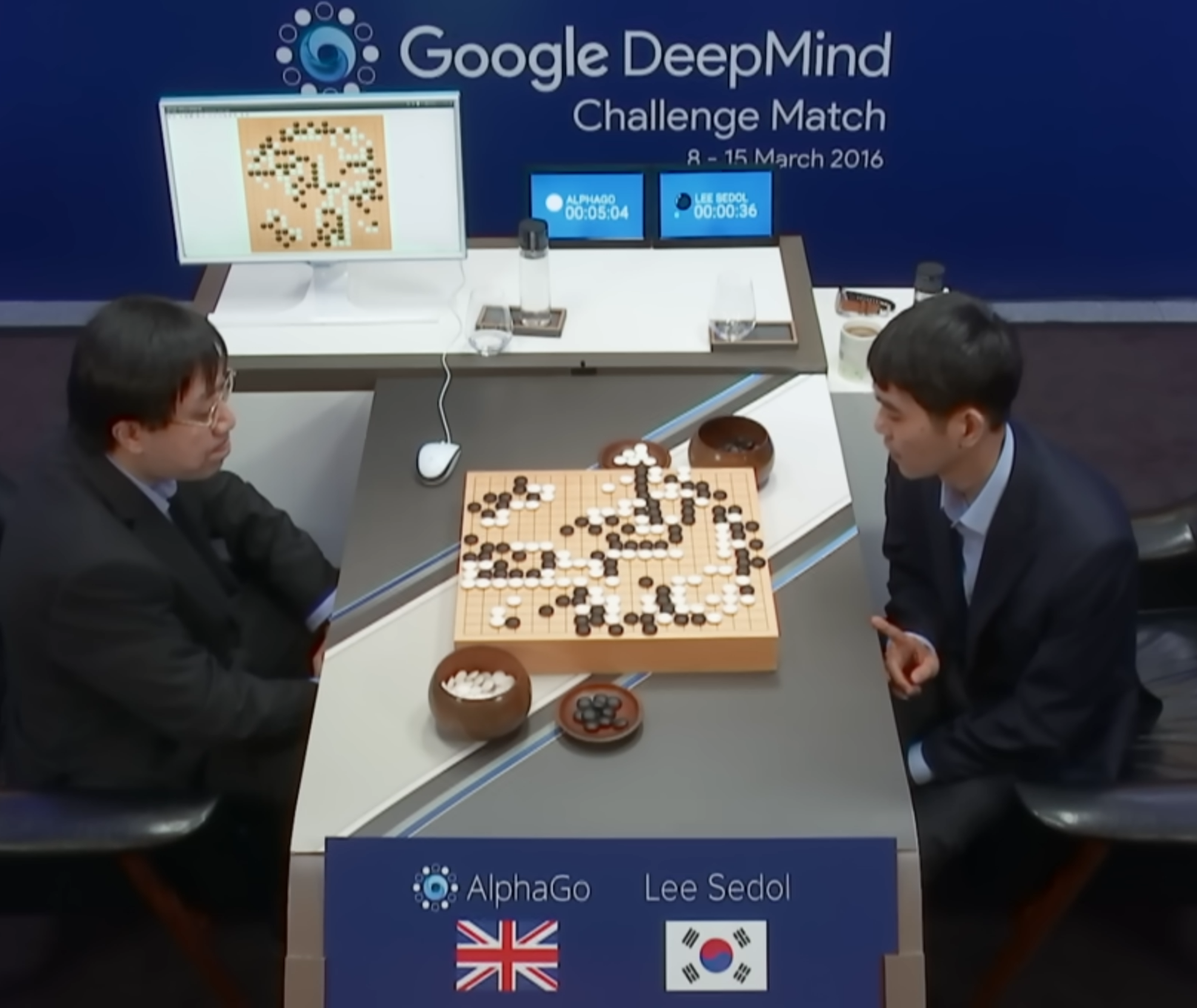}
        \captionof{figure}[International champion plays against AlphaGo]{International go champion Lee Sedol plays against AlphaGo in a five-game match. Source: \textcite{deepmind_alphago_2020}.}
        \label{fig:image2}
    \end{minipage}
\end{figure}

As recognised today, modern \gls{rl} emerged in the late 1980s. It synthesised principles from optimal control theory, trial-and-error learning, temporal-difference techniques, and other fields such as neuroscience, animal psychology and operation research. Most of its mathematical foundation, including notations, draws from optimal control and dynamic programming, notably, the Bellman Equations \autocite{sutton_reinforcement_2018}.

Recent advancements highlight the potent combination of Reinforcement Learning and Deep Learning, offering vast applications across various domains. Traditional tabular methods grapple with the curse of dimensionality in expansive state spaces, leading to storage challenges and extensive training requirements. To address these constraints, function approximation methods have been introduced. These methods aim to generalise across similar states. However, linear function approximators often fall short in complex tasks, where determining a relevant subset of features is non-trivial \autocite{sutton_reinforcement_2018, russell_artificial_2020}.

Nonlinear function approximators, such as \glspl{aan}, offer a solution. These networks are designed to automatically discern and craft features suitable for specific tasks, reducing the reliance on manual feature engineering. A significant advancement in this area has been the integration of Deep \glspl{cnn}. These networks have showcased their capability in end-to-end training, using raw images as state representations, leading to notable breakthroughs in the field \autocite{sutton_reinforcement_2018, russell_artificial_2020}.
Analysis of convergence properties of reinforcement learning algorithms using function approximation is complicated. Results for TD learning have been progressively strengthened for the case of linear function approximators, but several examples of divergence have been presented for nonlinear functions \autocite{tsitsiklis_analysis_1997, mnih_playing_2013}.

Recent successes in the field of \gls{drl} demonstrate the potential of these techniques. Despite the existing gap between theoretical guarantees and empirical performance, real-world applications of \gls{drl} have repeatedly yielded results surpassing human expertise. An early testament to this was Tesauro's TD-Gammon 0.0 \autocite{tesauro_td-gammon_1994, tesauro_temporal_1995}, a backgammon-playing program that used a multi-layer \gls{aan} for \gls{td}(lambda) function approximation \autocite{sutton_learning_1988}. Starting only with a raw representation of the board state and minimal game rules, this program reached master-level performance through self-play. However, while later versions of TD-Gammon improved play using handcrafted features, initial efforts to replicate this success in more complex games like chess, Go, and checkers fell short, leading to the widespread belief that TD-Gammon's nonlinear approximation strategy might be uniquely suited for backgammon \autocite{pollack_why_1996}.

This scepticism was soon dispelled by a team of researchers at DeepMind. Through their research, they demonstrated that deep convolutional \glspl{aan} could be seamlessly integrated into \gls{rl} to autonomously design features from raw visual inputs of the environment state \autocite{mnih_playing_2013, mnih_human-level_2015}. Their pioneering agent, using a \gls{dqn}, achieved a high level of play across a diverse set of 49 Atari games within the \gls{ale} \autocite{bellemare_arcade_2013}, all without any prior knowledge of game rules. Remarkably, a uniform network architecture, algorithm, and set of hyperparameters were employed across these games. This accomplishment showcased unprecedented generalisation abilities and marked a significant departure from the prevailing reliance on meticulously handcrafted features in \gls{drl}.

Building on their earlier achievements in \gls{drl}, DeepMind unveiled their groundbreaking AlphaGo program \autocite{silver_mastering_2016}. This software not only managed to defeat world champions at Go—a feat once considered years, if not decades, away from realisation—but also illustrated once again the transformative potential of merging \gls{dl} and \gls{rl}. AlphaGo was trained using \glspl{dnn} with a combination of initial \gls{sl} on a large dataset of expert human moves, followed by \gls{rl} via self-play on simulations of Go. Central to its design was a modified \gls{mcts}, guided by both policy and value functions, with \glspl{cnn} employed for function approximation. One year later, the next iteration of the program, AlphaGo Zero \autocite{silver_mastering_2017}, outperformed its predecessor. Impressively, it attained this without any human-derived data or input, relying solely on the game's basic rules. Its architecture was not only simpler by doing away with \gls{sl} and using a quicker \gls{mcts} but also more potent, achieving superhuman performance exclusively through self-play. The evolution continued with AlphaZero, which generalised the original program to master Chess and Shogi \autocite{silver_general_2018}. Finally, the advent of MuZero \autocite{schrittwieser_mastering_2020} marked a paradigm shift. Unlike its precursors, which needed to be provided with predefined models of their respective games, MuZero needed no prior knowledge and used \gls{drl} to learn and refine its internal model to simulate the environment dynamics, which would query to plan ahead.

The innovative streak of the AlphaGo series did not limit itself to games but underscored the versatility of \gls{drl} in various other scientific domains. AlphaFold, for instance, addressed a long-standing challenge in modern biology: it efficiently predicted the 3D structures of nearly all known proteins, marking a significant milestone \autocite{senior_improved_2020, jumper_highly_2021}. AlphaTensor achieved a breakthrough in computational mathematics by devising a 4x4 matrix multiplication technique that surpassed Strassen's algorithm—a standard that had held its ground for over half a century \autocite{fawzi_discovering_2022}. Furthermore, the foundational principles of the AlphaZero algorithm found utility in quantum computing, demonstrating its adaptability to diverse challenges \autocite{dalgaard_global_2020}.

Recently, there has been a noticeable shift towards embracing simple yet highly scalable \gls{drl} algorithms and architectures, particularly those that can harness enormous parallel computing power \autocite{schulman_proximal_2017}. A prominent instance of highly distributed \gls{rl} is seen with OpenAI's OpenAiFive. \textcite{openai_dota_2019} explains how they exploited the scalability of \gls{ppo} by training agents on an unprecedented scale using up to 1536 GPUs over a span of 10 months. Similarly, the GPT3-4 series of transformer models represents another remarkable application. These models were also trained using large-scale deployments of \gls{ppo} on \gls{dnn} followed by Reinforcement Learning from Human Feedback during the finetuning process, showcasing the versatility and strength of these approaches \autocite{brown_language_2020, openai_gpt-4_2023}.

\section{Motivation and Objectives}\label{sec:motivation}

The motivation for this thesis is threefold. First and foremost, it is to provide a general introduction to \gls{rl}, structured so that the reader should not need any preliminary knowledge of the field. We first give the definitions and background information of traditional \gls{rl} and describe some classic methods and algorithms. We then expand those methods to the function approximation case, introducing new concepts until the reader has all the necessary knowledge to understand a selection of state-of-the-art algorithms.
Second, we want to provide a standardised environment that can be used with those algorithms. One of the critiques against research in the Deep Reinforcement Learning field is the lack of standardisation. Significant efforts have recently been made to standardise algorithm implementations and agent environments. Here, we follow the Gymnasium APIs \autocite{towers_gymnasium_2023} and their best practices to make our environment fully standardised. We follow the Stable Baselines 3 framework \autocite{raffin_stable-baselines3_2021} for training our agents, and the seeds and hyperparameters are included here for reproducibility.
Last, we wish to benchmark and explore some widely adopted model-free \gls{drl} algorithms in our environment. Research shows that kinodynamic planning usually works well in physics-based environments. A key issue with some of these classic optimal control open-loop methods is that the computed solution is static and does not adapt to unexpected changes. This poses problems in highly dynamic environments and can be further compounded by the inaccuracies in the physics model used to compute these solutions. The planning is often onerous, so we cannot change it at execution time if we want to remain within the constraints of a real-time system. In our experiments, we use \gls{rl} methods, which are augmented by \gls{dl} function approximation to make them suited to learn real-time interaction in systems with large state spaces. Moreover, we also use a raw image representation of the environment to allow a potentially arbitrary number of obstacles to be in the environment without (in theory) needing retraining, provided that the model generalisation capabilities are good enough.

\section{Scope and Limitations}\label{sec:scope_and_limitations}

The scope of this thesis is limited to model-free methods for two primary reasons. First, including model-based approaches would significantly expand the breadth of the thesis, diverging from its introductory purpose. Second, model-based methods are known to be particularly challenging to implement, which would detract from this work's goal of serving as an approachable introduction to \gls{rl} rather than a display of more advanced concepts.

Computational restrictions also bound our experiments. While the studies mentioned in \cref{sec:brief_history} devoted thousands of hours to training their models, we have constrained our training time to a maximum of one day due to sharing computational resources with other groups. Allowing for a longer training duration might have led to better results.

Furthermore, this project remains within the bounds of a simulated environment, sidestepping the exploration of the "sim-to-real gap." Although a potential goal would be to use the robot gripper model trained in a simulation for a real-world application, we have yet to undertake any real-world testing or preliminary studies on transferring our models from the simulator to a real environment.

\section{Thesis Outline}\label{sec:outline}

In \vref{chapter:background}, we give the mathematical notation and foundational knowledge to understand the \gls{rl} concepts used in subsequent chapters. The \vref{chapter:algos} provides an overview of both value-based and policy-gradient \gls{rl} methods, starting with the basic tabular algorithms and leading up to the function approximation case. Here, we give the detailed description of \gls{dqn}, \gls{a2c} and \gls{ppo}, which we will then use in our experiments. In \vref{chapter:background}, we give an overview of other standardised environments and frameworks in \gls{drl}, along with relevant literature on physics-based object manipulation. The \vref{chapter:methodology} describes the \gls{pbge} environment along with preprocessing steps, our goal formulation and the agent network architecture. In \vref{chapter:experiments}, we present a series of experiments we run on our environment and their results. In the final \vref{chapter:conclusion}, we conclude with an analysis of our results, and we discuss possible directions for future work.

\chapter{Background} \label{chapter:background}

This chapter provides a summary of the background and the relevant theoretical foundations of \gls{rl} utilised in further chapters of this thesis. Initially, we delineate the general framework of \gls{rl}, introducing the fundamental concept of \acrfull{mdp}, which is central to defining interactions in terms of states, actions, and rewards. Subsequently, we articulate goals, rewards, returns, and the differentiation between episodic and continuing tasks. Moreover, we offer a brief overview of the various types of environments encountered in \gls{rl} and emphasise the significance of policies. The following sections detail the Bellman Equations and Bellman Optimality Equations, utilising their self-consistency property to introduce value functions. Following this, we illustrate the relationship between optimal policies and optimal value function, and we end the section by explaining the central trade-off between exploration and exploitation. We conclude the chapter by distinguishing between model-free and model-based \gls{rl} approaches, highlighting their respective strengths and weaknesses. The bulk of the definitions and explanations are taken from \textcite{sutton_reinforcement_2018}, \textcite{francois-lavet_introduction_2018}, \textcite{SpinningUp2018}, \textcite{silver_advanced_2015}, which the author recommends as additional resources. The thesis does not delve into \gls{dl} and \glspl{aan}, which we use here for nonlinear function approximators. Readers interested in \gls{dl} should consult additional resources such as \textcite{lecun_deep_2015} and \textcite{Goodfellow-et-al-2016}. The notation used throughout the text is taken from \textcite{sutton_reinforcement_2018} and is made available as \vref{appendix:notation}.

\section{Reinforcement Learning Framework}
\gls{rl} is a computational approach to automating and comprehending decision-making and goal-oriented learning. Its uniqueness stems from its reliance on direct environment-agent interactions for learning, sidestepping the need for model completeness or explicit supervision. \gls{rl} leverages \glspl{mdp} to define interactions between the agent and its environment, encompassing states, actions, and rewards. This model provides a concise representation of the Artificial Intelligence problem, including cause-effect relationships, inherent uncertainty, and the presence of an explicit goal. Value functions are fundamental to many \gls{rl} methods examined in this thesis. Incorporating value functions differentiates \gls{rl} from evolutionary methods that search the policy space guided by holistic policy evaluations. Another key aspect that makes \gls{rl} stand out from other machine learning paradigms is that the agent's actions affect subsequent data it receives, making the training datasets time-dependent and based on sequential, non i.i.d data. The decisions also unfold over time, making the rewards delayed.

We use \glspl{mdp} to frame the problem of learning from interaction to achieve a goal. In this setup, the "learner" or "decision-maker" is known as the agent. Everything that the agent interacts with outside of itself is called the environment.
The dividing line between the agent and its environment is not always a physical barrier, like the body of a robot or an animal. The rule of thumb we use is that anything the agent cannot change at will is considered part of the environment, regardless of whether the agent has partial or absolute knowledge of it. Thus, the agent-environment boundary marks the edge of the agent's control, not what it knows.
In this ongoing process, the agent chooses actions, and the environment reacts, creating new situations for the agent. The environment also offers rewards—specific numerical values—that the agent tries to increase over time through its chosen actions.

\subsection{Markov Decision Process}
\glspl{mdp} are a classic formalisation of sequential decision making, where actions influence immediate rewards, subsequent situations, or states, and through those future rewards. Thus, \glspl{mdp} involve delayed reward and the need to trade off immediate and delayed reward. We model the environment in which the agent acts as an \gls{mdp}:
\begin{definition}[Markov Decision Process]
A \textit{\acrfull{mdp}} is defined by a 4-tuple $(S, A, p, r)$ where:
\begin{itemize}
    \item $\S$ is a finite set of states.
    \item $\A$ is a finite set of actions.
    \item $p: \S \times \R \times \S \times \A \rightarrow [0, 1]$ defines the state transition probability such that $\p(s',r|s,a)$ is the probability of transitioning to state $s'$ from state $s$ with reward $r$ by taking action $a$.
    \item $r: \S \times \A \times \S \rightarrow \mathbb{R}$ is the reward function where $r(s,a,s')$ gives the immediate reward for taking action $a$ from state $s$ and transitioning to state $s'$.
\end{itemize}
\end{definition}
In a finite \gls{mdp}, the set of states, actions, and rewards $(\S,\A,$ and $\R)$ all have a finite number of elements. In this case, the random variables $R_t$ and $S_t$ have well-defined discrete probability distributions dependent only on the preceding state and action. That is, for particular values of these random variables, $s'\in\S$ and $r\in\R$, there is a probability of those values occurring at time $t$, given particular values of the preceding state and action:
\begin{equation}
\p(s',r|s,a) \defeq \Pr{S_t=s',R_t=r \mid S_{t-1}=s,A_{t-1}=a},
\end{equation}
for all $s',s\in\S,r\in\R$ and $a\in\A(s).$ The function $p$ defines the dynamics of the \gls{mdp}. The dynamics function $p$ is an ordinary deterministic function of four arguments which specifies a probability distribution for each choice of $s$ and $a$, that is, that
\[ \sum_{s'\in \S} \sum_{r \in \R} \p(s',r|s,a) = 1, \text{ for all } s\in\S,a\in\A(s). \]
Alternatively, we can also define $r(s,a)$ as a two-argument function that gives the expected rewards for a state-action pair:
\begin{equation}
    r(s,a) \defeq \E{R_t\mid S_{t-1}=s,A_{t-1}=a} = \sum_{r\in\R}r\sum_{s'\in \S} \p(s',r|s,a).
\end{equation}
From now on, we assume that the environment is an \gls{mdp} and the symbols $s, s' \in \S$, $a \in \A(s)$, $r \in \R$ and $p(s',r|s,a)$ gives the environment dynamics.

\subsection{Markov Property}
\begin{definition}[Markov Property] \label{def:markov_prop}
    A state $S_t$ is \textit{Markov} if and only if
    \[ \Pr{S_{t+1} \mid S_t=s} = \Pr{S_{t+1} \mid S_0=s_0,\dots,S_t=s}. \]
\end{definition}
A key property of an \gls{mdp} is that transitions only depend on the most recent state and action and no prior history. That is, the probability of each possible value for $S_t$ and $R_t$ depends on the immediately preceding state $S_{t-1}$ and action $A_{t-1}$ and, given them, not on all earlier states and actions. The state must include information about all aspects of the past agent–environment interactions that make a difference in the future. If it does, the state is said to have the \textit{Markov property}.

\subsection{Goals and Rewards} \label{sec:goal_reward}
The reward signal defines the goal of a \gls{rl} problem. The environment sends the agent a number called the reward on each timestep. The agent's sole objective is to maximise the cumulative reward, the total future reward it receives over the long run. \gls{rl} is based on the \textit{reward hypothesis} in \cref{def:reward_hypo}, meaning that the maximisation of expected cumulative reward can describe all goals.
\begin{definition}[The Reward Hypothesis] \label{def:reward_hypo}
    All of what we mean by goals and purposes can be well thought of as the maximisation of the expected value of the cumulative sum of a received scalar signal (called reward) \autocite[53]{sutton_reinforcement_2018}.
\end{definition}

\subsection{Return}
As we said in \cref{sec:goal_reward}, the agent's goal in \gls{rl} is to maximise the cumulative reward in the long run, namely the \textit{expected return}. We define the return $G_t$ (also known as \textit{reward-to-go}) as some specific function of the reward sequence:
\begin{definition}[Return]\label{def:return}
    The return $G_t$ is the total discounted reward from timestep t.
    \[ G_t \defeq R_{t+1} + \gamma R_{t+2} + \gamma^2 R_{t+3} + \dots = \sum_{k=0}^{\infty} \gamma^kR_{t+k+1}, \]
    where $\gamma$ is a parameter $0 \leq \gamma \leq 1$, called the \textit{discount rate}.
\end{definition} 
Returns at successive timesteps are related to each other in a way that is important for the theory and algorithms of \gls{rl}:
\begin{align}
G_t & \defeq R_{t+1} + \gamma R_{t+2} + \gamma^2 R_{t+3} + \gamma^3 R_{t+4} + \dots \nonumber\\
& = R_{t+1} + \gamma G_{t+1}, \label{eq:recursive_return}
\end{align}
To ensure the infinite sum in \cref{def:return} yields a finite value, we permit an undiscounted return (where $\gamma = 1$) exclusively in tasks where all sequences terminate.
We use discounting on the return for several rationales, including making returns more mathematically convenient, avoiding infinite returns in cyclic Markov processes, accounting for future uncertainties and regulating the emphasis between immediate and delayed rewards. We can adjust $\gamma$ to place more or less value on immediate rewards above delayed rewards, where $\gamma$ close to $0$ leads to a "myopic" evaluation and $\gamma$ close to $1$ leads to a "far-sighted" evaluation.

\subsection{Episodic and Continuing tasks} \label{sec:episodic_cont_task}
When the agent-environment naturally breaks into distinct subsequences, we refer to the task at hand as an \textit{episodic task}. Each episode ends in a special state called the \textit{terminal state}, followed by a reset to a standard starting state or a sample from a standard distribution of starting states. Every episode begins independently of how the previous one ended.
Conversely, a \textit{continuing task} is characterised by the agent-environment interaction going on continually without limit. The return formulation shown in \cref{def:return} holds for both episodic and continuing tasks.
We will use notation and algorithms assuming an episodic task in this and subsequent chapters. However, most concepts can be easily generalised to accommodate continuing tasks.

\subsection{Agent and Environment Interaction} \label{sec:agent_env_inter}
\begin{figure}[ht]
\centering
\includegraphics[width=0.9\textwidth]{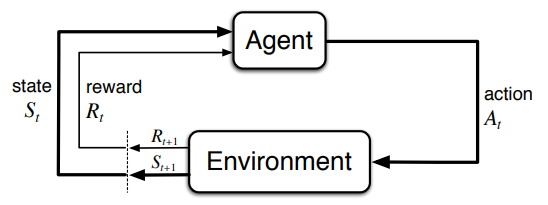}
\captionof{figure}[The agent-environment interaction in an \gls{mdp}.]{The agent-environment interaction in a Markov Decision Process. Source: \textcite[48]{sutton_reinforcement_2018}.}
\label{fig:rl_process}
\end{figure}
At each timestep, $t$, the agent receives some representation of the environment's state $S_t \in \S$ and, on that basis, selects an action, $A_t \in \A(s)$. One timestep later, the agent receives a numerical reward,$ R_{t+1} \in \R \subset \mathbb{R}$, and finds itself in a new state, $S_{t+1}$. The \gls{mdp} and agent together thereby give rise to a sequence or \textit{trajectory} $\tau$ that begins as follows:
\[ \tau = S_0,A_0,R_1,S_1,A_1,R_2,S_2,A_2,R_3,\dots \]
Where the very first state of the world $S_0$ is randomly sampled from the start-state distribution $\mu_0$, namely $S_0 \sim \mu_0(\cdot)$. We call a complete trajectory from a start state to a final state an \textit{episode} or \textit{rollout}.

\subsection{Full and Partial Observability} \label{sec:observability}
We defined $S_t$ as the environment's state representation, meaning whatever data the environment uses to pick the following observation and reward. However, in many cases, potentially essential aspects of the environment's state are not directly observable or may contain irrelevant information. In that case, the environment emits not its states but observations $O_t$ that provide only partial information about it. When the environment emits complete states $O_t = S_t$, we call the problem \textit{fully observable} and model it using an \gls{mdp}. Otherwise, it is \textit{partially observable}, and we use a Partially Observable Markov Decision Process. Here, we always assume the case where we have full observability, but parametrised function approximation in \vref{sec:value_approx} can be extended to partially observable environments with no changes. If a state variable is not observable, the parametrisation can be chosen so that the approximate value does not depend on that state variable.

\subsection{Action Spaces} \label{sec:action_space}
Different environments permit various types of actions, with the collective set of all legal actions in a given environment being known as the \textit{action space}. Environments such as Atari and Go possess \textit{discrete action spaces}, restricting the agent to a finite set of potential moves. In contrast, scenarios where, for example, the agent manages a robot in a physical realm have \textit{continuous action spaces}, where actions are articulated as real-valued vectors.
This differentiation between discrete and continuous action spaces is essential to the methods applied in \gls{drl}. Some families of algorithms can only be directly applied to a specific type of action space and would have to be substantially reworked for the other. In the scope of this thesis, the emphasis will be placed on environments characterised by discrete action spaces.

\subsection{Policies}
A policy is a rule used by an agent to decide what actions to take, determining the \textit{behaviour} of an agent at a given time. More formally, a policy is a mapping from states to probabilities of selecting each possible action. If the agent follows policy $\pi$ at time $t$, then $\pi(a|s)$ is the probability that $A_t=a$ if $S_t=s$.

\begin{definition}[Policy]
A \textit{policy} $\pi$ is a distribution of states over actions,
\[ \pi(a|s) \defeq \Pr{A_t=a \mid S_t = s} \]
and gives the probability of taking action $a$ when in state $s$ (also denoted by $a \sim \pi(\cdot|s)$)
\end{definition}
We denote the policy by $a = \pi(s)$ when the policy is deterministic. Otherwise, it is assumed to be stochastic. \gls{rl} methods specify how the agent's policy changes due to its experience.
In the context of \gls{drl}, it is standard to work with \textit{parametrised policies}, which are policies whose outputs are computable functions that depend on a set of parameters. These parameters can depend, for example, on a neural network's weights and biases, which can be fine-tuned to modify the agent's behaviour through some optimisation algorithm. Within \gls{drl}, the prevalent types of stochastic policies are categorical policies and diagonal Gaussian policies. Categorical policies can be used in discrete action spaces, while diagonal Gaussian policies are used in continuous action spaces. Two fundamental computations are key to the utilisation and training of stochastic policies: the sampling of actions from the policy and computing log-likelihoods of specific actions represented as $\log \pi_{\theta}(a|s)$. A more detailed discussion on parametrised policies will be presented in \vref{sec:policy_gradient}.

\section{Value Functions}
Value functions of states or state-action pairs are functions that serve to estimate the expected return for an agent when in a given state or state-action pair and operating under a specific policy forever after. Informally, a state-value function represents the total reward an agent expects to accumulate over time, starting from a particular state and following a policy.
Rewards indicate the immediate desirability of states, whereas values convey long-term desirability, factoring in the potential subsequent states and the associated rewards. In the decision-making process, the objective is to choose actions that lead to states with the highest value rather than the highest immediate reward. Central to nearly all \gls{rl} algorithms is a method for efficiently estimating values.
\begin{definition}[State-value function]
The \textit{state-value function} $\vpi(s)$ of a state $s$ under policy $\pi$ is the expected return when starting from state $s$ and then following $\pi$ thereafter.
\[ \vpi(s) \defeq \mathbb{E}_{\pi}\left[ G_t \mid S_t=s \right] \]
\end{definition}
\begin{definition}[Action-value function]
The \textit{action-value function} $\qpi(s,a)$ of taking action $a$ in a state $s$ under a policy $\pi$ is the expected return when starting from state $s$, taking action $a$, and thereafter following policy $\pi$.
\[ \qpi(s,a) \defeq \mathbb{E}_{\pi}\left[ G_t \mid S_t=s , A_t=a\right] \]
\end{definition}

\subsection{Bellman Equations}
All four value functions obey special self-consistency equations known as the Bellman equations and are a fundamental property of value function. These equations express a relationship between the state's value and its successor states' values.

The Bellman Equation for $\vpi$ is:
\begin{align}
\vpi(s) & \defeq \mathbb{E}_{\pi}\left[ G_t \mid S_t=s \right] \nonumber\\
& = \sum_a \pi(a|s) \sum_{s',r} \p(s',r|s,a) \left[ r + \gamma  \vpi(s') \right], \label{eq:bell_state}
\end{align}
and the Bellman Equation for $\qpi$ is given by:
\begin{align}
\qpi(s,a) & \defeq \mathbb{E}_{\pi}\left[ G_t \mid S_t=s , A_t=a\right] \nonumber\\
& = \sum_{s',r} \p(s',r|s,a) \left[ r + \gamma \sum_{a'} \pi(a'|s') \qpi(s',a') \right]. \label{eq:bell_action}
\end{align}

\subsection{Bellman Optimality Equations} \label{sec:bellmaneq}

The Bellman optimality equation is a system of equations, each corresponding to an individual state. Suppose the dynamics $\pi$ of the environment are known. In that case, one can theoretically solve this equation system for $\vstar$ using any one of a variety of methods for solving systems of nonlinear equations, such as \gls{dp}. 
Explicitly solving the equation provides one route to finding the optimal policy and solving the \gls{rl} problem. However, this solution is rarely useful as it is akin to an exhaustive search. This solution rests on three assumptions that are rarely true in practice: 
\begin{enumerate*}[label=(\arabic*)]
\item the dynamics of the environment are accurately known
\item computational resources are sufficient to complete the calculation and
\item the states have the Markov property.
\end{enumerate*}
In most tasks, a combination of these assumptions is violated. Hence, in \gls{rl}, one must settle for approximate solutions. Many \gls{rl} methods can be clearly understood as approximatively solving the Bellman optimality equation using actual sampled transitions instead of knowledge of expected transitions. This online nature of \gls{rl} makes it possible to approximate optimal policies in ways that put more effort into learning to make good decisions for frequently encountered states at the expense of infrequently encountered states.
\begin{figure}[ht]
\centering
\includegraphics[width=0.9\textwidth]{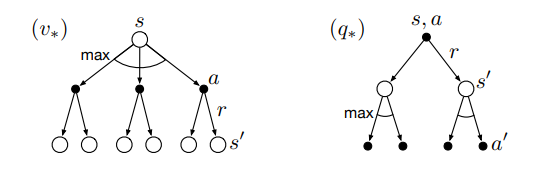}
\captionof{figure}[Backup diagrams for $\vstar$ and $\qstar$.]{Backup diagrams for $\vstar$ and $\qstar$. Source: \textcite[64]{sutton_reinforcement_2018}.}
\label{fig:bellman_opt}
\end{figure}
The Bellman Optimality Equation for $\vpi$ is
\begin{align}
\vstar(s) & = \max_{a \in \A(s)} q_{\pi_*}(s,a) \nonumber\\
& = \max_a \E{R_{t+1} + \gamma \vstar(S_{t+1}) \mid S_t=s, A_t = a} \nonumber\\
& = \max_a \sum_{s',r} \p(s',r|s,a) \left[r + \gamma \vstar(s') \right]. \label{eq:bellman_opt_vpi}
\end{align}
and the Bellman Optimality Equation for $\qpi$ is given by
\begin{align}
\qstar(s,a ) & = \E{R_{t+1} + \gamma \max_{a'} \qstar(S_{t+1}, a') \mid S_t=s, A_t = a} \nonumber\\
& = \sum_{s',r} \p(s',r|s,a) \left[r + \gamma \max_{a'} \qstar(s', a') \right]. \label{eq:bellman_opt_qpi}
\centering
\end{align}
The backup diagram on the left of \cref{fig:bellman_opt} graphically represents \cref{eq:bellman_opt_vpi}, and the backup diagram on the right of \cref{fig:bellman_opt} graphically illustrates \cref{eq:bellman_opt_qpi}.

\subsection{Optimal Value Functions and Optimal Policies} \label{sec:optimal_valuef_policies}
Solving a \gls{rl} task roughly means finding a policy that achieves a lot of reward over the long run. Value functions define a partial ordering over policies: a policy $\pi$ is defined to be better than or equal to a policy $\pi'$ if its expected return is greater or equal to that of $\pi'$ for all states, meaning $\pi \geq \pi'$ if and only if $\vpi(s) \geq v_{\pi'}(s)$ for all $v \in \S$.
At least one policy is always better than or equal to all other policies; we call it the \textit{optimal policy}. Although there may be more than one, we denote all the optimal policies by $\pi_*$. They all share the same state-value function, called the \textit{optimal state-value function}.
\begin{definition}[Optimal state-value function]
    The \textit{optimal state-value function} $\vstar(s)$ is the maximum state-value function over all policies.
\[ \vstar(s) \defeq \max_\pi \vpi(s), \text{ for all } s \in \S. \]
\end{definition}
\begin{definition}[Optimal action-value function]
    The \textit{optimal action-value function} $\qstar(s,a)$ is the maximum action-value over all policies.
\[ \qstar(s,a) \defeq \max_\pi \qpi(s,a), \text{ for all } s \in \S \text{ and } a \in \A(s). \]
\end{definition}
\begin{theorem}[Optimal policies]
For any Markov Decision Process:
\begin{itemize}
    \item There exists an optimal policy $\pi_*$ that is better than or equal to all other policies, $\pi_* \geq \pi, \forall \pi$
    \item There is always a deterministic optimal policy
    \item All optimal policies achieve the optimal value function $v_{\pi_*}(s) = \vpi(s)$
    \item All optimal policies achieve the optimal action-value function $q_{\pi_*}(s,a) = \qstar(s,a)$
\end{itemize}
\end{theorem}
Any policy that is greedy with respect to the optimal evaluation function $\vstar$ is an optimal policy; this implies that a one-step-ahead search using $\vstar$ yields long-term optimal actions. Similarly, by utilising $\qstar$, the agent can bypass the need for this one-step look ahead; it only needs to choose any action that maximises $\qstar(s,a)$ without knowledge about the environment's dynamics. Hence, an optimal policy can be found by maximising over $\qstar(s,a)$,
\begin{equation} \label{eq:qstar_yields_optimalp}
\pi_*(a|s) = 
\begin{cases}
1 & \text{if } a = \arg\max_{a\in \A(s)} \qstar(s,a) \\
0 & \text{otherwise}
\end{cases}
\end{equation}

\subsection{Advantage Function}
Sometimes, in \gls{rl}, we do not need to describe how good an action is in an absolute sense, but only how much better it is than others on average. That is to say, we want to know the relative advantage of that action. We make this concept precise with the \textit{advantage function}.
\begin{definition}[Advantage function] \label{def:advantage}
The advantage function $A_\pi(s,a)$ corresponding to a policy $\pi$ describes how much better it is to take a specific action $a$ in state $s$ over randomly selecting an action according to $\pi(\cdot|s)$, assuming you act according to $\pi$ forever after.
\[ \api(s,a) = \qpi(s,a) - \vpi(s). \]
\end{definition}

\subsection{Exploration and Exploitation Trade-off} \label{sec:exploration}
A central theme of \gls{rl} is the balance between exploitation — meaning leveraging past experiences to obtain reward — and exploration, which we use to discover better action selections in the future. The agent must, therefore, balance the two to complete the task by trying a variety of actions and then progressively favouring those that appear to be best. The exploration-exploitation dilemma has been intensively studied but remains unsolved \autocite{sutton_reinforcement_2018}.

\section{Model-Free and Model-Based}
\begin{figure}[ht]
    \centering
    \includegraphics[width=0.95\textwidth]{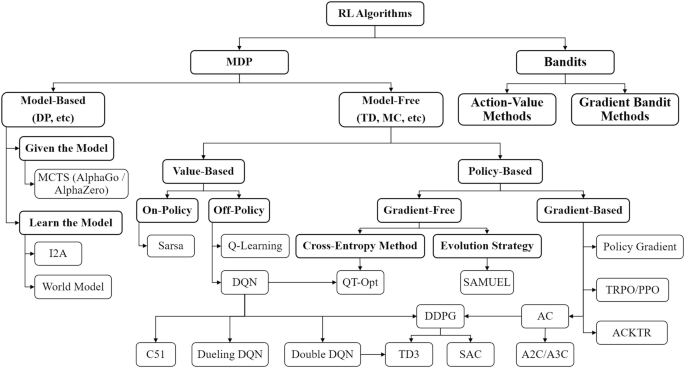}
    \caption[Taxonomy of \gls{rl} algorithms.]{A taxonomy of Reinforcement Learning methods and algorithms. Source: \textcite{zhang_taxonomy_2020}.}
    \label{fig:taxonomy}
\end{figure}
One of the most crucial branching points in an RL algorithm is whether the agent has access to a model of the environment. By a model of the environment, we mean a function which predicts state transitions and rewards. The model may be either given a priori or learned through experience.
The main upside to having a model is that it allows the agent to plan by thinking ahead, seeing what would happen for a range of possible choices, and explicitly deciding between its options. Agents can then distil the results from planning ahead into a learned policy. A particularly famous example of this approach is \textcite{silver_mastering_2017}, where AlphaGo Zero is given an exact model of the game of Go. When this works, it can substantially improve sample efficiency over methods that do not have a model.
The main downside is that a ground-truth model of the environment is usually not available. If an agent wants to use a model in this case, it has to learn it purely from experience, which creates several challenges. MuZero gives an example of successful model-learning in \textcite{silver_general_2018}. However, the biggest challenge is that bias in the model can be exploited by the agent, resulting in an agent which performs well with respect to the learned model but behaves sub-optimally or terribly in the real environment. Model learning is fundamentally hard, so even considerable investments of time and computational resources can fail to pay off. 
Algorithms which use a model are called \textit{model-based methods}, and those that do not are called \textit{model-free methods}. A potential classification of \gls{rl} algorithms is given in \cref{fig:taxonomy}. While model-free methods forego the potential gains in sample efficiency from using a model, they tend to be easier to implement and tune. This thesis predominantly explores model-free methods, focusing on their practicality in \gls{rl}.

\chapter{Reinforcement Learning Algorithms} \label{chapter:algos}

This chapter gives an overview of the \gls{rl} algorithms, starting with the basic tabular methods and leading up to the function approximation case. We begin by explaining value-based methods, including the central concepts of prediction, control and policy iteration. We show and explain the classic off-policy Q-learning algorithm. Later, we generalise value-based methods to the function approximation case using semi-gradient methods. We show the extension of tabular Q-learning to the famous \gls{dqn}. We then introduce policy gradient methods using REINFORCE as an example, which we modify by adding a baseline to move closer to actor-critic methods. We then show and detail the \gls{a2c} algorithm, which combines policy gradient and value-based methods. We conclude by showing \gls{ppo}, a state-of-the-art actor-critic algorithm that improves upon \gls{a2c}. 

There are two main approaches to representing and training agents, \textit{Policy Optimisation methods} and \textit{Value-based methods} (sometimes referred to as \textit{Q-Learning} methods). Methods in Policy Optimisation represent a policy explicitly as $\pi(a|s,\th)$. They optimise the parameters $\th$ directly by gradient ascent on the performance objective $J(\th)$ or indirectly by maximising local approximations of $J(\th)$. This optimisation is almost always performed on-policy, which means that each update only uses data collected while acting according to the most recent version of the policy. Policy optimisation also usually involves learning an approximator $\hat v(s,\w)$ for the on-policy value function $\vpi(s)$, which gets used to help with policy updates. Methods such as \gls{a2c} and 1\gls{ppo} fall under this category.
Value-based approaches predominantly focus on learning an approximator $\hat q(s,a,\w)$ for the optimal action-value function, $\qstar(s,a)$. The objective function used is typically derived from the Bellman equation. Optimisation under this method is often conducted off-policy, meaning that each update can use data collected at any point during training, regardless of how the agent chose to explore the environment when the data was obtained. The resultant policy is obtained through the relationship between $\qstar$ and $\pi_*$, as shown in \vref{eq:qstar_yields_optimalp}. \gls{dqn} is a notable example of value-based methods.
The primary strength of policy optimisation methods is their principled nature, in the sense that they directly optimise for the desired outcome. This tends to make them stable and reliable. By contrast, value-based methods only indirectly optimise agent performance by training $\hat q$ to satisfy a self-consistency equation. This kind of learning has many failure modes, so it tends to be less stable. However, value-based methods gain the advantage of being substantially more sample efficient when they do work because they can reuse data more effectively than policy optimisation techniques \autocite{SpinningUp2018}.

\section{Value-based Methods}
In value-based methods, we try to learn an optimal policy indirectly represented through value functions. \gls{dp} provides an essential foundation and collection of algorithms based on value-based methods. Value-based methods can be viewed as attempts to achieve the same goal as \gls{dp}, only with less computation and without assuming a perfect model of the environment. In value-based methods, we use value functions to organise and structure the search for good policies: we find an optimal value function, which is then used to derive an optimal policy for the agent. The goal is to learn a function that assigns a value to each state or to each state-action pair such that the value function accurately represents the expected return starting from that state (or state-action pair) and acting optimally thereafter.

\subsection{Policy Iteration} \label{sec:policy_iter_sec}
We can use $\vpi$ to improve $\pi$ and yield a better policy $\pi'$, and we can repeat this process by computing $v_{\pi'}$ and to generate an even better policy $\pi''$. We can thus obtain a sequence of monotonically improving policies and value functions. Each policy is guaranteed to be a strict improvement over the previous one (unless it is already optimal). Because a finite \gls{mdp} has only a finite number of deterministic policies, this process must converge to an optimal policy and the optimal value function in a finite number of iterations.
\begin{figure}[ht]
\centering
\begin{subfigure}{.5\textwidth}
  \centering
  \includegraphics[height=.65\linewidth]{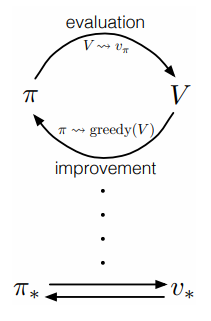}
\end{subfigure}%
\begin{subfigure}{.5\textwidth}
  \centering
  \includegraphics[height=.60\linewidth]{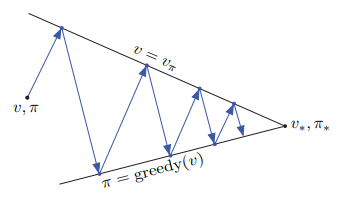}
\end{subfigure}
\caption[Generalised Policy Iteration]{The interaction between evaluation and improvement processes in \acrshort{gpi}. Source: \textcite[86-87]{sutton_reinforcement_2018}.}
\label{fig:gpi}
\end{figure}
As discussed before in \vref{sec:optimal_valuef_policies}, we can easily obtain the optimal policies once we have found the optimal value functions, namely $\vstar$ and $\qstar$, which satisfy the Bellman optimality equations shown in \vref{sec:bellmaneq}. To do so, we use a technique called \textit{policy iteration}, which is the foundation of all value-bases methods in \gls{rl}:
\begin{enumerate}
    \item \textit{Policy evaluation}, also known as \textit{prediction}, is where we iteratively compute the value function for a given policy.
    \item \textit{Policy improvement}, also known as \textit{control}, is where we compute an improved policy given the value function for that policy.
\end{enumerate}
In policy iteration, we first use policy evaluation to fully compute the value function $\vpi$ for a given policy until convergence. Then we apply policy improvement where we compute gradually build a new, improved policy $\pi'$ by greedily selecting actions from $\vpi$ by visiting every state until $\pi'$ has converged. \textit{\gls{gpi}} is a generalisation of policy iteration, where we do approximate policy evaluation and approximate policy improvement. One process takes the policy as given and performs some form of policy evaluation, changing the value function to be more like the true value function for the policy. The other process takes the value function as given. It performs some form of policy improvement, changing the policy to improve it, assuming that the value function is its value function. A policy or value function that remains unchanged by either process is optimal, and thus, it has converged. \textit{Value iteration} is a special case of \gls{gpi} where the policy evaluation is stopped after just one sweep of the state space: both \gls{gpi} and value iteration have been proven to converge with the same guarantees as policy iteration.

\subsection{Generalised Policy Iteration} \label{sec:gpi}
Policy iteration consists of two simultaneous, interacting processes, one making the value function consistent with the current policy, called policy evaluation, and the other making the policy greedy with respect to the current value function, policy improvement. In policy iteration, these two processes alternate, each completing before the other begins, but this is unnecessary. In value iteration, for example, only a single iteration of policy evaluation is performed between each policy improvement. Sometimes, a single state is updated in one process before returning to the other. As long as both processes continue to update all states, the ultimate result is typically the same—convergence to the optimal value function and an optimal policy. We use the term \acrfull{gpi} to refer to the general idea of letting policy-evaluation and policy improvement processes interact, independent of the granularity and other details of the two processes. Almost all \gls{rl} methods are well described as \gls{gpi}. All have identifiable policies and value functions, with the policy continuously being improved with respect to the value function and the value function always being driven toward the value function for the policy, as suggested by \cref{fig:gpi}. If both the evaluation and improvement processes stabilise and no longer produce changes, then the value function and policy must be optimal. The value function stabilises only when it is consistent with the current policy, and the policy stabilises only when it is greedy with respect to the current value function. Thus, both processes stabilise only when a policy is greedy with respect to its own evaluation function. This implies that the Bellman optimality equation \vref{eq:bellman_opt_vpi} holds, and thus, the policy and the value function are optimal.

\subsection{On-policy and Off-policy Methods}
In \textit{on-policy} algorithms, the agent learns the value function $\vpi(s)$ or the action-value function $\qpi(s)$ with respect to the policy $\pi$ that it is currently executing.
In \textit{off-policy} algorithms, the agent learns the value function $\vpi(s)$ or the action-value function $\qpi(s)$ for a deterministic optimal policy $\pi^*$ that may be unrelated to the policy followed. The policy that generates the data is called the \textit{behaviour policy} $b$, whereas the \textit{target policy} $\pi$ is the policy whose value function we want to learn. Since we use data from the behaviour policy to update a different target policy, methods such as \textit{importance sampling} are often needed to offset this mismatch. Thus, off-policy methods are often of greater variance and are slower to converge than on-policy methods. They are also particularly problematic when it comes to function approximation, as it will be later explained in \cref{sec:off_policy_approx}.

\subsubsection{The Exploration Dilemma}
As mentioned in \vref{sec:exploration} and \cref{sec:gpi}, learning control methods face a dilemma: they seek to learn action values conditional on subsequent optimal behaviour, but they need to behave non-optimally to explore all actions and find the optimal actions. On-policy and off-policy methods use two different approaches to learn about the optimal policy while behaving according to an exploratory policy. In on-policy methods, the exploration-exploitation tradeoff is particularly delicate, as the agent needs to find a near-optimal policy that still explores. The most straightforward idea for ensuring continual exploration is to make the policy \textit{soft} so that $\pi(a|s) > 0$ for all $s \in \S$ and all $a \in A(s)$, but gradually shifted closer and closer to a deterministic optimal policy. The \textit{$\epsilon$-greedy} policy is the $\epsilon$-soft policy that is closer to the greedy policy. It has probability $\epsilon$ to select an action at random and $1 - \epsilon + \frac{\epsilon}{|\A(s)|}$ probability to select the greedy action.
Guaranteeing exploration in off-policy methods seems more straightforward since the target policy can be an optimal greedy policy, whilst behaviour policy can be an exploratory \textit{$\epsilon$-greedy} policy.

\subsection{Monte Carlo Methods}
\gls{mc} methods are ways of solving the \gls{rl} problem based on averaging sample returns. \gls{mc} methods do not assume complete knowledge of the environment dynamics as they only require experience, meaning sample sequences of states, actions and rewards from actual or simulated interaction with an environment. A model is required only to generate sample transitions, not the complete probability distributions of all transitions required for \gls{dp}. We refer to "Monte Carlo" as methods based on averaging \textit{complete} returns, underlining the offline nature of these methods since they require complete episode rollouts. The first advantage of \gls{mc} methods is that they can be used to learn optimal behaviour directly from the environment, with no model of the environment's dynamics. Second, they can be used with simulation or sample models. Third, it is easy and efficient to focus \gls{mc} methods on a small subset of states that are of specific interest. A fourth advantage is that violations of the Markov property may less harm them because they do not bootstrap, meaning they do not use the Bellman equations consistency property for their updates. \gls{mc} control methods still follow the overall schema of \gls{gpi}.

\subsection{Temporal-Difference Methods}
\gls{td} learning is a core and innovative aspect of \gls{rl} that combines \gls{mc} aspects with \gls{dp} principles. Like \gls{mc} methods, \gls{td} methods can learn directly from raw experience without a model of the environment's dynamics. Like \gls{dp}, \gls{td} methods update estimates based partly on other learned estimates without waiting for the outcome. The general idea of updating estimates based on other estimates is called \textit{bootstrapping}. Whereas \gls{mc} methods must wait until the end of the episode to compute the return $G_t$ and determine the increment to $V(S_t)$, \gls{td} methods need to wait only until the next timestep. At time $t+1$, they immediately form a target and make a useful update using the observed reward $R_{t+1}$ and the estimate $V(S_{t+1})$. The simplest form of \gls{td} learning is a one-step method that makes the update
\begin{equation}
    V(S_t) \leftarrow V(S_t) + \alpha \left[ R_{t+1} + \gamma V(S_{t+1}) - V(S_t) \right] \label{eq:td_update}
\end{equation} immediately on transition to $S_{t+1}$ and receiving $R_{t+1}$.
We call $R_{t+1} + \gamma V(S_{t+1})$ the \textit{\gls{td} target} and we define $\delta_t$ as
\begin{equation}
    \delta_t \defeq R_{t+1} + \gamma V(S_{t+1}) - V(S_t) \label{eq:td_error}
\end{equation} the \textit{\gls{td} error}, which is used in various forms in many \gls{rl} algorithms.

We use the \gls{gpi} to build an on-policy \gls{td} control method that uses the action-value function rather than the state-value function in \cref{eq:td_update} for the \gls{td} update:
\begin{equation}
Q(S_t, A_t) \leftarrow Q(S_t, A_t) + \alpha \left[ R_{t+1} + \gamma Q(S_{t+1},A_{t+1}) - Q(S_t, A_t) \right]. \label{eq:sarsa_update}
\end{equation}
This leads to the \gls{sarsa} control method shown in \cref{algo:sarsa}. Note that the "(0)" notation refers to the fact that this is the one-step version.
\begin{algorithm}[ht]
\caption{\gls{sarsa}(0), On-policy TD Control} \label{algo:sarsa}
\begin{algorithmic}[1]
\State Initialize \( Q(s, a) \) for all \( s \in \mathcal{S} \), \( a \in \mathcal{A}(s) \), arbitrarily except that \( Q(\text{terminal state}, \cdot) = 0 \)
\For {each episode}
    \State Initialize \( S \)
    \State Choose \( A \) from \( S \) using policy derived from \( Q \) (e.g., \(\epsilon\)-greedy)
    \For {each step of episode}
        \State Take action \( A \), observe $R$, $S'$
        \State Choose \( A' \) from \( S' \) using policy derived from \( Q \) (e.g., \(\epsilon\)-greedy)
        \State \( Q(S, A) \leftarrow Q(S, A) + \alpha [ R + \gamma Q(S', A') - Q(S, A) ] \)
        \State \( S, A \leftarrow S', A' \)
    \EndFor
\EndFor
\end{algorithmic}
\end{algorithm}
In general, \gls{td} methods try to find estimates that would be correct for the Markov process's maximum-likelihood model. The maximum-likelihood estimate of a parameter is the parameter value whose probability of generating the data is greatest. In contrast, \gls{mc} methods find the estimates that minimise the mean-squared error on the training set. This is why, despite being biased, \gls{td} methods typically converge more quickly than \gls{mc}. At the same time, this makes \gls{td} methods more fragile to violations of the Markov property.

\subsection{Q-Learning}
One of the early breakthroughs in \gls{rl} was the development of an off-policy \gls{td} control algorithm known as \textit{Q-learning} \autocite{watkins_q-learning_1992}, whose update rule is defined by:
\begin{equation}
Q(S_t, A_t) \leftarrow Q(S_t, A_t) + \alpha \left[ R_{t+1} + \gamma \max_{a} Q(S_{t+1}, a) - Q(S_t, A_t) \right]. \label{eq:q_update}
\end{equation}
Here, the learned action-value function $Q$ directly approximates $\qstar$ independent of the policy being followed; hence the method being off-policy. Q-learning follows the pattern of value iteration described in \cref{sec:policy_iter_sec} and is implemented in \cref{algo:q_learning}.
\begin{algorithm}[ht]
\caption{Q-Learning, Off-policy TD Control} \label{algo:q_learning}
\begin{algorithmic}[1]
\State Initialise $Q(s, a)$ for all $s \in \S$, $a \in \A(s)$, arbitrarily except that $Q(\text{terminal state}, \cdot) = 0$
\For{each episode}
    \State Initialise \( S \)
    \For {each step of episode}
        \State Choose action $A$ using policy derived from $Q$ (e.g., $\epsilon$-greedy)
        \State Take action $A$, observe $R$, $S'$
        \State $Q(S, A) \leftarrow Q(S, A) + \alpha [R + \gamma \max_{a} Q(S', a) - Q(S, A)]$
        \State $S \leftarrow S'$
    \EndFor
\EndFor
\end{algorithmic}
\end{algorithm}

\section{Value-based Methods with Approximation} \label{sec:value_approx}
In value-based methods, the approximate value function is represented as a parametrised functional form with weight vector $\w \in \mathcal{R}^d$. We write $\hat v(s,\w) \approx \vpi(s)$ for the approximate value of state $s$ given vector $\w$. Usually, $\hat v$ is the function computed by a multi-layer \gls{aan}, with $\w$ the vector of connection weights in all the layers. Typically, the number of weights (the dimensionality of $\w$) is much less than the number of states ($d \ll \left| \S \right|$), and changing one weight changes the estimated value of many states. Consequently, when a single state is updated, the change generalises from that state to affect the values of many other states. This generalisation makes the learning potentially more powerful but potentially more challenging to manage and understand.

However, not all function approximation methods are equally well suited for use in \gls{rl}. The most sophisticated artificial neural network and statistical methods assume a static training set over which multiple passes are made. In \gls{rl}, it is vital that learning be able to occur online while the agent interacts with its environment or with a model of its environment. To do this requires methods that can learn efficiently from incrementally acquired data. In addition, \gls{rl} generally requires function approximation methods to handle nonstationary target functions (target functions that change over time). For example, in control methods based on \gls{gpi}, we often seek to learn $\qpi$ while $\pi$ changes. Even if the policy remains the same, the target values of training examples are nonstationary if they are generated by bootstrapping methods (such as \gls{dp} and \gls{td} learning). Methods that cannot easily handle such nonstationarity are less suitable for \gls{rl}.

To define an explicit learning objective for prediction, we must then say which states we care most about, as making one state’s estimate more accurate invariably means making others’ less accurate (by assumption, we have far more states than weights). We must specify a state distribution $\mu(s) \geq 0, \sum_s \mu(s) = 1$, representing how much we care about the error in each state $s$ error. We define the error in a state $s$ as the mean square of the difference between the approximate value $\hat v(s,\w)$ and the true value $\vpi(s)$, and we weight this over the state space by $\mu$. We obtain a natural objective function, the \textit{mean square value error}.
\begin{definition}[Mean Square Value Error]
    \[ \MSVEm(\w) \defeq \sum_{s \in \S} \mu(s) \left[ \vpi(s) - \hat v(s,\w) \right] ^2, \]
    where $\mu(s)$ is called the \textit{on-policy distribution} and is chosen as the fraction of time spent in $s$ under policy $\pi$.
\end{definition}
It is not completely clear that the $\MSVEm$ is the right performance objective for \gls{rl}. The ultimate purpose and the reason we are learning a value function is to find a better policy. The best value function for this purpose is not necessarily the best for minimizing $\MSVEm$. Nevertheless, it is not yet clear what a more useful alternative goal for value prediction might be. Complex function approximators may seek to converge instead to a local optimum, a weight vector $\w^*$ for which $\MSVEm(\w^*) \leq \MSVEm(\w)$ for all $\w$ in some neighbourhood of $w^*$. This is typically the best that can be said for nonlinear function approximators. For many cases of interest in \gls{rl}, there is no guarantee of convergence to an optimum or even to within a bounded distance of an optimum. Some methods may diverge, with their $\MSVEm$ approaching infinity in the limit.

\subsection{Stochastic-gradient Methods}
To minimise $\MSVEm$, we typically use methods based on \textit{\gls{sgd}}, which are well suited to online \gls{rl}. A common strategy is to minimise the $\MSVEm$ on observed examples, which we can do by using \gls{sgd} to adjust the weight vector $\w$ after each example by a small amount in the direction that would most reduce the error on that example:
\begin{align}
\w_{t+1} &\defeq \w_t - \frac{1}{2} \alpha \grad \left[ \vpi(S_t) - \hat v(S_t, \w_t) \right]^2 \label{eq:sgd_derivation}\\
        &= \w_t + \alpha \left[ \vpi(S_t) - \hat v(S_t, \w_t) \right] \grad \hat v(S_t, \w_t) \label{eq:sgd}
\end{align} where $\alpha$ is a positive step-size parameter, also called the \textit{learning rate}.

\subsection{Semi-gradient Methods} \label{sec:semi_gradient}
One problem with \cref{eq:sgd} is that in practice, we do not have the true value $\vpi(S_t)$ but some, possibly random or bootstrapped, approximation to it. Therefore, we cannot perform an exact update because $\vpi(S_t)$ is unknown, but we can approximate it using a target estimate $U_t$. If $U_t$ is an unbiased estimate, meaning that $\E{U_t|S_t=s} = \vpi(s)$ for each $t$, then $\w_t$ is guaranteed to converge to a local optimum under \gls{sgd}. An example of an unbiased estimate would be the \gls{mc} target $U_t \defeq G_t$.
If a bootstrapping estimate of $\vpi(S_t)$ is used as the $U_t$, such as in \gls{td} methods, then $U_t$ will also depend on the current value of the weight vector $\w_t$, which implies that they will be biased and will not produce a true gradient-descent method. The step from \cref{eq:sgd_derivation} to \cref{eq:sgd} relies on the target being independent of $\w_t$ and would not be valid if a bootstrapping estimate were used in place of $\vpi(S_t)$. Bootstrapping methods consider the effect of changing the weight vector $\w_t$ on the estimate but ignore its effect on the target. Because they include only a part of the gradient, they are accordingly called \textit{semi-gradient methods}.
Semi-gradient methods are still preferred for the usual reason that bootstrapped estimates enable significantly faster online learning.

We now extend the \gls{sarsa} algorithm in \cref{algo:sarsa} to function approximation. Similarly to what we did for \cref{eq:sarsa_update}, we use the update rule for the state-value function in \cref{eq:sgd} to derive an update rule using the approximate action-value function instead:
\begin{equation}
    \w_{t+1} \defeq \w_t + \alpha \left[ R_{t+1} + \gamma \hat q(S_{t+1},A_{t+1},\w_t) - \hat q(S_t,A_t,\w_t \right] \grad \hat q(S_t,A_t,\w_t). \label{eq:sarsa_grad_update}
\end{equation}
We have now obtained the update rule for the semi-gradient one-step on-policy control \gls{sarsa} in \cref{algo:sgd_sarsa}.

\begin{algorithm}[ht]
\caption{Semi-gradient \gls{sarsa}(0), On-policy TD Control} \label{algo:sgd_sarsa}
\begin{algorithmic}[1]
    \State Input: a differentiable function \( \hat{q} : \mathcal{S} \times \mathcal{A} \times \mathbb{R}^d \to \mathbb{R} \)
    \State Initialise value-function weights \( \textbf{w} \) arbitrarily $\mathbb{R}^d$ (e.g., \( \textbf{w} = \textbf{0} \))
    \For{each episode}
        \State Initialise \( S, A \) (first state and action)
        \For{each step of episode}
            \State Take action \( A \), observe \( R, S' \)
            \If{\( S' \) is terminal}
                \State \( \w \leftarrow \w + \alpha \left[ R - \hat q(S, A, \w) \right] \grad \hat q(S, A, \w) \)
                \State Go to the next episode
            \EndIf
            \State Choose \( A' \) as a function of \( \hat q(S',\cdot,\w) \) (e.g., \(\epsilon\)-greedy)
            \State \( \w \leftarrow \w + \alpha \left[ R + \gamma \hat q(S', A', \w) - \hat q(S, A, \w)\right] \grad \hat q(S, A, \w) \)
            \State \( S, A \leftarrow S', A' \)
        \EndFor
    \EndFor
\end{algorithmic}
\end{algorithm}

\subsection{Function Approximation with Off-policy Methods} \label{sec:off_policy_approx}
The extension to function approximation turns out to be significantly different and harder for off-policy learning than it is for on-policy learning. The tabular off-policy methods readily extend to semi-gradient algorithms, but these algorithms do not converge as robustly as they do under on-policy training. The fact that the distribution of updates in the off-policy case is not according to the on-policy distribution creates additional problems. Two general approaches have been explored to deal with this. One is to use importance sampling methods to warp the update distribution back to the on-policy distribution. The other is to develop true gradient methods that do not rely on any special distribution for stability, which we will not cover here.

By changing their update rule, we can easily convert \cref{algo:q_learning} and \cref{algo:sgd_sarsa} to off-policy semi-gradient form.
For the one-step off-policy semi-gradient \gls{sarsa} we need to introduce the \textit{per-step importance sampling ratio} $\rho_t$:
\begin{equation}
    \rho_t \defeq \frac{\pi(A_t|S_t)}{b(A_t|S_t)}. \label{eq:importance_sampling}
\end{equation}
Then we change the update rule of \cref{eq:sarsa_grad_update} to be:
\[ \w_{t+1} \defeq \w_t + \alpha \rho_t \left[ R_{t+1} + \gamma \hat q(S_{t+1},A_{t+1},\w_t) - \hat q(S_t,A_t,\w_t \right] \grad \hat q(S_t,A_t,\w_t). \]
In the case of Q-learning, we do not need importance sampling, as the update rule \cref{eq:q_update} matches exactly what the target policy is doing, namely learning about the optimal greedy policy. As a result, the update remains the same:
\begin{equation}
    \w_{t+1} \defeq \w_t + \alpha \left[ R_{t+1} + \gamma \max_{a} \hat q(S_{t+1}, a, \w_t) - \hat q(S_t, A_t, \w_t) \right] \grad \hat q(S_t, A_t, \w_t) \label{eq:gradient_q}
\end{equation}
\textcite{tsitsiklis_analysis_1997} has shown that off-policy Q-learning with non-linear function approximations can cause the Q-network to diverge, despite non-gradient Q-learning having the best convergence guarantees of all control methods \autocite{sutton_reinforcement_2018}. This shows the dangers and instability of using approximations combined with off-policy methods. In particular, \textcite{sutton_reinforcement_2018} shows how divergence arises whenever we combine all of the following three elements, called \textit{the deadly triad}:
\begin{enumerate*}
    \item Function approximation
    \item Bootstrapping and
    \item Off-policy training.
\end{enumerate*}

\subsection{Deep Q-Learning} \label{sec:dqn} 
We now introduce \acrfull{dqn}, a \gls{drl} algorithm first shown in \textcite{mnih_playing_2013}, \textcite{mnih_human-level_2015}, that generalises Q-learning to the function approximation case. In \ref{sec:off_policy_approx} we have talked about the instability and divergence of Q-learning with function approximation. \gls{dqn} attempts to mitigate these issues using a number of enhancements. First of all, it uses a \textit{minibatch method} that updates weights only after accumulating gradient information over a set threshold of observation. The original implementation used a gradient-ascent algorithm called RMSProp, but modern implementations (including the one we use in this thesis) often use Adam instead \autocite{kingma_adam_2015}. Moreover, it modifies the classic Q-learning algorithm in three ways.
(1) Through the use of a method called \textit{experience replay}, it stores the agent's experience in a replay memory $\D$ that is accessed to perform the weight updates. At each timestep, a transition including the agent's actions, the observation and the reward are added to $\D$. Then, a minibatch update is performed based on experiences \textit{sampled uniformly at random} from $\D$. Instead of $S_{t+1}$ becoming the new $S_t$ for the next update as in usual Q-learning, a new unconnected experience is drawn from $\D$ to supply data for the next update. This is enabled by the fact that Q-Learning is an off-policy algorithm. Thanks to experience replay \gls{dqn} gains several advantages over Q-Learning. First, it benefits from increased sample efficiency, as experiences are stored and used for many updates instead of being immediately discarded after the update. Second, it reduces the variance of the updates because successive updates are not correlated with one another as they would normally be. By removing the dependence of successive experiences on the current weights it eliminates one source of instability.
(2) A modified version of \cref{eq:gradient_q} with \textit{fixed Q-targets} is used to further improve stability. As seen in \cref{sec:semi_gradient}, Q-learning is a semi-gradient method whose update rule depends on a bootstrapped action-value function estimate. The problem with semi-gradient methods is that the target is a function of the same parameters that are being updated and this can lead to oscillations and divergence. Another Q-network called the \textit{target network} is used to address this. After a certain number of updates $C$ to the weights $\w$, the online network's current weights are inserted into the target network. The weights of the target network are then held fixed for the next $C$ updates of $\w$. This additional fixed network is then used to compute the Q-learning targets, thus avoiding the "moving Q-targets" problem. The update rule changes as follows:
\begin{equation}
\w_{t+1} \defeq \w_t + \alpha \left[ R_{t+1} + \gamma \max_{a} \hat q_{\text{tg}}(S_{t+1}, a, \w_t) - \hat q(S_t, A_t, \w_t) \right] \\ \grad \hat q(S_t, A_t, \w_t), \label{eq:dqn_update}
\end{equation} where $q_{tg}(s,a,\w_t)$ indicates the action-value estimate of the target Q-network.
(3) A final modification that improves the stability of standard Q-learning is the clipping of the \gls{td} error $R_{t+1} + \gamma \max_{a} \hat q_{dup}(S_{t+1}, a, \w_t) - \hat q(S_t, A_t, \w_t)$ so that it remains in the interval $[-1,1]$.
We show the pseudocode for \gls{dqn} in \cref{algo:dqn}.
\begin{algorithm}[ht]
\caption{Deep Q-Learning with Experience Replay} \label{algo:dqn}
\begin{algorithmic}[1]
\State Initialise replay memory $\D$ to capacity \(N\)
\State Initialise action-value function \(\hat q\) with random weights \(\theta\)
\State Initialise target action-value function \(\hat{q_{tg}}\) with weights \(\theta^- = \theta\)
\For{episode \(= 1, M\)}
    \State Initialize state \(S\)
    \For{timestep \(t = 1, T\)}
        \State With probability \(\epsilon\) select a random action \(A\)
        \State Otherwise select \(A = \arg\max_{a} \hat q(s, a, \theta)\)
        \State Take action $A$, observe $R$, $S'$
        \State Store transition \((S, A, R, S')\) in \(\D\)
        \State Sample random mini-batch of transitions from \(\D\)
        \For{each \((S, A, R, S')\)}
            \If{\(S'\) is terminal}
                \State \(y_j = R\)
            \Else
                \State \(y_j = R + \gamma \max_{a'} \hat{q_{tg}}(S', A', \theta^-)\)
            \EndIf
        \EndFor
        \State Perform a gradient descent step on \((y_j - \hat q(s, a, \theta))^2\) with respect to the parameters \(\theta\)
        \State Every \(C\) steps reset \(\hat{q_{tg}} = \hat q\)
    \EndFor
\EndFor
\end{algorithmic}
\end{algorithm}
The original \gls{dqn} has been substantially improved over the next years by using additional tricks to further stabilise learning. One of the problems of both classic Q-learning and \gls{dqn} is the \textit{maximisation bias}. This stems from the fact that a maximum overestimated value is implicitly used to estimate the maximum value, which leads to significant positive bias. For instance, consider a state where all the actions have zero true values but whose action-value estimates are distributed, some above and some below zero. The maximum of the true values is zero, but the maximum of the estimates is positive, leading to positive bias. \textcite{van_hasselt_deep_2016} overcomes this problem by decoupling action selection to action evaluation. The greedy policy is evaluated according to the online network but uses the target network to estimate its value. The \gls{dqn} update changes to:
\begin{flalign*}
\w_{t+1} & \defeq \w_t + \alpha \Big[ R_{t+1} + \gamma \hat{q}_{\text{tg}}(S_{t+1}, \arg\max \hat{q}(S_t, A_t, \w_t), \w_t) && \\
& - \hat{q}(S_t, A_t, \w_t) \Big] \grad \hat{q}(S_t, A_t, \w_t), &&
\end{flalign*}
and leads to significantly better performance. Another notable improvement to \gls{dqn} is the use of \textit{Prioritised Experience Replay} \autocite{schaul_prioritized_2016}. Instead of sampling the replay buffer uniformly at random, the experiences in $\D$ are sampled with respect to assigned priorities based on the magnitude of their \gls{td} error $\delta$, where higher $|\delta|$ values indicate that the experience is surprising and could result in significant Q-value updates.
Even more improvements to \gls{dqn} have been added over the years, culminating in the RAINBOW algorithm \autocite{hessel_rainbow_2018}, which is a version of \gls{dqn} that includes all improvements up to that date, namely what listed before with the addition of Dueling \gls{dqn}, multi-step learning, distributional \gls{dqn} and noisy nets.

\section{Policy Gradient Methods} \label{sec:policy_gradient}
Policy gradient methods directly learn a \textit{parametrised policy} that can select actions without consulting a value function. A value function may still be used to learn the policy parameter but is not required for action selection. We use the notation $\th \in \mathcal{R}^{d'}$ for the policy's parameter vector and $\pi(a|s,\th) = \Pr{A_t=a \mid S_t = s, \th_t = \th} $ for the probability that action $a$ is taken at time $t$ given the environment is in state $s$ at time $t$ with parameter $\th$. If a method also uses a value function, then the value function's weight vector is denoted $\w \in \mathcal{R}^d$.

We also define $J(\th)$ as the \textit{objective function}, a scalar performance measure with respect to the policy parameter. Policy gradient methods seek to maximise performance, so their updates approximate gradient \textit{ascent} in $J$:
\begin{equation}
    \theta_{t+1} = \theta_t + \alpha \widehat{\grad J(\th)},
\end{equation}
where $\widehat{\grad J(\th)} \in \mathcal{R}^{d'}$ is a stochastic estimate whose expectation approximates the gradient of the objective function with respect to its argument $\th_t$. All methods that follow this general schema are called \textit{policy gradient methods}: this includes \textit{actor-critic methods}, which will be explained in \cref{sec:actor_critic}. Policy gradient methods follow the exact policy gradient and thus usually have better convergence guarantees than semi-gradient methods in \cref{sec:semi_gradient}.

\subsection{The Policy Gradient Theorem}
We define the objective function as:
\[ J(\th) \defeq v_{\pi_\th}(s_0) = \mathbb{E}_{\pi_\th} \left[ G_t \mid S_t = s_0 \right], \]
that is, the objective is to learn a policy that maximises the cumulative future reward to be received, starting from an initial state $s_0$. We assume an episodic case with no discounting ($\gamma = 1$), but it can be extended for continuing discounted tasks. Thanks to the \textit{policy gradient theorem} (\cref{def:grad_theo}), we have an analytic expression for the gradient of the objective function with respect to the policy parameter that does not involve the derivative of the state distribution $\mu(s)$.
\begin{theorem}[The Policy Gradient Theorem] \label{def:grad_theo}
The policy gradient theorem for the episodic case is:
    \[ \grad J(\th) \propto \sum_s \mu(s) \sum_a \qpi(s,a) \grad \pi(a|s,\th), \]
\end{theorem} where $\pi$ denotes the policy corresponding to parameter vector $\th$, the constant of proportionality is the average episode length and $\mu$ is the on-policy distribution of states under $\pi$. 

\subsection{REINFORCE}
We use \cref{def:grad_theo} to derive the update rule for REINFORCE \autocite{williams_simple_1992}, a classic policy-gradient learning algorithm:
\begin{equation}
    \th_{t+1} = \th_t + \alpha G_t \grad \ln\pi(A_t|S_t,\th_t),\ \label{eq:reinforce_update}
\end{equation}
where $ \grad\ln\pi(A_t|S_t,\th_t) = \frac{\grad\pi(A_t|S_t,\th_t)}{\pi(A_t|S_t,\th_t)} $.

The right-hand side of \cref{eq:reinforce_update} is a quantity that can be sampled on each time step and whose expectation is proportional to the gradient. Each increment is proportional to the product of a return $G_t$ and a vector, the gradient of the probability of taking the action taken divided by the probability of taking that action $A_t$ on future visits on state $S_t$. The update increases the parameter vector in this direction proportional to the return and inversely proportional to the action probability. The former makes sense because it causes the parameter to move most in directions, favouring actions that yield the highest return. The latter makes sense because, otherwise, frequently selected actions have an advantage and might win out even if they do not yield the highest return.
REINFORCE is a \gls{mc} algorithm, so all updates are made offline as it needs the complete return from time $t$, which includes all future rewards up to the end of the episode. REINFORCE has good theoretical convergence properties as the expected update over an episode is in the same direction as the performance gradient. However, as a \gls{mc} method, it is of high variance and thus produces slow learning. We give the pseudocode in \cref{algo:reinforce}.
\begin{algorithm}
\caption{REINFORCE: Monte-Carlo Policy-Gradient Control} \label{algo:reinforce}
\begin{algorithmic}[1]
\State Input: a differentiable policy parametrisation $\pi(a|s,\th)$
\State Initialise policy parameter \( \th \in \mathbb{R}^{d'}\) arbitrarily
\For{each episode}
    \State Generate an episode \( S_0, A_0, R_1, \ldots, S_{T-1}, A_{T-1}, R_T \) following \( \pi(\cdot| \cdot, \th) \)
    \For{each step of the episode $t = 0,1,\dots,T-1$:}
        \State \( G \leftarrow \sum_{k=t+1}^{T} \gamma^{k-t-1} R_k \)
        \State \( \th \leftarrow \th + \alpha \gamma^t G \nabla \ln \pi(A_t|S_t, \th) \)
    \EndFor
\EndFor
\end{algorithmic}
\end{algorithm}

\subsection{REINFORCE with Baseline}
The policy gradient \cref{def:grad_theo} can be generalised to include a comparison of the action value to an arbitrary \textit{baseline} $b(s)$. The baseline $b(s)$ can be any function as long as it does vary with $a$, such that we can subtract it from the policy gradient without changing expectation:
\begin{equation}
    \grad J(\th) \propto \sum_s \mu(s) \sum_a \left( \qpi(s,a) - b(s) \right) \grad \pi(a|s,\th), \label{def:grad_theo_baseline}
\end{equation}
The policy gradient theorem with baseline \cref{def:grad_theo_baseline} can be used to derive an update rule for a new version of REINFORCE that includes a general baseline:
\begin{equation}
    \th_{t+1} = \th_t + \alpha \left( G_t - b(S_t) \right) \grad \ln\pi(A_t|S_t,\th_t),\ \label{eq:reinforce_update_baseline}
\end{equation}
The baseline leaves the expected value of the update unchanged but can significantly affect its variance. In some states, all actions have high values, necessitating a high baseline to help distinguish between the more valuable and less valuable actions. A low baseline is more suitable in other states where all actions hold low values. One natural choice for the baseline is an estimate of the state value, $\hat v(S_t,\w)$. Because REINFORCE is a \gls{mc} method for learning the policy parameter $\th$, we also use \gls{mc} to learn the state-value function weights $\w$.
\begin{algorithm}[ht]
\caption{REINFORCE with Baseline} \label{algo:reinforce_baseline}
\begin{algorithmic}[1]
\State Input: a differentiable policy parametrisation $\pi(a|s,\th)$
\State Input: a differentiable state-value function parametrisation $\hat v(s,\w)$
\State Initialise policy parameter \( \th \in \mathbb{R}^{d'}\) and state-value weights $\w \in \mathbb{R}^d$ arbitrarily
\For{each episode}
    \State Generate an episode \(S_0, A_0, R_1, \dots, S_{T-1}, A_{T-1}, R_T\) following \(\pi(\cdot|s,\th)\)
    \State \(G \leftarrow 0\)
    \For{\(t \leftarrow T-1, T-2, \dots, 0\)}
        \State \( G \leftarrow \sum_{k=t+1}^{T} \gamma^{k-t-1} R_k \)
        \State \(\delta \leftarrow G - \hat v(S_t, \w)\)
        \State \(\w \leftarrow \w + \alpha^\w \delta \grad \hat v(S_t, \w)\)
        \State \(\th \leftarrow \th + \alpha^\th \gamma^t \delta \grad \ln \pi(A_t|S_t,\th)\)
    \EndFor
\EndFor
\end{algorithmic}
\end{algorithm}
Note how in \cref{algo:reinforce_baseline}, the learned state-value function estimates the value of only the first state of each state transition. This estimate sets a baseline for the subsequent return but is made before the transition's action and thus cannot be used to assess that action.

\section{Actor-Critic Methods} \label{sec:actor_critic}
Actor-critic policy-gradient methods learn approximations of policy and value functions, where  \textit{actor} refers to the learned policy, and  \textit{critic} refers to the learned state-value function. The difference with \cref{algo:reinforce_baseline} is that the state-value function is also applied to the \textit{second} state of the transition. The estimated value of the second state is then discounted and added to the reward. It constitutes the one-step return $G_{t:t+1}$, which is a useful estimate of the actual return and can be used to assess the action. The one-step return is often superior to the actual return in terms of its variance and computational complexity, even though it introduces bias.

Here, we examine one-step actor-critic methods that replace the full return of REINFORCE in \cref{eq:reinforce_update_baseline} with the one-step return. To estimate the one-step return, we can use any \gls{td} method, including Q-learning, \gls{td}(0) and \gls{sarsa}(0). The one-step version is fully online and incremental, meaning that states, actions and rewards are processed as they occur and never revisited. We give a summary of the update rules of classic variants of actor-critic methods, with REINFORCE as a reference:
\begin{align}
\grad J(\theta) &= \mathbb{E}_{\pi}\left[ G_t \grad \log \pi(a_t | s_t) \right] &\text{(REINFORCE)} \nonumber\\
\grad J(\theta) &= \mathbb{E}_{\pi}\left[ \delta_t \nabla \log \pi(a_t | s_t) \right]  &\text{(\gls{sarsa} Actor-Critic)} \nonumber\\
\grad J(\theta) &= \mathbb{E}_{\pi}\left[ \hat q(S_t,A_t,\w) \grad \log \pi(a_t | s_t) \right] &\text{(Q Actor-Critic)} \nonumber\\
\grad J(\theta) &= \mathbb{E}_{\pi}\left[ \hat{A}(S_t,A_t,\w, \v) \grad \log \pi(a_t | s_t) \right] &\text{(Advantage Actor-Critic)} \nonumber
\end{align} 
where $\delta_t$ is the \gls{td} error \cref{eq:td_error} and $\hat{A}$ is the approximate advantage function described in \vref{def:advantage}:
\begin{align}
\delta_t &= R_{t+1} + \gamma \hat q(S_{t+1},A_{t+1}\w) - \hat q(S_t,A_t,\w) \\
\hat{A}(S_t, A_t, \w, \v) &= \hat{q}(S_t, A_t, \w) - \hat{v}(S_t, \v) \label{eq:approx_advantage}
\end{align}
We implement the one-step \gls{sarsa} Actor-Critic in \cref{algo:sarsa_actor-critic}
\begin{algorithm}[ht]
\caption{\gls{sarsa}(0) Actor-Critic} \label{algo:sarsa_actor-critic}
\begin{algorithmic}[1]
\State Input: differentiable action-value function \(\hat{q}(s, a, \w)\) and  policy \(\pi(a|s, \th)\) parametrisations
\State Initialize \(\w \in \mathbb{R}^d, \th \in \mathbb{R}^{d'}\) arbitrarily
\For{each episode}
    \State Initialize \(S\) (first state of episode)
    \State \(I \leftarrow 1\)
    \State Sample \(A \sim \pi(\cdot|S, \th)\)
    \For{each timestep of episode}
        \State Take action \(A\), observe \(S', R\)
        \State Sample \(A' \sim \pi(\cdot|S', \th)\)
        \State \(\delta \leftarrow R + \gamma \hat{q}(S', A', \w) - \hat{q}(S, A, \w)\)
        \State \(\w \leftarrow \w + \alpha^{\w} \delta \grad \hat q(S, A, \w)\)
        \State \(\th \leftarrow \th + \alpha^{\th} I \delta \grad \ln \pi(A|S, \th)\)
        \State \(I \leftarrow \gamma I\)
        \State \(S, A \leftarrow S', A'\)
    \EndFor
\EndFor
\end{algorithmic}
\end{algorithm}

\subsection{Advantage Actor-Critic} \label{sec:a2c}
\gls{a2c} is considered a state-of-the-art algorithm that uses the advantage function as the baseline. The advantage function makes an optimal baseline, as it intuitively indicates how much better it is to take a specific action in a state compared to the average value of that state. We previously defined the advantage function \cref{eq:approx_advantage} to use two distinct approximators to predict state and state-action values separately. However, this is rarely the case in practice, as it would be extremely inefficient. We use the fact that
\begin{equation}
\qpi(s, a) = \mathbb{E}_{\pi} \left[ R_t + \gamma \vpi(S_{t+1}) \mid S_t = s, A_t = a \right]
\end{equation}
to approximate the advantage function using the \gls{td} error computed with a single approximator for the state-value function:
\begin{equation}
\hat{A}(s, a, \w) = r + \gamma \hat{v}(s', \w) - \hat{v}(s, \w). \label{eq:advantage_calc}
\end{equation}
\textcite{mnih_asynchronous_2016} details an implementation of an asynchronous version of the Advantage Actor-Critic called \gls{a3c}. However, it was later found that the asynchronous aspect was a hindrance as it only increased the algorithm's complexity with no tangible performance benefits. In \textcite{wu_scalable_2017}, the synchronous version \gls{a2c} was found to perform better and is now widely adopted.
In \cref{algo:a2c}, we give the pseudocode for a basic version of \gls{a2c}.
\begin{algorithm}[ht]
\caption{(Synchronous) One-step Advantage Actor-Critic (A2C)} \label{algo:a2c}
\begin{algorithmic}[1]
\State Input: differentiable state-value function \(\hat{v}(s, \w)\) and  policy \(\pi(a|s, \th)\) parametrisations
\State Initialize \(\w \in \mathbb{R}^d, \th \in \mathbb{R}^{d'}\) arbitrarily
\For{each episode}
    \State Initialize \( S \) (first state of episode)
    \State \(I \leftarrow 1\)
    \For{each timestep of episode}
        \State Sample \(A \sim \pi(\cdot|S', \th)\)
        \State Take action \(A\), observe \(S', R\)
        \State \( \delta = R + \gamma \hat{v}(S', \w) - \hat{v}(S, \w) \)
        \State \( \w \gets \w + \alpha^{\w} \delta \nabla \hat{v}(S, \w) \)
        \State \( \th \gets \th + \alpha^{\th} I \delta \nabla \ln \pi(A|S, \th) \)
        \State \(I \leftarrow \gamma I\)
        \State \( S \gets S' \)
    \EndFor
\EndFor
\end{algorithmic}
\end{algorithm}
\gls{a2c} is a powerful algorithm that can be scaled easily using multiple workers. Its implementations often use workers that collect a small number of transitions in parallel and immediately use them for a single minibatch update, after which the transitions are discarded. These implementations use an n-step version of the Advantage Actor-Critic presented in \cref{algo:a2c}, where the change would be merely replacing the one-step return $G_t$ with $G_{t:t+n}$, computed by using the transitions stored in transient memory.

\subsection{Proximal Policy Optimisation (Clip)} \label{sec:ppo}

\gls{ppo} \autocite{schulman_proximal_2017} is a state of the art actor-critic algorithm based on Trust Region Policy Optimisation \autocite{schulman_trust_2015}.
Here, we consider the \gls{ppo}-Clip variant of the algorithm, which is more straightforward to implement and the most widely adopted \autocite{SpinningUp2018}. 
\gls{ppo} updates policies by taking the largest step possible while satisfying a special constraint on how close the new and old policies are allowed to be. 
This differs from the normal policy gradient, which keeps new and old policies close in parameter space. However, even slight differences in parameter space can substantially differ in performance, so a single bad step can collapse the policy performance. This makes it dangerous to use large step sizes with vanilla policy gradients, thus hurting its sample efficiency. PPO avoids this kind of collapse and quickly and monotonically improves performance.
\begin{algorithm}[ht]
\caption{Proximal Policy Optimisation (Clip)} \label{algo:ppo}
\begin{algorithmic}[1]
\State Initialize \(\w \in \mathbb{R}^d, \th \in \mathbb{R}^{d'}\) arbitrarily
\For{each iteration}
    \State Collect a set of trajectories \( \tau \) by running policy \( \pi(a|s, \th) \)
    \State Compute rewards-to-go \( G_t \) and advantages \( \hat{A}_t \) for each \( \tau \)
    \State Store old policy parameters \( \th_{\text{old}} \)
    \For{each optimization epoch}
        \For{each mini-batch \( (s, a, G, \hat{A}) \) in \( \tau \)}
            \State Compute \( r_t(\th) = \frac{\pi(a|s, \th)}{\pi(a|s, \th_{\text{old}})} \)
            \State Compute clipped surrogate objective:
            \[
            L(\th) = \min\left(r_t(\th) \hat{A}_t, \text{clip}(r_t(\th), 1 - \epsilon, 1 + \epsilon) \hat{A}_t\right)
            \]
            \State Update \( \th \) by maximizing \( L(\th) \)
            \State Update \( \w \) using \( G_t \)
        \EndFor
    \EndFor
\EndFor
\end{algorithmic}
\end{algorithm}
\gls{ppo} updates maximises a clipped \textit{surrogate objective function} instead of directly indirectly maximising the performance of the policy. By maximizing the surrogate objective, it encourages updates that increase the probability of actions that had positive advantages, i.e., were better than the average action in that state according to the old policy, thus encouraging the policy to take more beneficial actions in the future, while simultaneously preventing the policy from changing too drastically and potentially destabilizing the learning process. Before proceeding, we define the probability ratio $r_t(\th)$, similar to the importance sampling ratio we defined in \cref{eq:importance_sampling}.
\begin{equation}
    r_t(\th) \defeq \frac{\pi(a|s, \th)}{\pi(a|s, \th_{\text{old}})}.
\end{equation}
$r_t(\th)$ is used in \cref{eq:ppo_surrogate_obj} to essentially weigh the advantage of each action by how much more likely it is under the new policy compared to the old policy.
\gls{ppo} maximises the surrogate objective function $L^{\text{CLIP}}$:
\begin{equation}
L^{\text{CLIP}}(\th) \defeq \mathbb{E} \left[ \min \left( r_t(\th) \hat A_t, \text{clip} \left( r_t(\th), 1 - \epsilon, 1 + \epsilon \right) \hat A_t \right) \right] \label{eq:ppo_surrogate_obj}
\end{equation}
The clipping in the surrogate objective is governed by the small hyperparameter $\epsilon$, which prevents the policy from changing too much in a single update. If an action's probability under the new policy becomes too different from its probability under the old policy (outside of the range $\left[ 1 - \epsilon, 1 + \epsilon \right]$), the objective is clipped, which prevents the policy from receiving any additional benefit from increasing that action's probability any further. This prevents the policy from changing too drastically and helps maintain policy updates within a 'trust region', which can prevent the policy from destabilizing due to large updates.

The pseudocode for \gls{ppo} is given in \cref{algo:ppo}. Compared to \gls{a2c}, we see that \gls{ppo} has better sample efficiency due to having a trajectory buffer $\mathcal{D}$ used for multiple epochs of minibatch \gls{sgd} updates (or Adam updates, depending on the implementation).
Whereas \gls{a2c} stored transitions in transient memory up to the minibatch size before performing a single update and discarding them, in \gls{ppo}, the buffer $\mathcal{D}$ is large and contains transitions from multiple trajectories. Following shuffling, these transitions are divided into minibatches to update the policy and value function, and the procedure is repeated across multiple epochs, thereby increasing sample efficiency.
\chapter{Related Work}\label{chapter:related_work}

\acrfull{drl} has seen rapid and significant advancements in recent years, with environment frameworks pivotal in this progression. These frameworks offer structured, reproducible, and often realistic spaces where algorithms can be trained, verified, and benchmarked. Thanks to these environments, researchers can analyse and compare the performance of \gls{drl} agents in various conditions, ranging from simple bidimensional games to highly complex tridimensional simulators. Here, we give an overview of relevant literature concerning both algorithms and environment frameworks. We then conclude the discussion with studies that use \gls{drl} techniques to augment agents manipulating objects in physics-based environments.

\section{Algorithms}
\textcite{mnih_playing_2013} introduced a \gls{dqn} model that successfully learned control policies from high-dimensional sensory input using \gls{rl}. This model maintained a constant architecture and learning algorithm across seven Atari 2600 games from the \gls{ale}, outperforming previous approaches and even a human expert in some games, setting a precedent for further exploration and development in \gls{rl}.
Building upon this, \textcite{mnih_human-level_2015} presented a \gls{dqn} agent that surpassed the performance of all previous algorithms and achieved human-level play across 49 classic Atari 2600 games using end-to-end \gls{drl}. The agent, which used deep \gls{cnn} architecture and a novel variant of Q-learning, demonstrated robust learning across diverse tasks and showed that it could develop representations supporting adaptive behaviour from high-dimensional sensory inputs.
\textcite{hester_deep_2018} introduced a deep reinforcement learning algorithm called \gls{DQfD}, which leverages small sets of demonstration data to accelerate the learning process significantly. Unlike prior approaches limited to imitation learning, \gls{DQfD} combined \gls{td} updates with supervised classification of the demonstrator's actions, enabling it to learn from both demonstration data and self-generated data efficiently.
\textcite{mnih_asynchronous_2016} proposed \gls{a3c}, a reinforcement learning method that used parallel actor-learners to stabilise training. \gls{a3c} stabilised learning and reduced training time and resource requirements by using parallelism.
\textcite{schulman_proximal_2017} introduced \gls{ppo} algorithms, a family of policy gradient methods for \gls{rl} that balance between sample complexity, simplicity, and wall-time. The \gls{ppo} algorithms were tested on benchmark tasks, including simulated robotic locomotion and Atari game playing, outperforming other online policy gradient methods.

\section{Environments}
The \gls{ale} introduced by \textcite{bellemare_arcade_2013} is a platform for evaluating artificial intelligence agents using Atari 2600 game environments. \gls{ale} transformed each game into a standard \gls{rl} problem. It proposed a training and testing methodology for agents, becoming the de facto benchmark for \gls{drl} algorithms in discrete action spaces.
VizDoom, introduced by \textcite{Kempka2016ViZDoom}, is a research platform based on the classic first-person shooter game Doom. It allows for developing agents that operate based on visual inputs from the screen buffer and offers highly customisable options. As of 2023, VizDoom is part of the Farama Foundation, a non-profit organisation working to develop and maintain open-source \gls{rl} tools.
OpenAI Gym, developed by \textcite{brockman_openai_2016}, was a toolkit for developing and comparing \gls{rl} algorithms. It offered a comprehensive collection of benchmark environments, from simple control tasks to simulated robotic tasks and popular games from the \gls{ale}. It quickly became the standardised API for communication between learning algorithms and environments used in most research.
Gymnasium \autocite{towers_gymnasium_2023} is an open-source Python evolved from OpenAI's Gym designed for developing and comparing \gls{rl} algorithms. Gymnasium is currently maintained by the Farma Foundation and is the official replacement of the discontinued OpenAI Gym, the latter not receiving any future updates or bug fixes. \textcite{gymnasium_robotics2023github} is a version of Gymnasium for \gls{rl} in robotic environments.
The Minigrid and Miniworld libraries, introduced by \textcite{MinigridMiniworld23}, are libraries for \gls{rl} in goal-oriented tasks. These environments are structured as partially observable \glspl{mdp} and have been used in research for safe \gls{rl}, curiosity-driven exploration, and meta-learning.
\textcite{terry2021pettingzoo} introduced the PettingZoo library, an extension to Gymnasium designed to facilitate research of \gls{marl} algorithms. The library operates based on a novel Agent Environment Cycle games model to make \gls{marl} research more accessible and reproducible.

\section{Physics Object Manipulation}
\textcite{zhong_value_2013} presents a method for enhancing the efficiency of \gls{mpc} in solving optimal control problems, particularly in high-dimensional systems. The authors propose two strategies to derive a descriptive final cost function to aid \gls{mpc} in policy selection without extensive future planning or meticulous tuning of cost functions. The first strategy leverages the substantial data generated during \gls{mpc} simulations to learn the global optimal value function offline for use as a final cost. The second strategy directly solves the Bellman equation using aggregation methods for linearly-solvable \glspl{mdp} (LMDPs) to approximate the value function and the optimal policy. When integrated into the \gls{mpc} framework, this approach maintained controller performance quality even with a significantly shortened horizon, demonstrating its potential in real applications.
\textcite{bejjani_planning_2018} presented a method for real-time manipulation in cluttered environments using a Receding Horizon Planner integrated with a learned value function. The approach interleaves planning and execution in a closed-loop system to dynamically respond to changes during task execution, addressing the challenges posed by the uncertainty in modelling real-world physics.
Building on this, \textcite{bejjani_learning_2019} integrated end-to-end \gls{rl} and planning-based look-ahead to real-world manipulation tasks. The research leverages model-free \gls{drl} to train a \gls{cnn} and \gls{dnn} (CNN+DNN) that map visual observations to actions, one for pushing and the other for grasping, all using a Q-learning framework. The system demonstrated improved grasping success rates and efficiencies in picking experiments.
A separate study by \textcite{zeng_learning_2018} also explored the synergies between non-prehensile (pushing) and prehensile (grasping) actions in robotic manipulation. The authors used model-free \gls{drl} to train two fully convolutional networks (FCNs) that map from visual observations to actions, one for pushing and the other for grasping, in a Q-learning framework. The networks were self-supervised through trial and error, with rewards given for successful grasps, enabling the policy to learn pushing motions that facilitate future grasps and vice versa.
A paper by \textcite{song_multi-object_2020} addresses the planar non-prehensile sorting task where a robot must sort densely packed objects of different classes into distinct homogeneous clusters, optionally navigating around immovable obstacles. The task is solved using a \gls{mcts} algorithm, which is well-suited for high-dimensional state spaces and does not require explicit target states, utilising a discriminative function to identify goal states based on the relative positions of objects.

\chapter{Methodology}\label{chapter:methodology}

\section{Environment}

\begin{figure}[ht]
    \centering
    \includegraphics[width=0.35\textwidth]{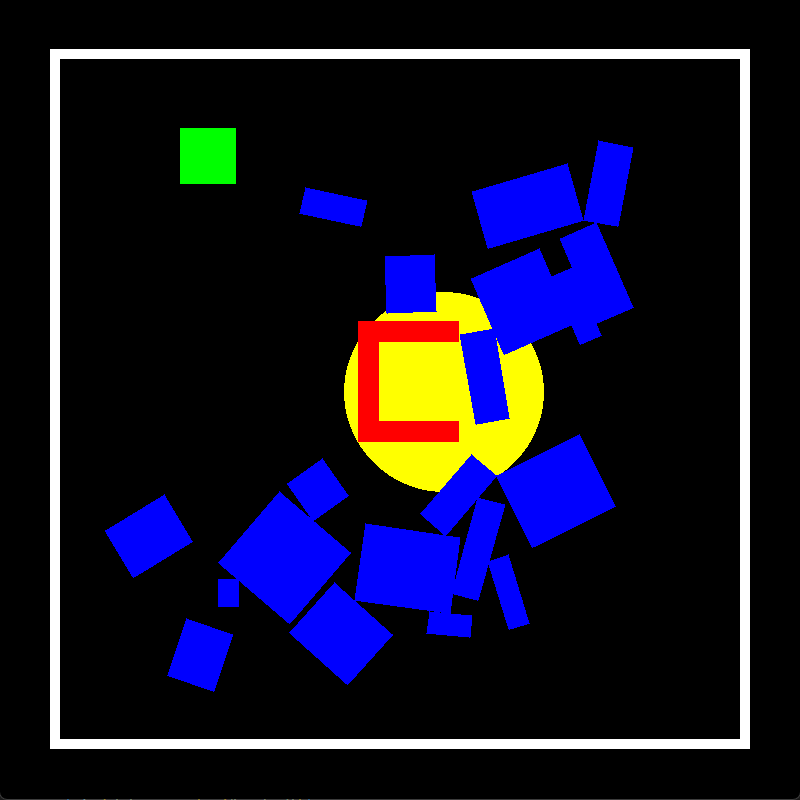}
    \caption[\gls{pbge} environment]{An initial state of \gls{pbge}. Pictured in blue is the clutter, the green square is the goal object, and the yellow circle is the target position.}
    \label{fig:env}
\end{figure}

\gls{pbge} is a custom environment that is compliant to the v0.26.2 version of the Gymnasium API standard for \gls{rl} \autocite{towers_gymnasium_2023}. The environment simulates a task-oriented scenario where a robotic gripper is manoeuvred to move a goal object to a target goal location. This simulation occurs in a two-dimensional space delineated by predefined borders and uses Pymunk to simulate 2D physics \autocite{blomqvist_pymunk_2023}. The task is episodic (\vref{sec:episodic_cont_task}), and every episode ends in either success, failure or \textit{truncation}. Failure conditions include the gripper or any other object going out of bounds, whereas success is achieved when the goal object is moved onto the target position. Truncation occurs when the episode terminates in neither success nor failure but because the episode takes too long and surpasses a predefined timestep limit. \cref{fig:env} graphically represents the environment.

The action space (\vref{sec:action_space}) is discrete and consists of four possible actions the agent can take at each step: move forward, move backwards, turn left and turn right. The observation space is defined as a raw $800 \times 800$ RGB image that represents the current simulator environment state, but changes described in \cref{sec:obs_preprocessing} are made such that the final agent observation is a $12 \times 84 \times 84$ vector of luminescence values.

The environment parameters are listed in \cref{tbl:env_params}. The \textit{seed} instantiates the numpy random generator \autocite{harris2020array} and is used for reproducibility. \textit{grayscale} and \textit{transpose} enable their respective observation preprocessing steps. \textit{noops} is the number of frames executed at every environment reset before the agent can take action. This is used to both allow the physic objects to come to rest after initial spawn and to populate the starting observation with the stacked frames. \textit{randomise} enables random spawn positions and angles for clutter objects, goal object, target position and gripper agent. \textit{randomise\_domain} enables full domain randomisation as described in \cref{sec:domain_random}. \textit{agent\_history\_len} controls the number of stacked frames used for observations. \textit{agent\_act\_repeat} controls the amount of "frame-skipping", meaning that the agent only takes action at intervals defined by agent\_act\_repeat, repeating the last action in the skipped frames. The simulation and physics simulation operate at a rate of 50 Frames Per Second, yielding an effective agent decision rate of $\frac{50}{\textit{agent\_act\_repeat}}$ FPS. This speeds up learning as the agent would otherwise have to make $50$ action selections per second. \textcite{mnih_human-level_2015} had a decision rate of around 15 FPS. \textit{clutter\_items} and \textit{clutter\_mass} define the number of clutter objects and the weight used for physics calculation. \textit{reward\_func} describes the reward evaluation function used to describe the agent goal \ref{sec:goal_reward}, and the possible functions are described in detail in \ref{sec:env_reward_funcs}. \textit{max\_timesteps} defines an upper bound on the maximum timesteps the agent can use to complete the task. This truncates episodes where the agent is "stuck", and nothing useful would be learnt. \textit{frictiton\_coeff} sets the ground friction coefficient for all object physics calculations.

\subsection{Agent-environment Interaction as an MDP} \label{sec:env_agent_inter_mdp}
We consider tasks where an agent interacts with \gls{pbge}, a process delineated in \vref{sec:agent_env_inter}. At each timestep, the agent chooses an action from the available actions $\A = \{\text{{left}}, \text{{right}}, \text{{forward}}, \text{{backward}}\}$. This chosen action is passed to the simulator and influences its internal state, which can also vary randomly due to the environment's stochastic nature. The agent does not observe the internal state. Instead, it observes an image denoted by $X_t \in \mathbb{R}^d$ that represents the current screen in terms of pixel values. Alongside, it receives a reward signal, noted as $R_t$, which represents the overarching goal.
It is important to note that this reward signal could be influenced by the entire history of actions and observations up to that point; feedback on a particular action might be \textit{delayed} by a substantial number of timesteps. Since the agent only sees the current screen, the environment is partially observable, and the agent can't fully grasp the existing circumstances from the image $X_t$ alone. This leads to a situation where many simulator states are perceptually aliased.
To counteract this, we feed sequences of actions and observations ($S_t \sim X_0,A_0,X_1,A_1,\dots,A_{t-1},X_t$) as input to the agent to facilitate learning strategies that take into account these series of inputs. Since all sequences are assumed to terminate in a finite number of timesteps, we create a large but finite \gls{mdp}, wherein each unique sequence represents a distinct state.
Therefore, we can apply standard reinforcement learning methods for \glspl{mdp} described in \vref{chapter:algos} by utilising the full series $S_t$ as the state representation at each timestep $t$. The agent's goal is to interact with the simulator by selecting actions to maximise future rewards and, by doing so, completing the goal.

\section{Observation Preprocessing} \label{sec:obs_preprocessing}
\begin{figure}[ht]
    \centering
    \includegraphics[width=0.95\textwidth]{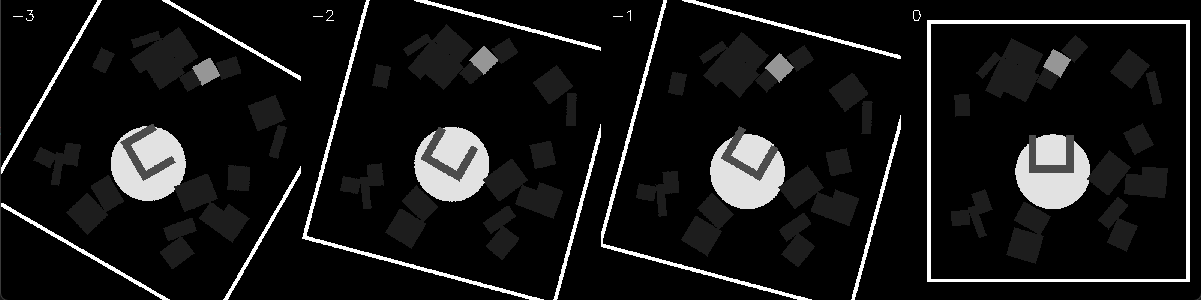} \label{fig:agent_obs}
    \caption[Agent observation]{Preprocessed and stacked frames forming the agent observation. The frames are ordered right to left, with the most recent frame being the rightmost one.}
\end{figure}
We first re-centre and rotate the input frame onto the gripper to obtain an \textit{agent-centric view} of the environment, meaning that the image tracks the robot from a top-view perspective. \textcite{bejjani_learning_2021-1} shows that this step significantly improves learning because it enables the agent to take advantage of the symmetries of the scene when compared to a fixed view. 
We then follow similar preprocessing steps to \textcite{mnih_playing_2013}, \textcite{mnih_human-level_2015}, which have been shown to facilitate learning and processing at the expense of some visual fidelity. Essential structures and patterns in the image are largely preserved.
First, we \textit{downsample} the high-resolution $3 \times 800 \times 800$ RGB input frame into a lower-resolution $3 \times 84 \times 84$ RGB output frame. We use a function that utilises a nearest-neighbour interpolation technique, which is computationally efficient but might introduce aliasing artefacts due to the omission of result sampling.
We then \textit{grayscale} the RGB observation by collapsing the colour information into a single channel. To do so, we extract the $Y$ channel, also known as luminance, from the RGB frame by combining the different colour channels with appropriate weights:
\[ Y \leftarrow 0.299R + 0.0587G + 0.114B. \]
resulting in a frame with dimensions $1 \times 84 \times 84$.
This still image would not be sufficient to determine the environment's full dynamics, including the velocities and directions of moving objects. To alleviate this problem, we frame stack the last four frames and use that as the agent observation. We make the observation more Markovian by incorporating some temporal information over a short period. The final dimensions are $4 \times 84 \times 84$. The result of all the preprocessing steps is shown in \cref{fig:agent_obs}.
Before feeding the observation into the neural network, we also normalise the luminescence values to fall within the $[0,1]$ range.

\section{Goal Formulation and Evaluation Metrics} \label{sec:env_reward_funcs}
\begin{figure}[ht]
    \centering
    \begin{minipage}[t]{.49\textwidth}
        \centering
        \includegraphics[width=0.67\linewidth]{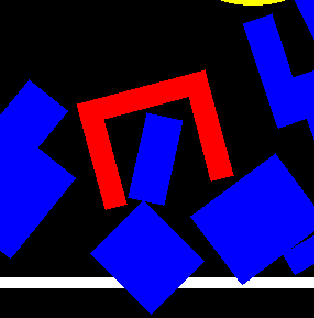}
        \captionof{figure}[Unsuccessful episode]{The agent pushes a clutter object past the boundary lines, failing the episode.}
        \label{fig:env_failure}
    \end{minipage}%
    \hfill
    \begin{minipage}[t]{.49\textwidth}
        \centering
        \includegraphics[width=0.7\linewidth]{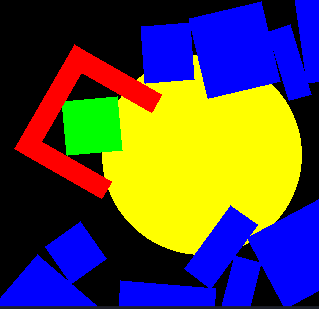}
        \captionof{figure}[Successful episode]{The agent successfully pushes the goal object onto the goal position.}
        \label{fig:env_success}
    \end{minipage}
\end{figure}
We formulate the goal of our task by defining our reward function (see \vref{sec:goal_reward}). The agent's goal should be to grasp the goal object and move it to the target position. We define as \textit{success} the states in which the goal object is positioned on the target position, as in \cref{fig:env_success}. We define \textit{failure} the states in which any object is outside the world boundaries, as in \cref{fig:env_failure}. We craft a diverse set of reward functions to achieve the goal:

\subsection{Binary Sparse Rewards}
This reward function is a classic reward signal that emits $1$ whenever the goal is accomplished, $-1$ when the task is failed and $0$ otherwise:
\begin{equation}
r(s, a, s') = 
\begin{cases} 
      1 & \text{if } s' \in \texttt{success}\\
      -1 & \text{if } s' \in \texttt{failure}\\
      0 & \text{otherwise}
\end{cases}
\end{equation} \label{eq:sparse_rewards}
In \gls{rl}, binary sparse rewards pose a significant challenge since the only measure of the agent's efficacy is the anticipated return. Sparse rewards do not guide the agent sufficiently, potentially preventing it from reaching states associated with positive rewards. Consequently, the agent might not improve performance, lacking a clear understanding of its objective.

\subsection{Reward Shaping}
\textit{Reward shaping} refers to the technique where we incorporate domain knowledge into the reward signal to make the \gls{rl} algorithms converge more quickly. In this case, we generally aim to reduce the distance between the agent and the goal object and between the goal object and the target position. We can incorporate this information (the domain knowledge) into our reward function to better guide our agent. We can also give a negative base reward at every timestep to encourage the agent to complete the task more quickly.

We define the following notation, where the numbers in the parenthesis are the default values:
\begin{itemize}[noitemsep,nolistsep,label={}]
\item \( r_{\text{fail}} \): the reward returned upon failure (100).
\item \( r_{\text{success}} \): the reward returned upon success (-100).
\item \( d_{\text{gt}} \): distance from gripper to target.
\item \( d_{\text{gtt}} \): distance from goal to target.
\item \( d_{\text{gt}}^{\text{prev}} \): previous distance from gripper to target.
\item \( d_{\text{gtt}}^{\text{prev}} \): previous distance from goal to target.
\item \( d_{\text{gt}}^{\text{init}} \): the initial distance from gripper to target.
\item \( d_{\text{gtt}}^{\text{init}} \): the initial distance from goal to target.
\item \( d_{\text{total}} = d_{\text{gt}} + d_{\text{gtt}} \)
\item \( d_{\text{total}}^{\text{prev}} = d_{\text{gt}}^{\text{prev}} + d_{\text{gtt}}^{\text{prev}} \)
\item \( d_{\text{total}}^{\text{init}} = d_{\text{gt}}^{\text{init}} + d_{\text{gtt}}^{\text{init}} \)
\end{itemize}
whereby \textit{previous distance}, we mean the distances computed at the last timestep unless otherwise specified, and by \textit{initial distance}, we mean the distances calculated at the beginning of the episode.

\subsubsection{Shaped Reward 1}
Here, we give a simple positive reward if the total distance decreases.
\begin{equation} \label{eq:dense_naive}
r(s,a,s') = 
\begin{cases} 
  r_{\text{fail}} & \text{if } s' \in \texttt{success} \\
  r_{\text{success}} & \text{if }  s' \in \texttt{failure} \\
  2 & \text{if } d_{\text{gt}} < d_{\text{gt}}^{\text{prev}} \text{ and } d_{\text{gtt}} < d_{\text{gtt}}^{\text{prev}} \\
  1 & \text{if } d_{\text{gt}} < d_{\text{gt}}^{\text{prev}} \text{ or } d_{\text{gtt}} < d_{\text{gtt}}^{\text{prev}} \\
  -1 & \text{otherwise}
\end{cases}
\end{equation}

\subsubsection{Shaped Reward 2}
Here, we use a predefined budget spread over the entire distance the agent has to close. In this function, the \textit{previous} distances save the shortest distances the agent has ever reached, meaning that the agent will not get a new reward by increasing and then decreasing the distances. We give a budget of $r_{\text{budget}} = 100$.
\begin{equation} \label{eq:rew_budget}
r(s,a,s') = 
\begin{cases} 
  r_{\text{fail}} & \text{if } s' \in \texttt{success} \\
  r_{\text{success}} & \text{if } s' \in \texttt{failure} \\
  (d_{\text{total}}^{\text{prev}} - d_{\text{total}}) \cdot \frac{r_{\text{budget}}}{d_{\text{total}}^{\text{init}}} & \text{if } d_{\text{total}} < d_{\text{total}}^{\text{prev}} \\
  0 & \text{otherwise}
\end{cases}
\end{equation}

\subsubsection{Shaped Reward 3}
We now define a complex reward clipped into the $\in [-1,1]$ range. We first define \( w_{\text{gt}}=0.15 \), the weight for the distance from gripper to target and \( w_{\text{gtt}}=0.15 \), the weight for the distance from goal to target.
\begin{equation} \label{eq:complex_shaping}
r(s, a, s') = 
\begin{cases} 
  r_{\text{fail}} & s' \in \texttt{success} \\
  r_{\text{success}} & \text{if } s' \in \texttt{failure} \\
  r_{\text{computed}} & \text{otherwise}
\end{cases}
\end{equation}
\( r_{\text{computed}} = \text{clip}\left(2 \cdot \left( \frac{r - r_{\text{min}}}{r_{\text{max}} - r_{\text{min}}} \right), -1, 1\right)\) where \( r \) is defined below as:
\[
r = -0.33 - 0.5 \cdot \text{notmoving} + w_{\text{gt}} (d_{\text{gt}}^{\text{best prev}} - d_{\text{gt}}) + w_{\text{gtt}} (d_{\text{gtt}}^{\text{best prev}} - d_{\text{gtt}})
\]
Where \textit{notmoving} is a variable that increments each time the gripper and goal remain stationary and resets to zero otherwise.

\subsection{Curriculum Learning}
In \textit{curriculum learning}, we start with a simplified version of the original task and gradually scale back to the standard settings as the agent performance improves. Here, we implement curriculum learning by reducing the initial spawn radius of the goal object and target position and making them closer to the agent. We also steadily increase the number of clutter objects, starting with only a few. Curriculum learning can be used in conjunction with reward shaping.

\subsection{Domain Randomisation} \label{sec:domain_random}
\begin{figure}[ht]
    \centering
    \includegraphics[width=0.95\textwidth]{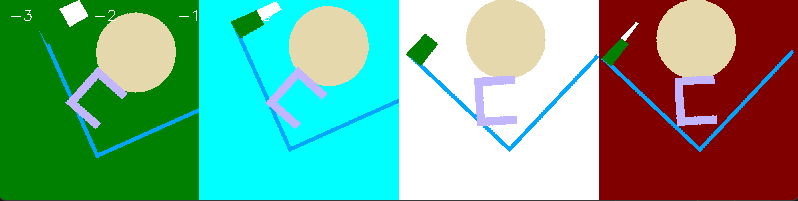} \label{fig:domain_rando}
    \caption[Domain randomisation]{Frames with different colours and environment parameters.}
\end{figure}
By \textit{domain randomisation}, we refer to a technique used to improve the generalization of a trained model to new, unseen environments. This is done by randomly varying aspects of the training environment, such as visual, physical, and even geometrical properties. Here, we use domain randomisation in several manners, including having random initial arrangements for every episode, randomising the number and size of clutter objects, changing up the colours of the environment and varying the world friction coefficient as shown in \cref{fig:domain_rando}. This should, in theory, make our models more resilient to potential noise and avoid overfitting.

\subsection{Evaluation Metrics} \label{sec:eval_metrics}
In our agent evaluations, we use two sets of metrics based on the environment type: dynamic and static.
For dynamic environments, where elements can change with every reset, we use metrics from 10 separate evaluation environments. We collect data every 10,000 timesteps to create the following charts:
\begin{enumerate}
    \item \textbf{Evaluation mean reward over time}: This is the mean episodic evaluation reward, averaged over 10 episodes. The shaded areas in the chart show the standard deviation.
    \item \textbf{Evaluation mean episode length over time}: This chart shows the average episodic evaluation length, found by averaging over 10 episodes.
    \item \textbf{Evaluation success rate over time}: This is the ratio of successful evaluation episodes to the total number of evaluation episodes, in this case, 10.
    \item \textbf{Evaluation efficiency over time}: This metric is the mean episodic evaluation reward divided by the mean episodic evaluation length, both averaged over 10 episodes.
\end{enumerate}
We also present tables that list the average episodic reward, average episodic length, and average efficiency at the final training checkpoint. These numbers are averaged over 100 episodes. In all charts and tables, we include a random agent as a \textit{BASELINE}, using statistics averaged over 100 episodes. All charts use a smoothing window of 2 to present the data clearly.

We use metrics from training rollouts for static environments, where everything is fixed and deterministic. Separate evaluation rollouts would show no difference and would be a waste of computational resources. We provide the following plots:
\begin{enumerate}
    \item \textbf{Training reward over time}: This shows the mean episodic training reward averaged over 100 episodes and the number of environment timesteps.
    \item \textbf{Training episode length over time}: This chart shows the mean episodic training length, averaged over 100 episodes and the number of timesteps.
    \item \textbf{Training efficiency over time}: This is calculated as the mean episodic training reward divided by the mean episodic length, both averaging over 100 episodes.
\end{enumerate}
We also present tables listing the average episodic reward, average episodic length, and average efficiency at the last training checkpoint for a single deterministic episode.

\section{Agent Architecture} \label{sec:netarch}
The overall agent architecture varies depending on the algorithms used, but we use the same feature extraction \gls{cnn} layers across all architectures. The input to the neural network consists of a $4 \times 84 \times 84$ image produced by the preprocessing, which is then processed as follows:
\begin{itemize}
    \item Convolutional Layer 1: $32$ filters, kernel size $8 \times 8$, stride $4$, followed by ReLU activation (rectifier nonlinearity). Input size is $12 \times 84 \times 84$, output size is $32 \times 20 \times 20$.
    \item Convolutional Layer 2: $64$ filters, kernel size $4 \times 4$, stride $2$, followed by ReLU activation. The input size is $32 \times 20 \times 20 $, and the output size is $64 \times 9 \times 9$.
    \item Convolutional Layer 3: $64$ filters, kernel size $3 \times 3$, stride $1$, followed by ReLU activation. Input size is $64 \times 9 \times 9$, output size is $64 \times 7 \times 7$, flattened to $3136$.
\end{itemize}
Figures \cref{fig:dqn_arch,fig:a2carch} show max-pooling layers equivalent to the strides listed here. We use PyTorch to build the \glspl{dnn} \autocite{paszke_pytorch_2019}.

\subsection{Deep Q-Network Architecture}
\begin{figure}[ht]
    \centering
    \includegraphics[width=1\textwidth]{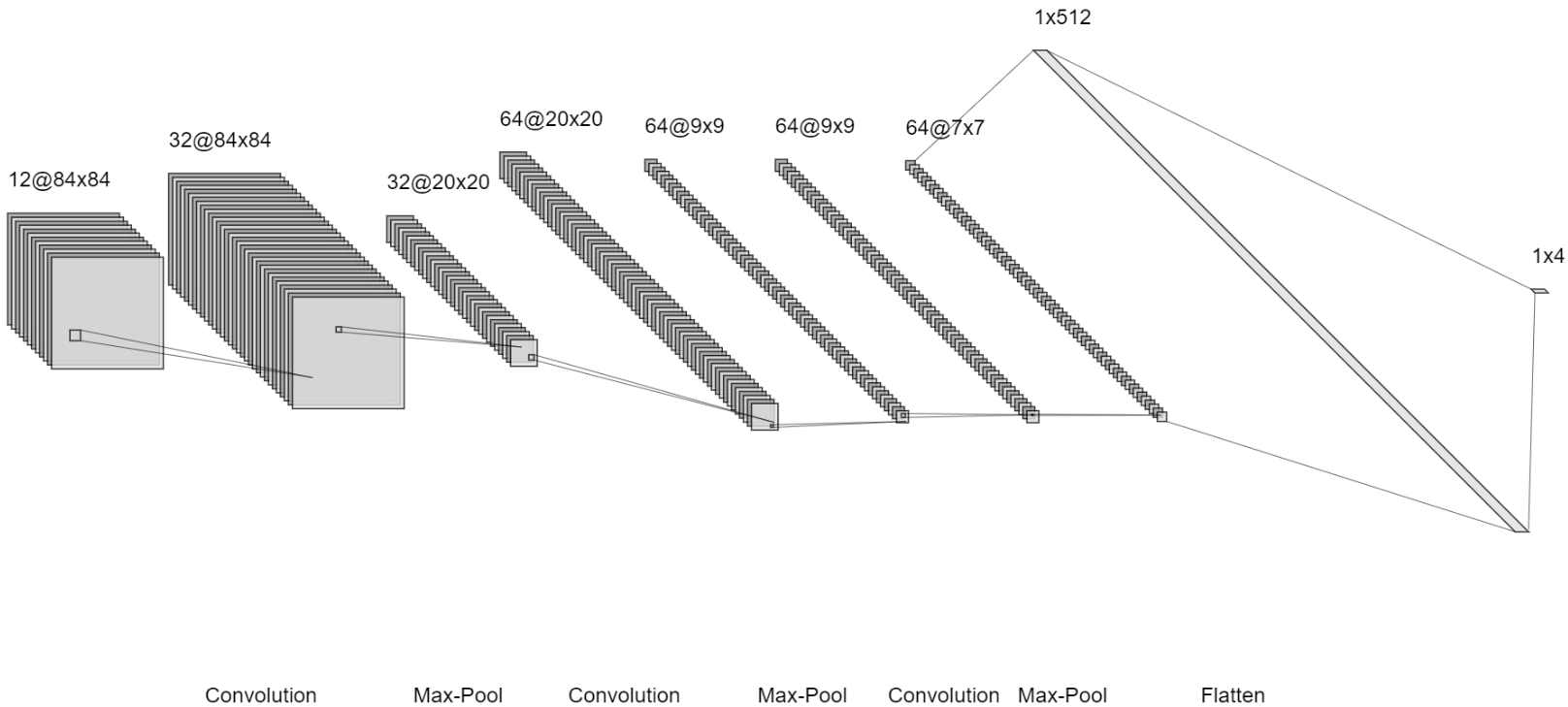}
    \caption[Network architecture for DQN]{Network architecture for \gls{dqn}. A \gls{cnn} is used for featured extraction and is followed by a hidden, fully connected layer.}
    \label{fig:dqn_arch}
\end{figure}

The \gls{dqn} agent architectures have an additional final hidden layer of $512$ fully connected rectifier units in addition to the layers listed in \cref{sec:netarch}. The output layer is a fully connected linear layer with a single output for each of the $4$ valid actions, where the final action is selected according to an $\epsilon$-greedy policy. As described in \vref{algo:dqn}, \gls{dqn} uses a copy of the Q-network to stabilise training. We use two copies of the full \gls{dnn} shown in \cref{fig:dqn_arch}.

\subsection{Actor-Critic Architecture}
\begin{figure}[ht]
    \centering
    \includegraphics[width=0.95\textwidth]{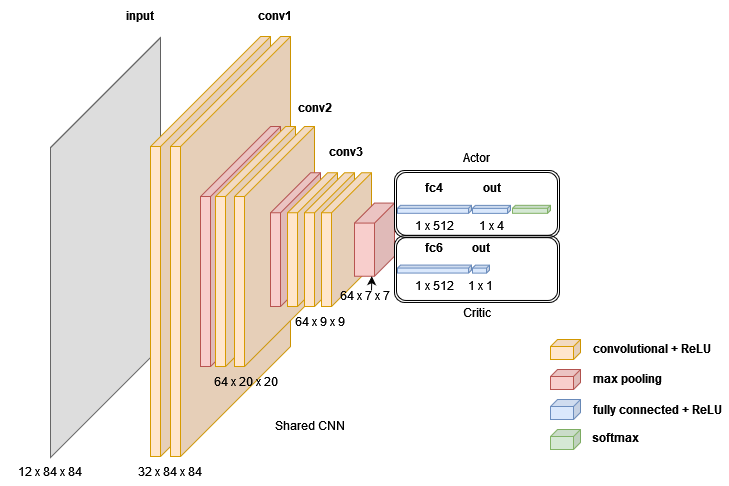}
    \caption[Network architecture for PPO and A2C]{Actor-critic network architecture used for both \gls{ppo} and \gls{a2c}. A \gls{cnn} is shared between the actor and critic for featured extraction, followed by separate fully connected layers for the actor and the critic.}
    \label{fig:a2carch}
\end{figure}

We use the same actor-critic architecture for both \gls{ppo} and \gls{a2c}. We feed the outputs of the convolutional layers described in \cref{sec:netarch} to two different fully connected layers of $512$ rectifier hidden units each, one for the actor and one for the critic. The actor connects the fully connected layer to an output layer of $4$ units, one for each action, and the action is then selected using softmax. The critic connects the fully connected layer to a single output unit representing the value of the state. The architecture is shown schematically in \cref{fig:a2carch}, and its motivation is discussed in \vref{sec:actor_critic}.

\section{Training and Evaluation}
For each experiment in \ref{chapter:experiments}, we train our agents for $2$ Million timesteps. Algorithm-specific training information is logged and monitored through Tensorboard, including metrics such as the learning rate, value loss, policy gradient loss, and current entropy coefficient. The models are evaluated every $10000$ timesteps with $5$ evaluation environments where the mean episode length, mean episodic reward and success rate are recorded and saved. The primary evaluation metric remains the mean episodic training reward, averaging over $100$ episodes. The mean episodic length, averaged over $100$ episodes, determines the agent's efficiency.
We retrain a different network for each experiment for every algorithm, but we keep the same hyperparameters to test for generalisation. Example training code is given in \cref{algo:train_code}.
\begin{algorithm}
\caption{Example code for training an agent on \gls{pbge}} \label{algo:train_code}
\begin{algorithmic}[1]
\Procedure{TrainAgent}{max\_episodes}
    \State $env \gets$ initialise the environment
    \State $agent \gets$ \Call{Agent.Init}{env.observation\_space, env.action\_space}
    
    \State $episode \gets 0$

    \While{$episode < max\_episodes$}
        \State $state \gets env.$\Call{Reset}{}
        \State $episodic\_reward \gets 0$
        \State $terminated \gets \text{False}$
        
        \While{not $terminated$}
            \State $action \gets agent.$\Call{SelectAction}{$state$}
            \State $next\_state, reward, terminated, \text{\textit{info}} \gets env.$\Call{Step}{$action$}
            \State $agent.$\Call{StoreTransition}{$state, action, reward, next\_state, done$}
            \State $agent.$\Call{OptimizeModel}{}
            \State $state \gets next\_state$
            \State $total\_reward \gets total\_reward + reward$
        \EndWhile
        \State \textbf{print} "Episode: ", $episode$, "Total Reward: ", $total\_reward$
        \State $episode \gets episode + 1$
    \EndWhile
\EndProcedure
\end{algorithmic}
\end{algorithm}

\subsection{Algorithms and Environment Parameters}
We give tables for the hyperparameters used for all experiments. A non-exhaustive hyperparameter search was conducted using Optuna \autocite{akiba_optuna_2019}. For details about how the hyperparameters affect their respective algorithms, see \gls{dqn} in \vref{sec:dqn}, \gls{a2c} in \vref{sec:a2c} and \gls{ppo} in \vref{sec:ppo}.
\begin{table}[tb]
    \caption{DQN Hyperparameters} \label{tbl:dqn_hyperparams}
    \begin{tabular}{ p{4.5cm}  p{1.7cm}  p{6cm} }
        \toprule
        \textbf{Parameter} & \textbf{Value} & \textbf{Description} \\
        \midrule
        learning\_rate & \(1 \times 10^{-4}\) & Learning rate \\
        buffer\_size & \(1,000,000\) & Size of the replay buffer \\
        learning\_starts & \(50,000\) & Steps before learning starts \\
        batch\_size & \(32\) & Minibatch size for gradient update \\
        gamma & \(0.99\) & Discount factor \\
        train\_freq & \(4\) & Training frequency \\
        gradient\_steps & \(1\) or \(-1\) & Number of gradient steps after each rollout \\
        target\_update\_interval & \(10,000\) & Target network update interval \\
        exploration\_fraction & \(0.1\) & Fraction of training period for exploration \\
        exploration\_initial\_eps & \(1.0\) & Initial exploration probability \\
        exploration\_final\_eps & \(0.05\) & Final exploration probability \\
        max\_grad\_norm & \(10\) & Maximum gradient clipping value \\
        \bottomrule
    \end{tabular}
\end{table}
\begin{table}[tb]
    \caption{PPO Hyperparameters}
    \begin{tabular}{ p{4.5cm}  p{1.7cm}  p{6cm} }
        \toprule
        \textbf{Parameter} & \textbf{Value} & \textbf{Description} \\
        \midrule
learning\_rate & \(3 \times 10^{-4}\) & Learning rate \\
n\_steps & \(2048\) & Number of steps per update \\
batch\_size & \(64\) & Minibatch size \\
n\_epochs & \(10\) & Number of epochs for loss optimization \\
gamma & \(0.99\) & Discount factor \\
gae\_lambda & \(0.95\) & GAE trade-off factor \\
clip\_range & \(0.2\) & Clipping parameter \\
clip\_range\_vf & None & Value function clipping parameter \\
normalize\_advantage & True & Normalize advantage or not \\
ent\_coef & \(0.0\) & Entropy coefficient \\
vf\_coef & \(0.5\) & Value function coefficient \\
max\_grad\_norm & \(0.5\) & Maximum gradient clipping value \\
        \bottomrule
    \end{tabular}
\end{table}
\begin{table}[tb]
    \caption{A2C Hyperparameters}
    \begin{tabular}{ p{4.5cm}  p{1.7cm}  p{6cm} }
        \toprule
        \textbf{Parameter} & \textbf{Value} & \textbf{Description} \\
        \midrule
learning\_rate & \(7 \times 10^{-4}\) & Learning rate \\
n\_steps & \(5\) & Number of steps per update \\
gamma & \(0.99\) & Discount factor \\
gae\_lambda & \(1.0\) & GAE trade-off factor \\
ent\_coef & \(0.0\) & Entropy coefficient \\
vf\_coef & \(0.5\) & Value function coefficient \\
max\_grad\_norm & \(0.5\) & Maximum gradient clipping value \\
rms\_prop\_eps & \(1 \times 10^{-5}\) & RMSProp epsilon \\
normalize\_advantage & False & Normalize advantage or not \\
        \bottomrule
    \end{tabular}
\end{table}
\begin{table}[tb]
    \caption{Description of environment parameters} \label{tbl:env_params}
    \begin{tabular}{ p{4.5cm}  p{1.7cm}  p{6cm} }
        \toprule
        \textbf{Parameter} & \textbf{Type} & \textbf{Brief description} \\
        \midrule
        seed & integer & Sets the seed for the environment \\
        grayscale & boolean & Enables grayscaled observations \\
        transpose & boolean & Transposes the observation onto the gripper point of view \\
        noops & integer & Number of noops executed on environment reset \\
        randomise & boolean & Enables spawn positions randomisation \\
        randomise\_domain & boolean & Enables full domain randomisation \\
        agent\_history\_len & integer & Number of stacked frames \\
        agent\_act\_repeat & integer & Controls the agent decision rate \\
        agent\_speed & float & Agent speed \\
        agent\_ang\_speed & float & Agent angular speed \\
        clutter\_items & integer & Number of clutter items \\
        clutter\_mass & float & Mass of the clutter objects \\
        reward\_func & callable & Specifies the reward evaluation function \\
        max\_timesteps & integer & Maximum timesteps before the environment is truncated \\
        friction\_coeff & float & Ground friction coefficient \\
        \bottomrule
    \end{tabular}
\end{table}
\chapter{Experiments}\label{chapter:experiments}
In this chapter, we will show experiments whose results are meaningful and add value to the discussion. Many other preliminary experiments have been conducted and discarded, often yielding predictable and uninteresting results. In \cref{table:default_params}, we give the default environment parameters, which will be used in the experiments unless explicitly stated otherwise.

\begin{table}[ht] \label{table:default_params}
    \centering
    \caption[Environment parameters for experiments]{Global Environment Parameters}
    \begin{tabular}{ p{4.5cm}  p{3cm}}
        \toprule
        \textbf{Parameter} & \textbf{Default Value} \\
        \midrule
        grayscale & True \\
        transpose & True \\
        noops & 50 \\
        randomise & True \\
        randomise\_domain & False \\
        agent\_history\_len & 4 \\
        agent\_act\_repeat & 4 \\
        agent\_speed & 300 \\
        agent\_ang\_speed & 4.91 \\
        clutter\_items & 10 \\
        clutter\_mass & 1.0 \\
        max\_timesteps & 300 \\
        friction\_coeff & 0.2 \\
        \bottomrule
    \end{tabular}
\end{table}

\section{Experiment \RN{1}} \label{sec:exp1}
In this experiment, we use the static environment setup and the sparse reward function shown in \vref{eq:sparse_rewards}.
\begin{table}[ht]
    \centering
    \caption[Environment parameters for experiment I]{Environment Parameters}
    \begin{tabular}{p{5.5cm}p{4.5cm}}
        \toprule
        \textbf{Parameter} & \textbf{Value} \\
        \midrule
        seed & 756765 \\
        randomise & False \\
        clutter\_items & 1 \\
        \bottomrule
    \end{tabular}
    \vspace{1em}
    
    \caption[Baseline performance for experiment I]{Baseline Performance}
    \begin{tabular}{p{5.5cm}p{4.5cm}}
        \toprule
        \textbf{Metric} & \textbf{Value} \\
        \midrule
        Mean Reward & -0.98 $\pm$ 0.14\\
        Mean Length & 70.20 \\
        Mean Efficiency & -0.014 \\
        Success Rate & 0.01 \\
        \bottomrule
    \end{tabular}
    \vspace{1em}
    
    \caption[Final performance for experiment I]{Final Performance} \label{tbl:exp1_models}
    \begin{tabular}{p{3cm}p{1.5cm}p{1.5cm}p{1.5cm}p{1.5cm}}
        \toprule
        \textbf{Metric} & \textbf{PPO} & \textbf{A2C} & \textbf{DQN} & \textbf{DQN} \\
        \midrule
        Timesteps & 1,000,000 & 1,000,000 & 1,000,000 & 900,000 \\
        Reward & 0.00 & 0.00 & -1.00 & 0.00 \\
        Length & 300 & 300 & 28 & 300\\
        Efficiency & 0 & 0 & -0.04 & 0\\
        Success Rate & 0 & 0 & 0 & 0\\
        \bottomrule
    \end{tabular}
\end{table}
\begin{landscape}
    \begin{figure}
        \begin{subfigure}{0.75\textwidth}
            \includegraphics[width=\textwidth]{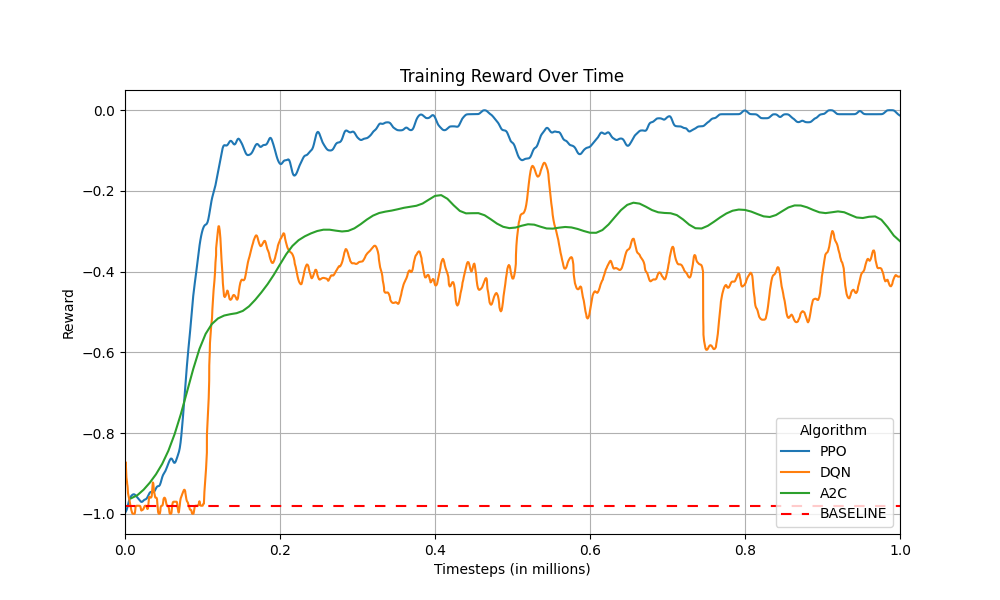}
            \caption{Training reward over time.}
        \end{subfigure}%
        \begin{subfigure}{0.75\textwidth}
            \includegraphics[width=\textwidth]{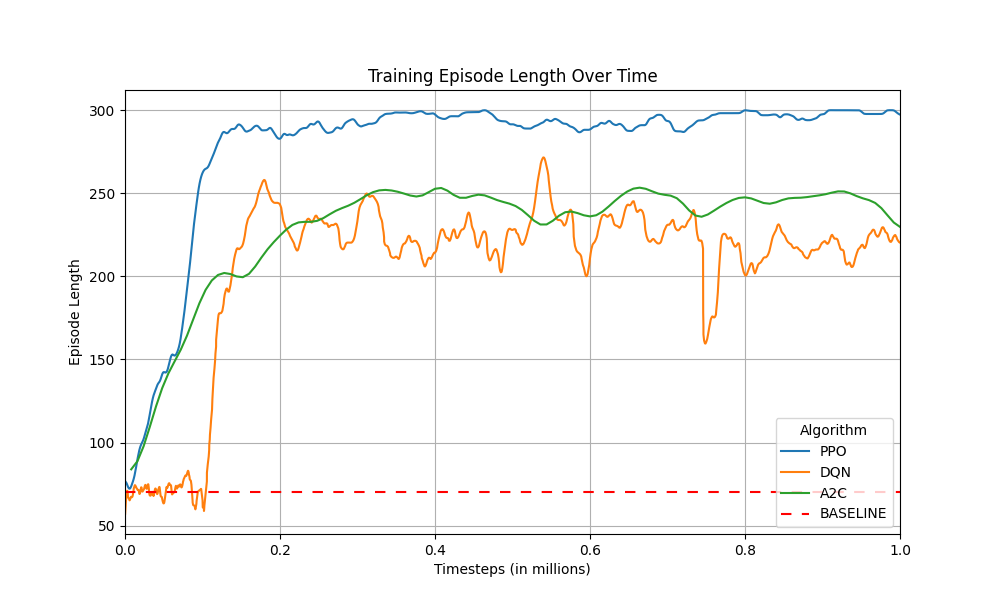}
            \caption{Training episode length over time.}
        \end{subfigure}
        \centering
        \begin{subfigure}{0.75\textwidth}
            \includegraphics[width=\textwidth]{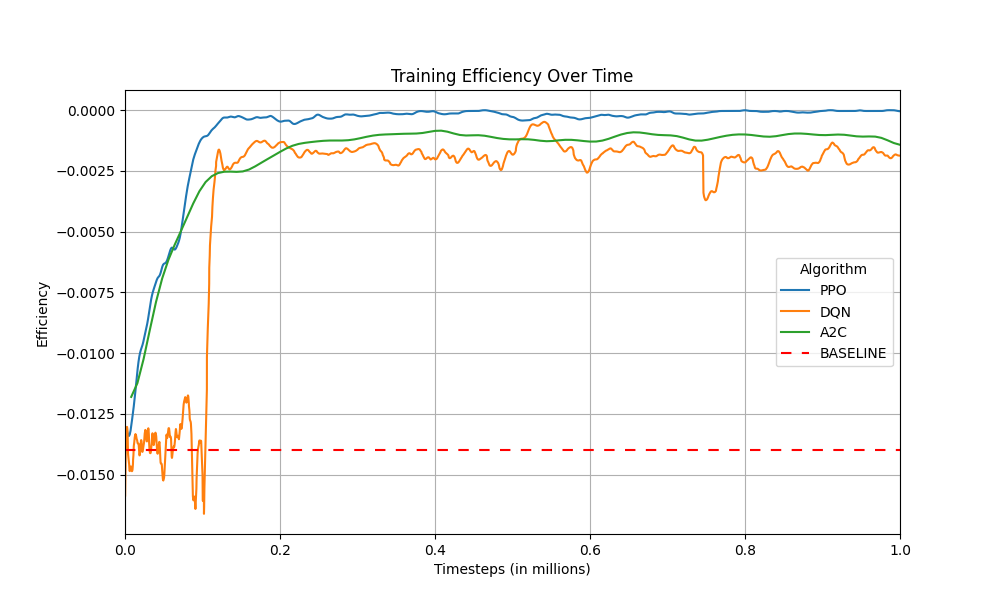}
            \caption{Training efficiency over time.}
        \end{subfigure}
        
        \caption{Agent performance in experiment \RN{1}.} \label{fig:experiment1}
    \end{figure}
\end{landscape}
\subsection{Results}
The graphs in the graphs in \cref{fig:experiment1} show that the agent fails to learn how to solve the task, as they never get to see the positive reward for successful completion. In the final models shown in \ref{tbl:exp1_models}, \gls{ppo} learns to rotate in place until the episode terminates continuously, and \gls{a2c} has a similar behaviour. The \gls{dqn} agent fails the task at the latest checkpoint but behaves similarly to the other at $900000$ timesteps. The agents all perform better than the baseline.

\section{Experiment \RN{2}}
In this experiment, we use the static environment setup and seed of \cref{sec:exp1} but use the naive dense reward function shown in \vref{eq:dense_naive}.
\begin{table}[ht]
    \centering
    \caption[Environment parameters for experiment II]{Environment Parameters}
    \begin{tabular}{p{4.5cm}p{4cm}}
        \toprule
        \textbf{Parameter} & \textbf{Value} \\
        \midrule
        seed & 756765 \\
        randomise & False \\
        clutter\_items & 1 \\
        \bottomrule
    \end{tabular}
    \vspace{1em}
    
    \caption[Baseline performance for experiment II]{Baseline Performance}
    \begin{tabular}{p{4.5cm}p{4cm}}
        \toprule
        \textbf{Metric} & \textbf{Value} \\
        \midrule
        Mean Reward & -88.62 $\pm$ 31.85 \\
        Mean Length & 23.68 \\
        Mean Efficiency & -3.74 \\
        Success Rate & 0.00 \\
        \bottomrule
    \end{tabular}
    \vspace{1em}
    
    \caption[Final performance for experiment II]{Final Performance} \label{tbl:exp2}
    \begin{tabular}{p{3cm}p{1.5cm}p{1.5cm}p{1.5cm}p{1.5cm}}
        \toprule
        \textbf{Metric} & \textbf{PPO} & \textbf{A2C} & \textbf{DQN} \\
        \midrule
        Timesteps & 1,000,000 & 1,000,000 & 1,000,000 \\
        Reward & 122.0 & -78.0 & 404.0 \\
        Length & 24 & 15 & 30 \\
        Efficiency & 5.08 & -5.2 & 13.47 \\
        Success Rate & 1 & 0 & 0 \\
        \bottomrule
    \end{tabular}
\end{table}
\begin{landscape}
    \begin{figure}
        \begin{subfigure}{0.75\textwidth}
            \includegraphics[width=\textwidth]{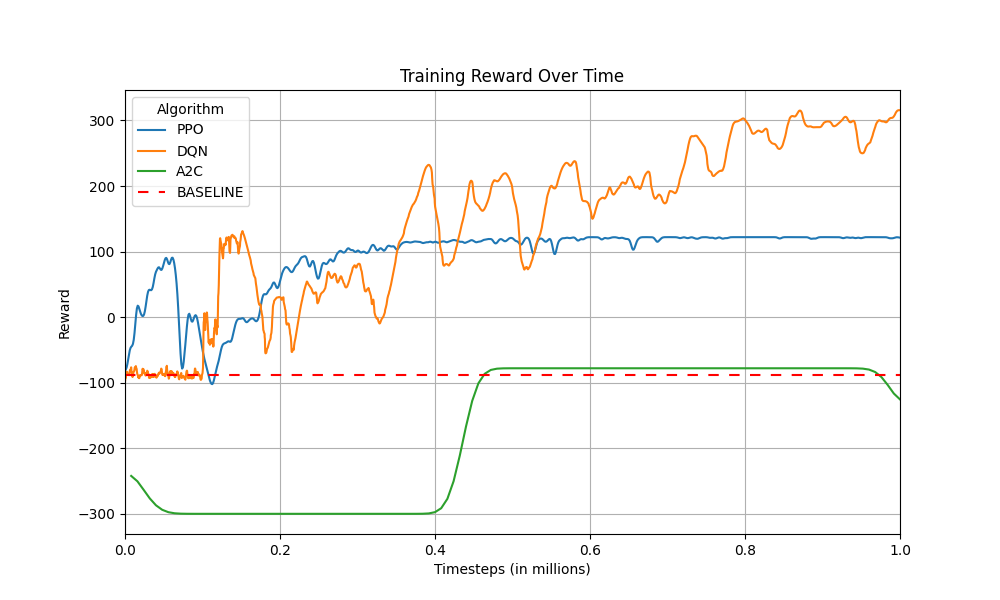}
            \caption{Training reward over time.}
        \end{subfigure}%
        \begin{subfigure}{0.75\textwidth}
            \includegraphics[width=\textwidth]{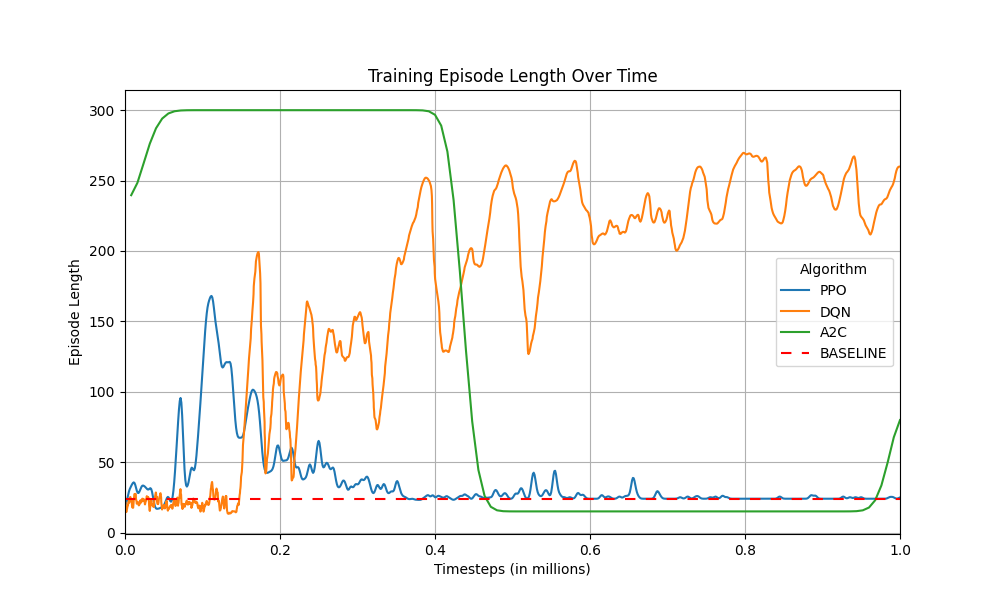}
            \caption{Training episode length over time.}
        \end{subfigure}
        \centering
        \begin{subfigure}{0.75\textwidth}
            \includegraphics[width=\textwidth]{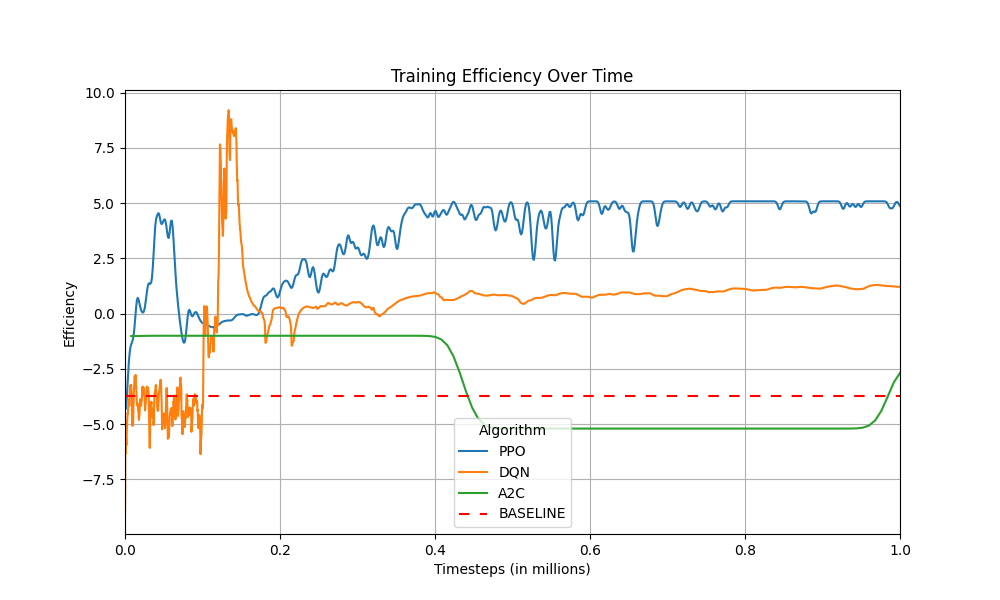}
            \caption{Training efficiency over time.}
        \end{subfigure}
        \caption{Agent performance in experiment \RN{2}.} \label{fig:experiment2}
    \end{figure}
\end{landscape}
\subsection{Results}
Looking at the charts in \cref{fig:experiment2}, it appears \gls{dqn} performs best. However, that is not the case. \cref{tbl:exp2} shows how despite achieving $404$ of mean episodic reward, its success rate is $0$. This is because the agent has learnt to exploit our naive reward function and keeps moving back and forth to accumulate reward instead of completing the real goal. On the other hand, \gls{ppo}, who initially appears to have lower performance, actually learned to solve the task. \gls{a2c} behaves similarly to experiment \RN{1}.

\section{Experiment \RN{3}}
We start examining fully randomised environments. We use the budgeted reward function defined in \vref{eq:rew_budget}.

\begin{table}[ht]
    \centering
    \caption[Environment parameters for experiment III]{Environment Parameters}
    \begin{tabular}{p{7cm}p{5.5cm}}
        \toprule
        \textbf{Parameter} & \textbf{Value} \\
        \midrule
        seed & 934612 \\
        randomise & True \\
        clutter\_items & 3 \\
        \bottomrule
    \end{tabular}
    \vspace{1em}
    
    \caption[Baseline performance for experiment III]{Baseline Performance}
    \begin{tabular}{p{7cm}p{5.5cm}}
        \toprule
        \textbf{Metric} & \textbf{Value} \\
        \midrule
        Mean Reward & -97.82 $\pm$ 4.18 \\
        Mean Length & 16.44 \\
        Mean Efficiency & -5.95 \\
        Success Rate & 0.00 \\
        \bottomrule
    \end{tabular}
    \vspace{1em}
    
    \caption[Final performance for experiment III]{Final Performance} \label{tbl:exp3}
    \begin{tabular}{p{3cm}p{3cm}p{3cm}p{3cm}}
        \toprule
        \textbf{Metric} & \textbf{PPO} & \textbf{A2C} & \textbf{DQN} \\
        \midrule
        Timesteps & 1,000,000 & 1,000,000 & 1,000,000 \\
        Mean Reward & 6.00 $\pm$ 7.93 & 6.87 $\pm$ 7.56 & 41.43 $\pm$ 66.34 \\
        Mean Length & 300.0 $\pm$ 0.0 & 300.0 $\pm$ 0.0 & 202.62 $\pm$ 135.76 \\
        Mean Efficiency & 0.02 & 0.02 & 0.20 \\
        Success Rate & 0.00 & 0.00 & 0.22 \\
        \bottomrule
    \end{tabular}
\end{table}
\begin{landscape}
    \begin{figure}
        \begin{subfigure}{0.75\textwidth}
            \includegraphics[width=\textwidth]{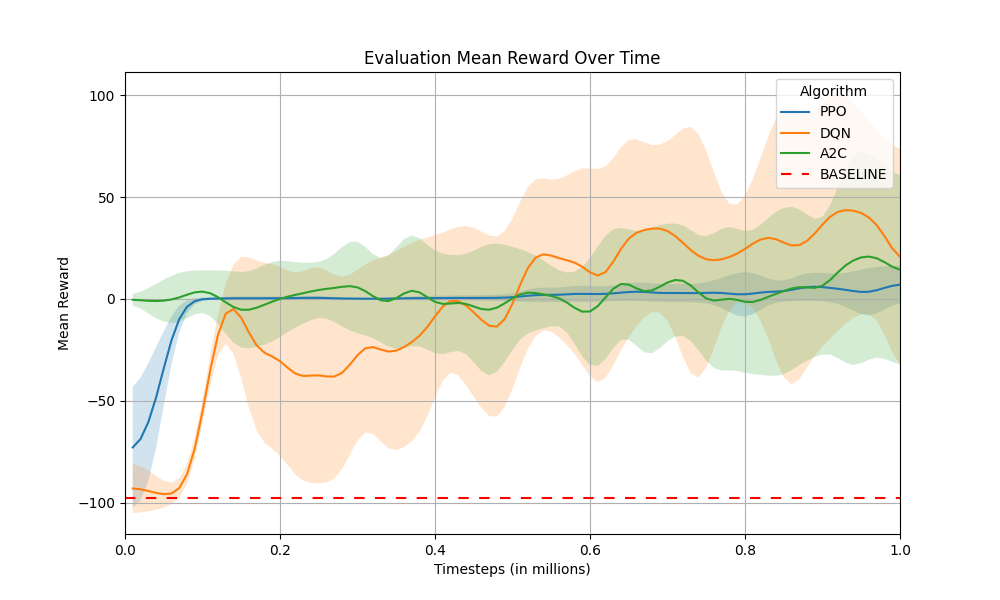}
            \caption{Evaluation mean reward over time.}
        \end{subfigure}%
        \begin{subfigure}{0.75\textwidth}
            \includegraphics[width=\textwidth]{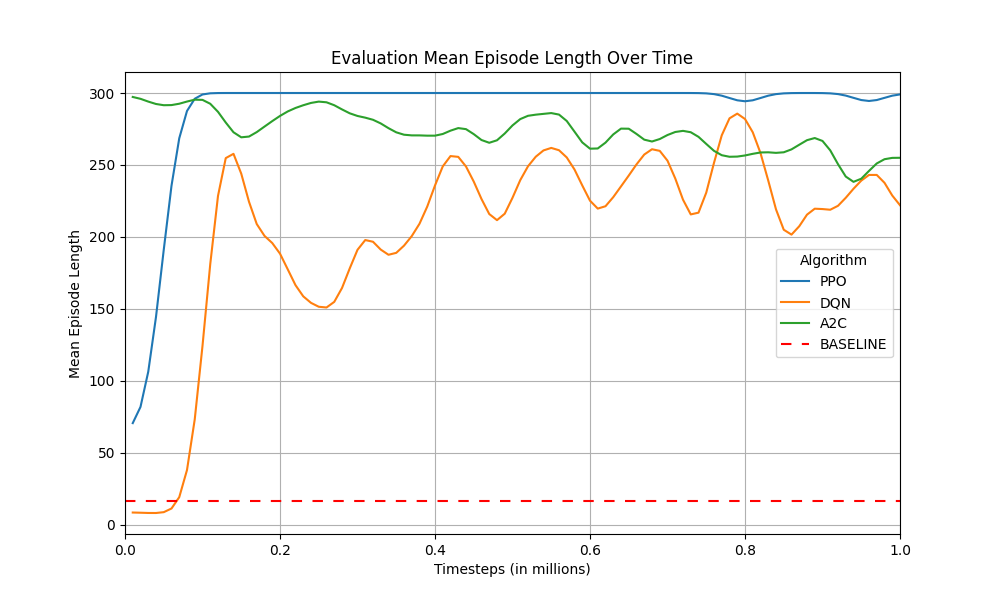}
            \caption{Evaluation mean episode length over time.}
        \end{subfigure}
        \begin{subfigure}{0.75\textwidth}
            \includegraphics[width=\textwidth]{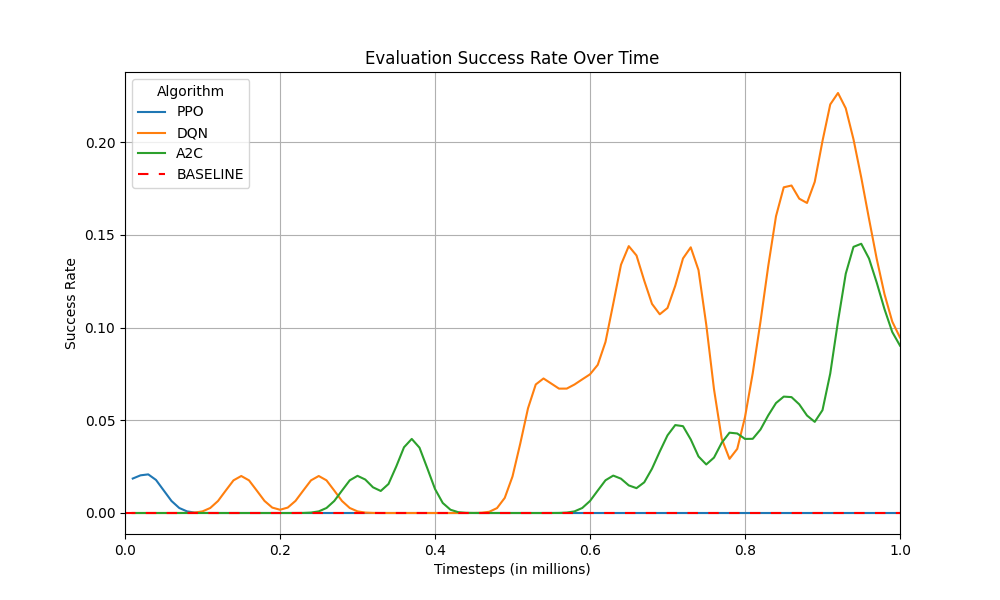}
            \caption{Evaluation success rate over time.}
        \end{subfigure}%
        \begin{subfigure}{0.75\textwidth}
            \includegraphics[width=\textwidth]{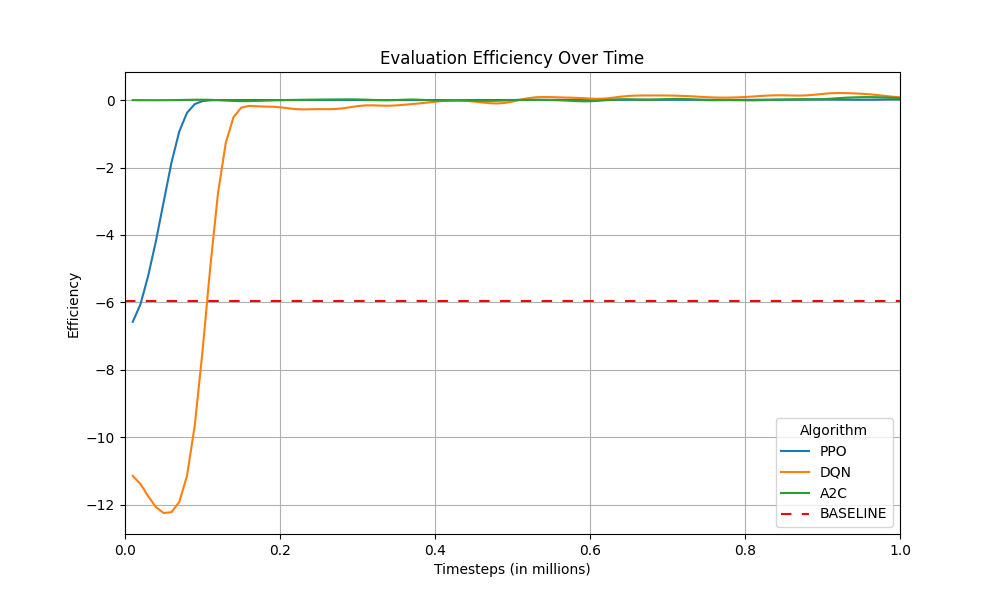}
            \caption{Evaluation efficiency over time.}
        \end{subfigure}
        \caption{Agent performance in experiment \RN{3}.} \label{fig:exp3}
    \end{figure}
\end{landscape}

\subsection{Results}
We see from plots in \cref{fig:exp3} that the reward function seems more representative of agent performance this time. \gls{ppo} and \gls{a2c} have again converged to the local optimum of rotating in place until episode termination to avoid any negative reward.
\gls{dqn} in \cref{tbl:exp3} only rarely achieves the goal.

\section{Experiment \RN{4}}
We try using domain randomisation along with the complex reward function in \vref{eq:complex_shaping} to guide the agent.

\begin{table}[ht]
    \centering
    \caption[Environment parameters for experiment IV]{Environment Parameters}
    \begin{tabular}{p{7cm}p{5.5cm}}
        \toprule
        \textbf{Parameter} & \textbf{Value} \\
        \midrule
        seed & 467328 \\
        randomise & True \\
        randomise\_domain & True \\
        clutter\_items & 3 \\
        \bottomrule
    \end{tabular}
    \vspace{1em}
    
    \caption[Baseline performance for experiment IV]{Baseline Performance}
    \begin{tabular}{p{7cm}p{5.5cm}}
        \toprule
        \textbf{Metric} & \textbf{Value} \\
        \midrule
        Mean Reward & -126.55 $\pm$ 23.03 \\
        Mean Length & 80.12 \\
        Mean Efficiency & -1.58 \\
        Success Rate & 0.00 \\
        \bottomrule
    \end{tabular}
    \vspace{1em}
    
    \caption[Final performance for experiment IV]{Final Performance} \label{tbl:exp4}
    \begin{tabular}{p{3cm}p{3cm}p{3cm}p{3cm}}
        \toprule
        \textbf{Metric} & \textbf{PPO} & \textbf{A2C} & \textbf{DQN} \\
        \midrule
        Timesteps & 1,000,000 & 1,000,000 & 1,000,000 \\
        Mean Reward & -249.27 $\pm$ 94.82 & -285.42 $\pm$ 20.32 & -92.67 $\pm$ 40.41 \\
        Mean Length & 261.13 $\pm$ 96.40 & 300.0 $\pm$ 0.0 & 180.48 $\pm$ 138.09 \\
        Mean Efficiency & -0.95 & -0.95 & -0.51 \\
        Success Rate & 0.05 & 0.0 & 0.04 \\
        \bottomrule
    \end{tabular}
\end{table}

\begin{landscape}
    \begin{figure}
        \begin{subfigure}{0.75\textwidth}
            \includegraphics[width=\textwidth]{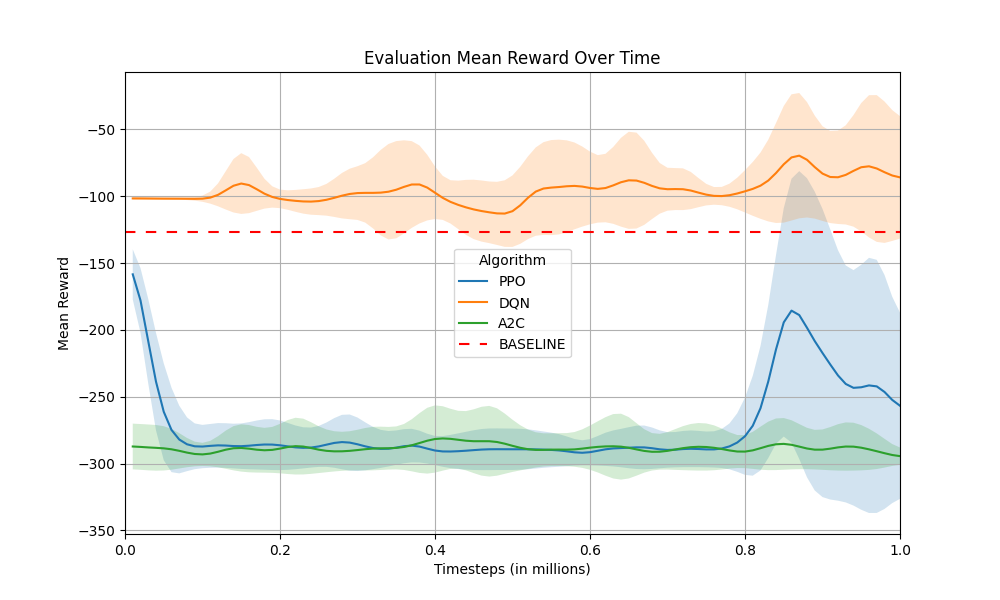}
            \caption{Evaluation mean reward over time.}
        \end{subfigure}%
        \begin{subfigure}{0.75\textwidth}
            \includegraphics[width=\textwidth]{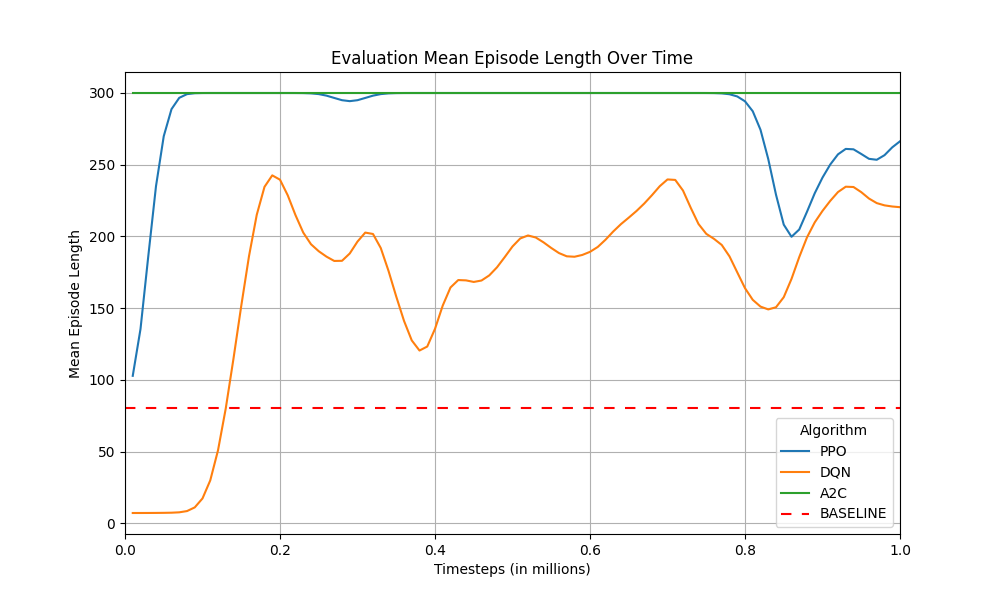}
            \caption{Evaluation mean episode length over time.}
        \end{subfigure}
        \begin{subfigure}{0.75\textwidth}
            \includegraphics[width=\textwidth]{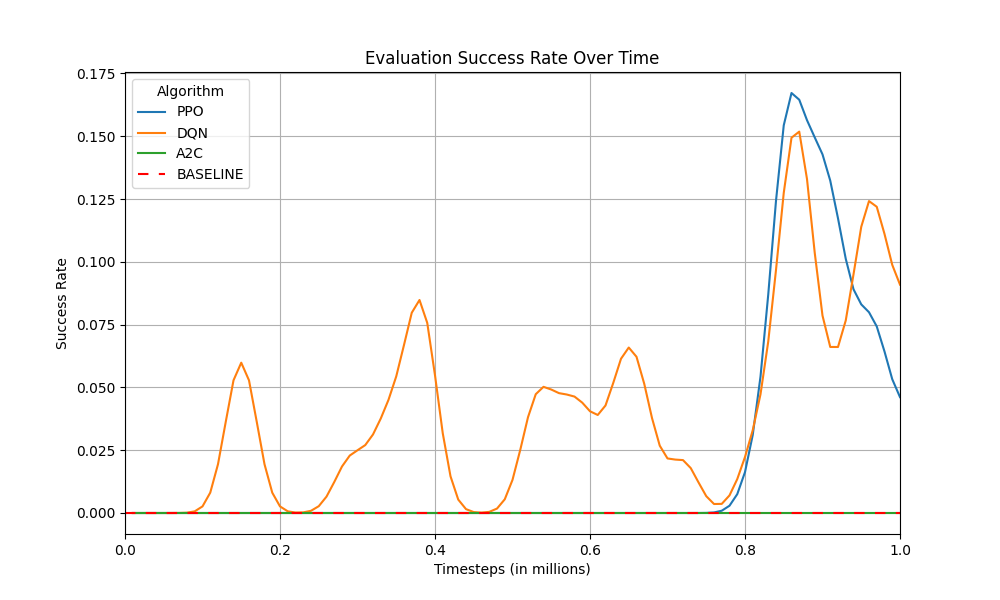}
            \caption{Evaluation success rate over time.}
        \end{subfigure}%
        \begin{subfigure}{0.75\textwidth}
            \includegraphics[width=\textwidth]{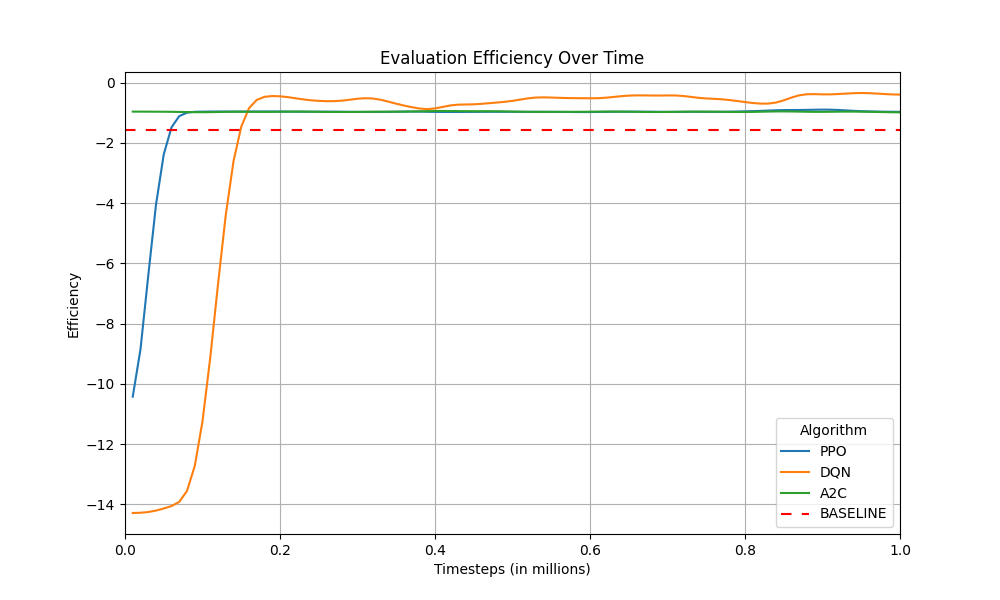}
            \caption{Evaluation efficiency over time.}
        \end{subfigure}
        \caption{Agent performance in experiment \RN{4}.} \label{fig:exp4}
    \end{figure}
\end{landscape}

\subsection{Results}
\cref{fig:exp4} and \cref{tbl:exp4} show that the random baseline performs better than the agents, with \gls{dqn} only marginally better. This is probably caused by the domain randomisation, making it impossible for the agent to distinguish the goal object from the clutter.

\section{Experiment \RN{5}}
We revisit \vref{eq:sparse_rewards} with a slight modification. At every timestep, the reward is $-1$ instead of $0$, and $+1$ for successfully completing the episode.

\begin{table}[ht]
    \centering
    \caption[Environment parameters for experiment V]{Environment Parameters}
    \begin{tabular}{p{7cm}p{5.5cm}}
        \toprule
        \textbf{Parameter} & \textbf{Value} \\
        \midrule
        seed & 115545 \\
        randomise & True \\
        clutter\_items & 1 \\
        \bottomrule
    \end{tabular}
    \vspace{1em}
    \centering
    \caption[Baseline performance for experiment V]{Baseline Performance}
    \begin{tabular}{p{7cm}p{5.5cm}}
        \toprule
        \textbf{Metric} & \textbf{Value} \\
        \midrule
        Mean Reward & $-90.36 \pm 69.73$ \\
        Mean Length & $90.36$ \\
        Mean Efficiency & $-1.0$ \\
        Success Rate & $0.0$ \\
        \bottomrule
    \end{tabular}
    \vspace{1em}
    \caption[Final performance for experiment V]{Final Performance}
    \begin{tabular}{p{3cm}p{3cm}p{3cm}p{3cm}}
        \toprule
        \textbf{Metric} & \textbf{PPO} & \textbf{A2C} & \textbf{DQN} \\
        \midrule
        Timesteps & $1,000,000$ & $1,000,000$ & $1,000,000$ \\
        Mean Reward & $-8.3 \pm 1.82$ & $-300.0 \pm 0.0$ & $-5.89 \pm 1.84$ \\
        Mean Length & $8.3 \pm 1.82$ & $300.0 \pm 0.0$ & $5.89 \pm 1.84$ \\
        Mean Efficiency & $-1.0$ & $-1.0$ & $-1.0$ \\
        Success Rate & $0.0$ & $0.0$ & $0.0$ \\
        \bottomrule
    \end{tabular}
\end{table}
\begin{landscape}
    \begin{figure}
        \begin{subfigure}{0.75\textwidth}
            \includegraphics[width=\textwidth]{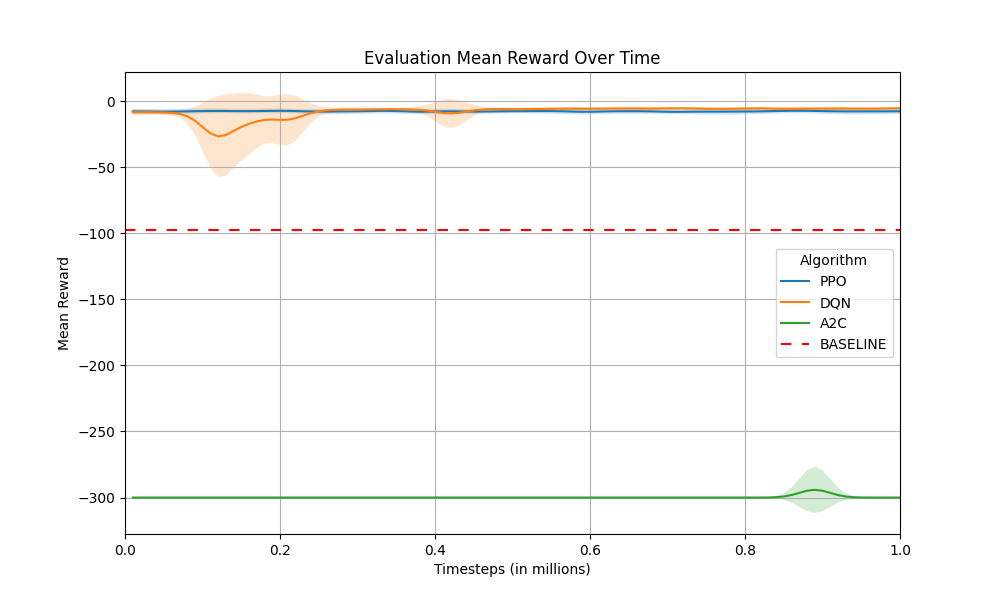}
            \caption{Evaluation mean reward over time.}
        \end{subfigure}%
        \begin{subfigure}{0.75\textwidth}
            \includegraphics[width=\textwidth]{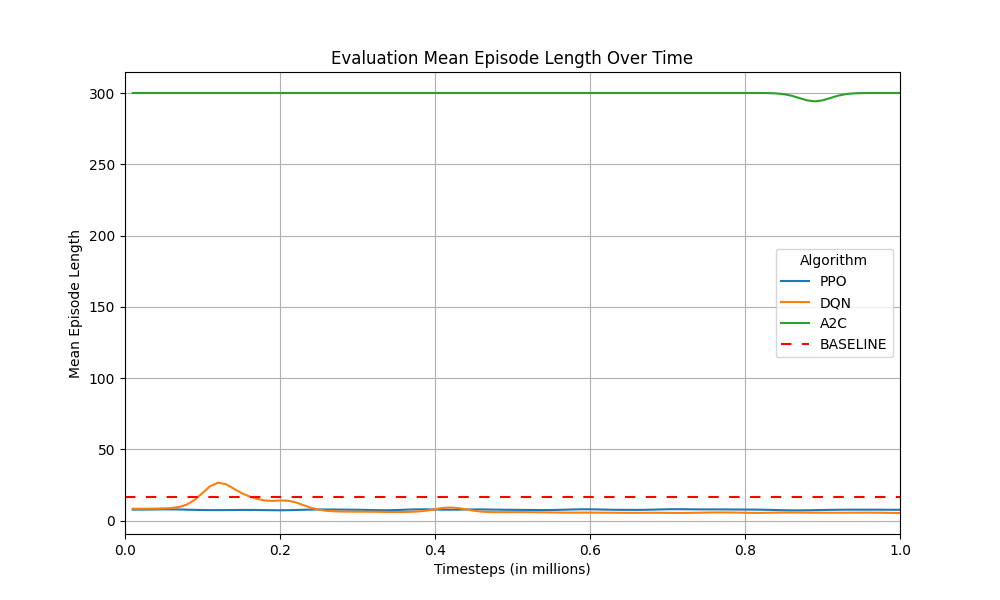}
            \caption{Evaluation mean episode length over time.}
        \end{subfigure}
        \begin{subfigure}{0.75\textwidth}
            \includegraphics[width=\textwidth]{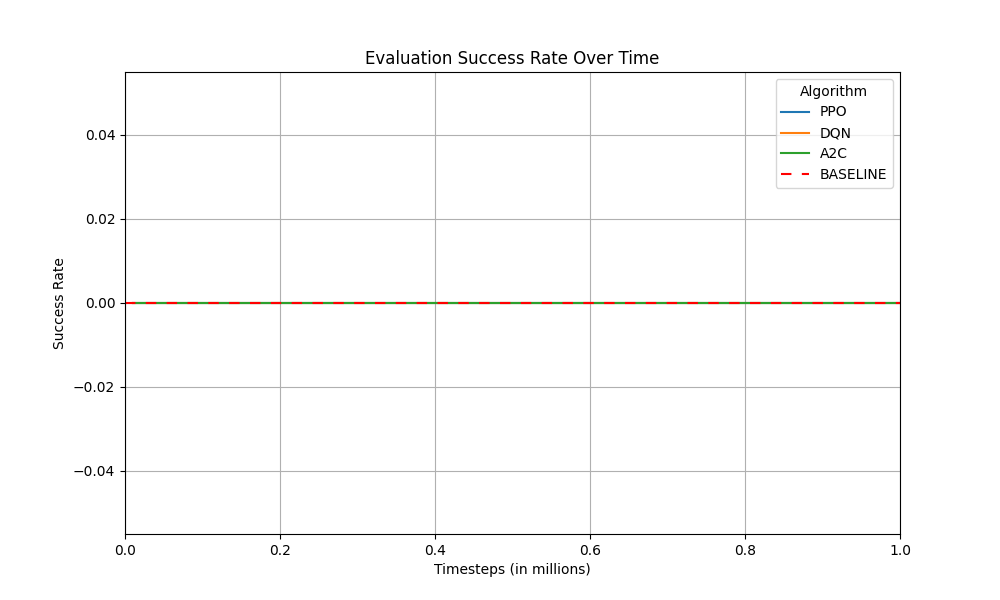}
            \caption{Evaluation success rate over time.}
        \end{subfigure}%
        \begin{subfigure}{0.75\textwidth}
            \includegraphics[width=\textwidth]{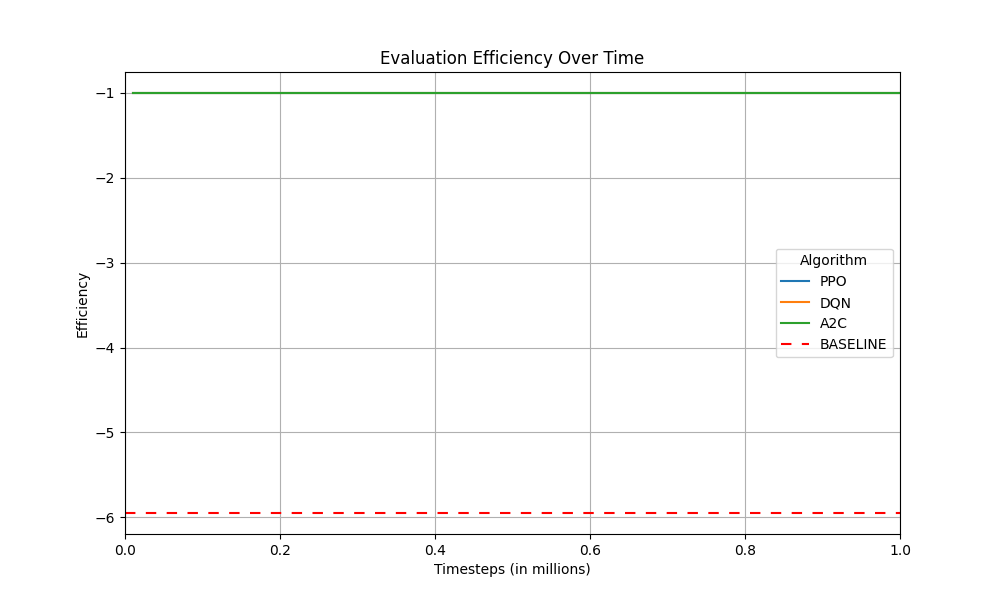}
            \caption{Evaluation efficiency over time.}
        \end{subfigure}
        \caption{Agent performance in experiment \RN{5}.} \label{fig:exp5}
    \end{figure}
\end{landscape}
\subsection{Results}
\cref{fig:exp5} clearly shows that \gls{ppo} and \gls{dqn} have found the same strategy to maximise reward. Once the episode tasks, they find the quickest way to self-destruct, as they predict they would get more negative reward if they attempted any other course of action. \gls{a2c} rotates around its initial position.

\section{Experiment \RN{6}: Extended Training }
We now try training our agents for a longer number of timesteps. We select the more promising algorithms, \gls{ppo} and \gls{dqn}, and the reward function in \vref{eq:rew_budget}, which seems to represent the goal better. We use curriculum learning in the following way to facilitate training: the environment starts with the goal object spawning near the target position, and we slowly increase the randomness radii upon each successful episode completion. We run the training for a total of $9$ million timesteps and we evaluate the agents in fully randomised environments.

\begin{table}[ht]
    \centering
    \caption[Environment parameters for experiment VI]{Environment Parameters}
    \begin{tabular}{p{7cm}p{5.5cm}}
        \toprule
        \textbf{Parameter} & \textbf{Value} \\
        \midrule
        seed & 433854 \\
        randomise & True \\
        clutter\_items & 3 \\
        \bottomrule
    \end{tabular}
    \vspace{1em}
    \centering
    \caption[Baseline performance for experiment VI]{Baseline Performance}
    \begin{tabular}{p{7cm}p{5.5cm}}
        \toprule
        \textbf{Metric} & \textbf{Value} \\
        \midrule
        Mean Reward & $-95.07 \pm 5.69$ \\
        Mean Length & $99.17$ \\
        Mean Efficiency & $-0.96$ \\
        Success Rate & $0.00$ \\
        \bottomrule
    \end{tabular}
    \vspace{1em}
    \caption[Final performance for experiment VI]{Final Performance}
    \begin{tabular}{p{3cm}p{3cm}p{3cm}p{3cm}}
        \toprule
        \textbf{Metric} & \textbf{PPO} & \textbf{DQN (4mln)} & \textbf{DQN} \\
        \midrule
        Timesteps & $1,000,000$ & $4,000,000$ & $1,000,000$ \\
        Mean Reward & $100.14 \pm 70.83$ & $86.62 \pm 81.76$ & $-28.75 \pm 53.06$ \\
        Mean Length & $79.9 \pm 117.04$ & $103.03 \pm 126.22$ & $184.95 \pm 140.96$ \\
        Mean Efficiency & $1.25$ & $0.84$ & $-0.16$ \\
        Success Rate & $0.73$ & $0.58$ & $0.01$ \\
        \bottomrule
    \end{tabular}
\end{table}

\begin{landscape}
    \begin{figure}
        \begin{subfigure}{0.75\textwidth}
            \includegraphics[width=\textwidth]{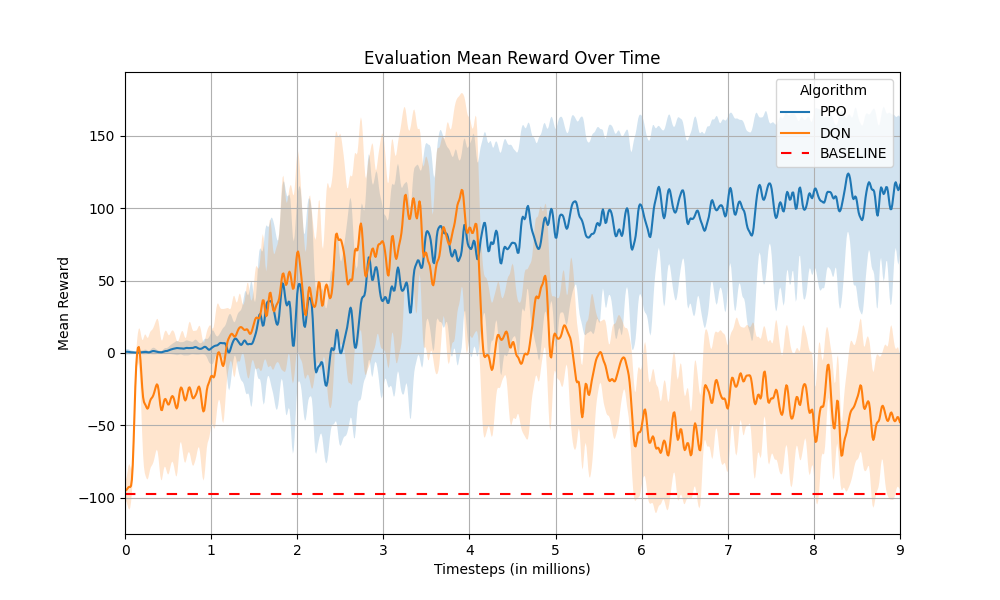}
            \caption{Evaluation mean reward over time.}
        \end{subfigure}%
        \begin{subfigure}{0.75\textwidth}
            \includegraphics[width=\textwidth]{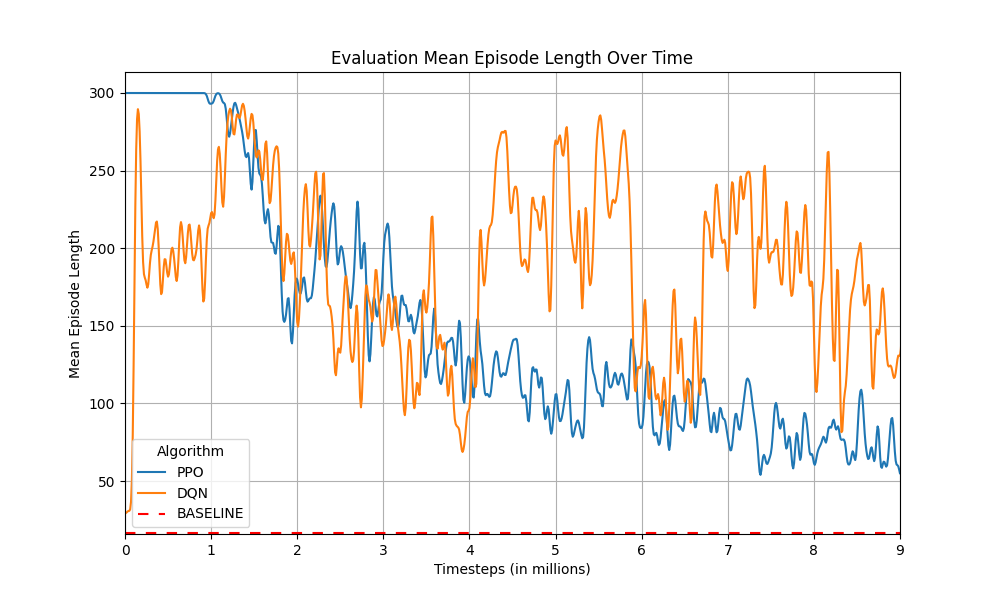}
            \caption{Evaluation mean episode length over time.}
        \end{subfigure}
        \begin{subfigure}{0.75\textwidth}
            \includegraphics[width=\textwidth]{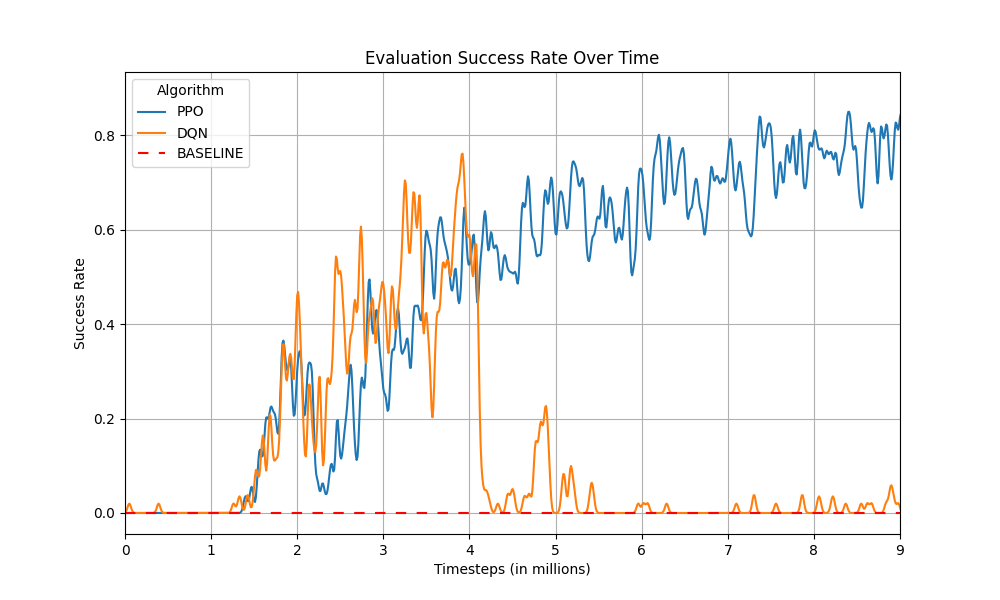}
            \caption{Evaluation success rate over time.}
        \end{subfigure}%
        \begin{subfigure}{0.75\textwidth}
            \includegraphics[width=\textwidth]{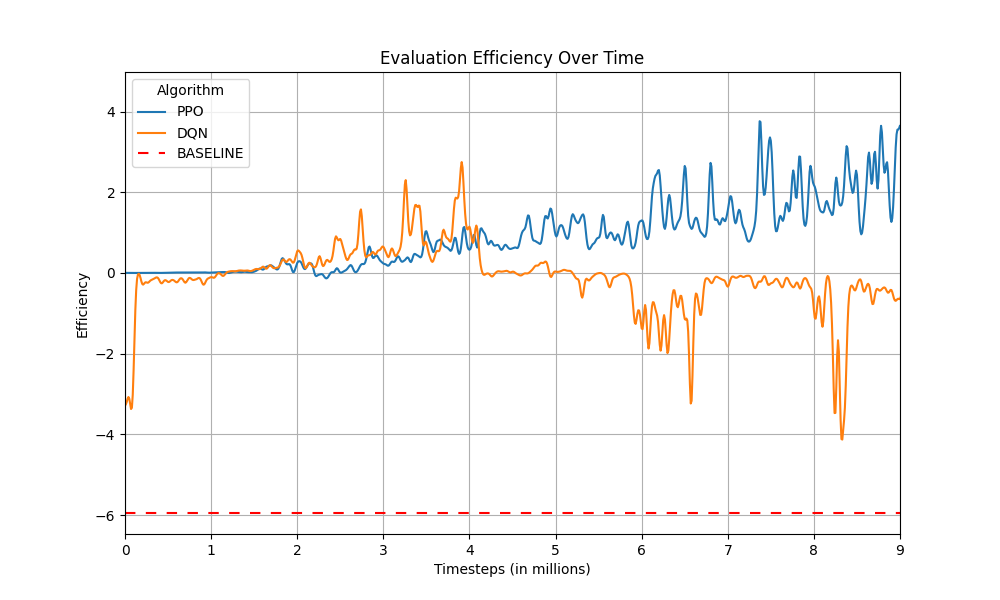}
            \caption{Evaluation efficiency over time.}
        \end{subfigure}
        \caption{Agent performance in experiment \RN{6}.} \label{fig:exp6}
    \end{figure}
\end{landscape}

\subsection{Results}
\cref{fig:exp6} sees a significant drop in performance for \gls{dqn} around $4$ million timesteps. This is a common phenomenon in \gls{dqn} called \textit{catastrophic forgetting}. Remember that we run \gls{dqn} with a replay buffer of $1$ million transitions, as shown in its hyperparameters \vref{tbl:dqn_hyperparams}. Once the buffer is full, \gls{dqn} discards the oldest observations and replaces them with new ones without any prioritisation mechanism. Because of this, it can start overestimating the Q-values of all actions since it has forgotten experiences that lead to failure. Hence, catastrophic forgetting occurs.
On the other hand, we see that \gls{ppo} maintains good performance with a $0.73$ success rate. Still, testing shows that transferring the trained \gls{ppo} model to an environment with $5$ clutter objects plummeted the success rate to $0.22$. Upon investigation, it appears that the \gls{ppo} agent refuses to complete the task whenever there is a clutter object between him and the goal object, preferring instead to avoid failure by rotating in place until the truncation of the episode.
%
%
\pagestyle{plain}
\chapter{Conclusion and Future Work} \label{chapter:conclusion}

\section{Results Discussion}
The results show significant difficulties for all agents in solving the task, even under simplified circumstances. Part of this can be linked to the goal formulation. Using binary sparse rewards settings seems to be the fairest way as it directly rewards the agent for doing what we want. However, this approach didn't work well within the reasonable computational limits we had. Without clear guidance, the agent rarely finds a successful action sequence on its own. This is more pronounced when using the $\epsilon$-greedy exploratory policy, which doesn't maintain a steady exploration strategy.
Similarly, policy-gradient methods often get stuck at a local optimum despite using techniques like entropy regularisation to encourage exploration. Using reward shaping is also problematic, often leading to unforeseen consequences. When we defined the goal as moving an object to a specific position, it was hard to do this without influencing the agent's behaviour. Defining the goal of moving an object to a position is difficult to express as a shaped reward signal without imparting bias to the agent. Suppose we define the reward signal in relation to reducing the distance between the agent and the goal, as we have done in most of our experiments. In that case, it will discourage other valid strategies, such as using other objects to push the goal. The experiments have shown that a poorly shaped reward can cause the agent to terminate the scenario without any attempt to solve it. The task we defined here is fundamentally different from the well-studied Atari games. Atari games are deterministic, and the levels have similar sequences of actions to complete the task. During training, stochasticity is artificially introduced in the environment using techniques like random starts or randomly repeated actions. Atari games also have a well-defined, non-sparse reward signal, namely the underlying game score. A reward increase represents something we want, not some proxy heuristic function for some overarching goal. 

\section{Future Work}
There are many ways to build upon this research. In future, we want to explore different methods to give the agent some prior knowledge. According to \textcite{sutton_reinforcement_2018}, domain knowledge should be given using not complex reward shaping but direct policy initiation. For instance, \textcite{hester_deep_2018} has shown that it is possible to significantly speed up training by using even a few expert demonstrations of successful completion. The training is typically done through standard supervised learning or some other imitation learning method. This technique is particularly effective for the largely deterministic Atari games, where the desired trajectories are somewhat similar to the expert trajectories. Here, similar attempts have yielded no significant change in performance. \textcite{bejjani_learning_2021} proposed a more applicable practical approach where a large number of expert trajectories are generated using classic open-loop planning. Adding imitation learning to the training procedure is a potential area to delve into.
We also aim to improve existing algorithms. We used a classic \gls{dqn} version based on \textcite{mnih_human-level_2015},  but many enhancements have been made since then. Using the latest \gls{dqn} algorithms or trying massively parallel policy-gradient methods could yield better results, as suggested by the extended training experiment. Other improvements could also be made on the feature extraction \gls{dnn} side: using more modern \glspl{cnn} architectures that have better feature extraction can lead to significant improvements in generalisation and performance, as shown in \textcite{cobbe_quantifying_2019}. Another possible avenue for research would be to use a model-based \gls{drl} approach, possibly with learned environment dynamics to predict physics interactions. In addition, here, we have strictly considered a discrete action space. Using a continuous action space makes it possible to utilise \gls{drl} algorithms that would be otherwise incompatible.

\section{Conclusions}
This thesis has provided a general overview and introduction to modern Reinforcement Learning, including all the mathematical background needed for three well-established algorithms: Deep Q-Network, Proximal Policy Optimisation and Advantage Actor-Critic. We developed an environment called the \acrfull{pbge} that simulates a physics-based top-down world reminiscent of a table, where a gripper agent has to push a goal object onto a goal position. We have implemented preprocessing steps to make the environment observation more Markovian, and we have made it compatible with a \gls{cnn} for end-to-end training and automatic feature extraction. We have used the \gls{cnn} as part of \gls{drl} agent architectures and then trained said agents in our environment. We reported the result of six experiments with different combinations of reward function, randomness and clutter obstacles. The inconclusive results show that these Model-free algorithms fail to accomplish the task reliably, especially in a way that would be transferable to real-world systems. Reward shaping has also proven problematic, and our goal formulation as a non-sparse reward signal is unclear. In many cases, the agent preferred to maximise the given heuristic instead of actually completing the task. We concluded by providing recommendations for augmenting our \gls{rl} algorithms, including a call for better feature-extracting networks, better policy initialisation through imitation learning methods and more advanced or more distributed algorithms.
%
\cleardoublepage
\phantomsection
\addcontentsline{toc}{chapter}{\bibname}
\printbibliography

\cleardoublepage
\phantomsection
\addcontentsline{toc}{chapter}{\listfigurename}
\listoffigures    

\cleardoublepage
\phantomsection
\addcontentsline{toc}{chapter}{\listalgorithmname}
\listofalgorithms 

\cleardoublepage
\phantomsection
\addcontentsline{toc}{chapter}{\listtablename}
\listoftables     

\cleardoublepage
\phantomsection
\addcontentsline{toc}{chapter}{List of Acronyms}
\printglossary[style=index,type=\acronymtype, title=List of Acronyms]

\clearpage 
\appendix
\pagenumbering{Roman}  
\setcounter{page}{1}   
\chapter{Summary of Notation}\label{appendix:notation}
\fontsize{10pt}{12pt}\selectfont
\bigskip\noindent
Capital letters are used for random variables, whereas lower case letters are used for the values of random variables and for scalar functions. Quantities that are required to be real-valued vectors are written in bold and in lower case (even if random variables). Matrices are bold capitals.
\begin{tabbing}
\=~~~~~~~~~~~~~~~~~~  \= \kill
\>$\defeq$            \> equality relationship that is true by definition\\
\>$\approx$           \> approximately equal\\
\>$\propto$           \> proportional to\\
\>$\Pr{X\!=\!x}$      \> probability that a random variable $X$ takes on the value $x$\\
\>$X\sim p$           \> random variable $X$ selected from distribution $p(x)\defeq\Pr{X\!=\!x}$\\
\>$\E{X}$             \> expectation of a random variable $X$, i.e., $\E{X}\defeq\sum_x p(x)x$\\
\>$\arg\max_a f(a)$   \> a value of $a$ at which $f(a)$ takes its maximal value\\
\>$\ln x$             \> natural logarithm of $x$\\
\>$e^x$               \> the base of the natural logarithm, $e\approx 2.71828$, carried to power $x$; $e^{\ln x}=x$\\
\>$\Re$               \> set of real numbers\\
\>$f:\X\rightarrow\Y$ \> function $f$ from elements of set $\X$ to elements of set $\Y$\\
\>$\leftarrow$        \> assignment\\
\>$(a,b]$             \> the real interval between $a$ and $b$ including $b$ but not including $a$\\
\\
\>$\e$                \> probability of taking a random action in an \e-greedy policy\\
\>$\alpha, \beta$     \> step-size parameters\\
\>$\gamma$            \> discount-rate parameter\\
\>$\lambda$           \> decay-rate parameter for eligibility traces\\
\>$\ind{\text{\emph{predicate}}}$ \>indicator function ($\ind{\text{\emph{predicate}}}\defeq1$ if the \emph{predicate} is true, else 0)\\
\\
\>In a Markov Decision Process:\\
\>$s, s'$             \> states\\
\>$a$                 \> an action\\
\>$r$                 \> a reward\\
\>$\S$                \> set of all nonterminal states \\
\>$\S^+$              \> set of all states, including the terminal state \\
\>$\A(s)$             \> set of all actions available in state $s$\\
\>$\R$                \> set of all possible rewards, a finite subset of $\Re$\\
\>$\subset$           \> subset of; e.g., $\R\subset\Re$\\
\>$\in$               \> is an element of; e.g., $s\in\S$, $r\in\R$\\
\>$|\S|$              \> number of elements in set $\S$\\
\\
\>$t$                 \> discrete time step\\
\>$T, T(t)$           \> final time step of an episode, or of the episode including time step $t$\\ 
\>$A_t$               \> action at time $t$\\
\>$S_t$               \> state at time $t$, typically due, stochastically, to $S_{t-1}$ and $A_{t-1}$\\
\>$R_t$               \> reward at time $t$, typically due, stochastically, to $S_{t-1}$ and $A_{t-1}$\\
\>$\pi$               \> policy (decision-making rule)\\
\>$\pi(s)$            \> action taken in state $s$ under {\it deterministic\/} policy $\pi$\\
\>$\pi(a|s)$          \> probability of taking action $a$ in state $s$ under {\it stochastic\/} policy $\pi$\\
\\
\>$G_t$               \> return following time $t$\\
\>$h$                 \> horizon, the time step one looks up to in a forward view\\
\>$G_{t:t+n}, G_{t:h}$\> $n$-step return from $t+1$ to $t+n$, or to $h$ (discounted and corrected) \\
\>$\bar G_{t:h}$      \> flat return (undiscounted and uncorrected) from $t+1$ to $h$\\
\>$G^\lambda_t$       \> $\lambda$-return\\
\>$G^\lambda_{t:h}$   \> truncated, corrected $\lambda$-return\\
\>$G^{\lambda s}_t$, $G^{\lambda a}_t$    \> $\lambda$-return, corrected by estimated state, or action, values \\
\\
\>$\p(s',r|s,a)$      \> probability of transition to state $s'$ with reward $r$, from state $s$ and action $a$\\
\>$\p(s'|s,a)$        \> probability of transition to state $s'$, from state $s$ taking action $a$\\
\>$r(s,a)$            \> expected immediate reward from state $s$ after action $a$\\
\>$r(s,a,s')$         \> expected immediate reward on transition from $s$ to $s'$ under action $a$\\
\\
\>$\vpi(s)$           \> value of state $s$ under policy $\pi$ (expected return)\\
\>$\vstar(s)$         \> value of state $s$ under the optimal policy \\
\>$\qpi(s,a)$         \> value of taking action $a$ in state $s$ under policy $\pi$\\
\>$\qstar(s,a)$       \> value of taking action $a$ in state $s$ under the optimal policy \\
\\
\>$V, V_t$            \> array estimates of state-value function $\vpi$ or $\vstar$\\
\>$Q, Q_t$            \> array estimates of action-value function $\qpi$ or $\qstar$\\
\>$\bar V_t(s)$       \> expected approximate action value, e.g., $\bar V_t(s)\defeq\sum_a\pi(a|s)Q_{t}(s,a)$\\
\>$U_t$               \> target for estimate at time $t$\\
\\
\>$\delta_t$          \> temporal-difference (TD) error at $t$ (a random variable) \\
\>$\delta^s_t, \delta^a_t$ \> state- and action-specific forms of the TD error \\
\>$n$                 \> in $n$-step methods, $n$ is the number of steps of bootstrapping\\
\\
\>$d$                 \> dimensionality---the number of components of $\w$\\
\>$d'$                \> alternate dimensionality---the number of components of $\th$\\
\>$\w,\w_t$           \> $d$-vector of weights underlying an approximate value function\\
\>$w_i,w_{t,i}$ \> $i$th component of learnable weight vector\\
\>$\hat v(s,\w)$      \> approximate value of state $s$ given weight vector $\w$\\
\>$v_\w(s)$           \> alternate notation for $\hat v(s,\w)$\\
\>$\hat q(s,a,\w)$    \> approximate value of state--action pair $s,a$ given weight vector $\w$\\
\>$\grad \hat v(s,\w)$\> column vector of partial derivatives of $\hat v(s,\w)$ with respect to $\w$\\
\>$\grad \hat q(s,a,\w)$\> column vector of partial derivatives of $\hat q(s,a,\w)$ with respect to $\w$\\
\\
\>$\x(s)$             \> vector of features visible when in state $s$\\
\>$\x(s,a)$           \> vector of features visible when in state $s$ taking action $a$\\
\>$x_i(s), x_i(s,a)$  \> $i$th component of vector $\x(s)$ or $\x(s, a)$\\
\>$\x_t$              \> shorthand for $\x(S_t)$ or $\x(S_t,A_t)$\\
\>$\w\tr\x$           \> inner product of vectors, $\w\tr\x\defeq\sum_i w_i x_i$; e.g., $\hat v(s,\w)\defeq\w\tr\x(s)$\\
\>$\v,\v_t$           \> secondary $d$-vector of weights, used to learn $\w$ \\
\>$\z_t$              \> $d$-vector of eligibility traces at time $t$ \\
\\
\>$\th, \th_t$        \> parameter vector of target policy \\
\>$\pi(a|s,\th)$      \> probability of taking action $a$ in state $s$ given parameter vector $\th$\\
\>$\pi_\th$           \> policy corresponding to parameter $\th$\\
\>$\grad\pi(a|s,\th)$ \> column vector of partial derivatives of $\pi(a|s,\th)$ with respect to $\th$\\
\>$J(\th)$            \> performance measure for the policy $\pi_\th$\\
\>$\grad J(\th)$      \> column vector of partial derivatives of $J(\th)$ with respect to $\th$\\
\>$h(s,a,\th)$        \> preference for selecting action $a$ in state $s$ based on $\th$\\
\\
\>$b(a|s)$            \> behavior policy used to select actions while learning about target policy $\pi$ \\
\>$b(s)$              \> a baseline function $b:\S\mapsto\Re$ for policy-gradient methods\\
\>$b$                 \> branching factor for an MDP or search tree \\
\>$\rho_{t:h}$        \> importance sampling ratio for time $t$ through time $h$ \\
\>$\rho_{t}$          \> importance sampling ratio for time $t$ alone, $\rho_t\defeq\rho_{t:t}$\\
\>$r(\pi)$            \> average reward (reward rate) for policy $\pi$ \\
\>$\bar R_t$          \> estimate of $r(\pi)$ at time $t$\\
\\
\>$\mu(s)$            \> on-policy distribution over states \\
\>$\bm\mu$            \> $|\S|$-vector of the $\mu(s)$ for all $s\in\S$\\
\>$\norm{v}$          \> $\mu$-weighted squared norm of value function $v$, i.e., $\norm{v}\defeq\sum_{s\in\S} \mu(s)v(s)^2$\\
\>$\eta(s)$           \> expected number of visits to state $s$ per episode\\
\>$\Pi$               \> projection operator for value functions \\
\>$B_\pi$             \> Bellman operator for value functions \\
\\
\>${\bf A}$           \> $d\times d$ matrix ${\bf A}\defeq\E{\x_t\bigl(\x_t-\g\x_{t+1}\bigr)\tr}$\\
\>${\bf b}$           \> $d$-dimensional vector ${\bf b}\defeq\E{R_{t+1}\x_t}$\\
\>$\w_{\rm TD}$       \> TD fixed point $\w_{\rm TD}\defeq {\bf A}^{-1}{\bf b}$ (a $d$-vector\\
\>${\bf I}$           \> identity matrix\\
\>${\bf P}$           \> $|\S|\times |\S|$ matrix of state-transition probabilities under $\pi$\\
\>${\bf D}$           \> $|\S|\times |\S|$ diagonal matrix with $\bm\mu$ on its diagonal\\
\>${\bf X}$           \> $|\S|\times d$ matrix with the $\x(s)$ as its rows\\
\\
\>$\bar\delta_\w(s)$  \> Bellman error (expected TD error) for $v_\w$ at state $s$\\
\>$\bar\delta_\w$, BE \> Bellman error vector, with components $\bar\delta_\w(s)$\\
\>$\MSVEm(\w)$        \> mean square value error $\MSVEm(\w)\defeq\norm{v_\w-\vpi}$\\
\>$\MSBEm(\w)$        \> mean square Bellman error $\MSBEm(\w)\defeq\norm{\bar\delta_\w}$\\
\>$\MSPBEm(\w)$       \> mean square projected Bellman error $\MSPBEm(\w)\defeq\norm{\Pi\bar\delta_\w}$\\
\>$\MSTDEm(\w)$       \> mean square temporal-difference error $\MSTDEm(\w)\defeq\EE{b}{\rho_t\delta_t^2}$ \\
\>$\MSREm(\w)$        \> mean square return error\\
\end{tabbing}
\end{document}